\documentclass[11pt]{article}

\usepackage{PRIMEarxiv}
\usepackage{longtable}
\usepackage[utf8]{inputenc} % allow utf-8 input
\usepackage[T1]{fontenc}    % use 8-bit T1 fonts
\usepackage{url}            % simple URL typesetting
\usepackage{booktabs}       % professional-quality tables
\usepackage{amsfonts}       % blackboard math symbols
\usepackage{nicefrac}       % compact symbols for 1/2, etc.
\usepackage{microtype}      % microtypography
\usepackage{lipsum}
\usepackage{fancyhdr}       % header
\usepackage{graphicx}       % graphics
\graphicspath{{media/}}     % organize your images and other figures under media/ folder
\usepackage{tabularx}
\usepackage{listings} % Include the package at the beginning
\usepackage{fancyvrb} % Import the fancyvrb package

\usepackage[numbers]{natbib}
\usepackage{amsmath,amssymb}
\usepackage{amsthm}
\usepackage{mathtools}
\usepackage{xspace}
\usepackage[noend]{algorithmic}
\usepackage[ruled,vlined]{algorithm2e}
\usepackage{enumitem}

\usepackage{pdfpages}

\usepackage{chngcntr}
\usepackage{mdframed} % For creating framed boxes
\usepackage{caption}

\usepackage{color}
\usepackage{xcolor}

\usepackage{subfigure}
\usepackage[most]{tcolorbox}
\usepackage[frozencache=true, finalizecache=false, cachedir=./minted-cache]{minted}

\usepackage{float}
\usepackage{alltt}
\usepackage{soul}
\usepackage{fancyvrb}
\usepackage{multirow}
\usepackage[final]{hyperref}

\usepackage{csquotes}

\newcounter{textbox}
\renewcommand{\thetextbox}{\arabic{textbox}} % Adjust according to where you reset this

\newcommand{\textboxcaption}[1]{
  \refstepcounter{textbox}
  \noindent\small\textbf{Text Box \thetextbox:} #1\par\medskip
}
\usepackage{listings} % For listings, which allows more control over verbatim text
\newcommand{\VG}{GPT-4\xspace}

\newcommand{\XLora}{X-LoRA\xspace}
\newcommand{\BioMixtral}{BioinspiredLLM-Mixtral\xspace}
\newcommand{\MistralOpenOrca}{Mistral-7B-OpenOrca\xspace}
\newcommand{\ZephyrBeta}{Zephyr-7B-$\beta$\xspace}

\newcommand{\VGV}{GPT-4V\xspace}

\tcbset{
  custombox/.style={
    width=474.18663pt,
    top=3pt,%top=10pt,
    fonttitle=\bfseries\small\sffamily, % Bold, larger, and sans serif title font
    colback=white,
    colframe=black,
    colbacktitle=black,
    boxrule=0.5pt, % Border thickness
    colback=white, % Background color
    enhanced,
    rounded corners, % Rounded corners
    center,
    boxed title style={
            sharp corners,
            size=small,
            colframe=blue!75!black, % Title frame color
        },
    attach boxed title to top left={yshift=-0.1in,xshift=0.15in},
    boxed title style={boxrule=0pt,colframe=white,},
  }
}
\newtcolorbox{LLMbox}[2][]{custombox,title=#2,#1}

\tcbset{
  customboxmultipage/.style={
    width=474.18663pt,
    top=3pt,%top=10pt,
    fonttitle=\bfseries\small\sffamily, % Bold, larger, and sans serif title font
    colback=white,
    breakable,
    colframe=black,
    colbacktitle=black,
    boxrule=0.5pt, % Border thickness
    colback=white, % Background color
    enhanced,
    rounded corners, % Rounded corners
    center,
    boxed title style={
            sharp corners,
            size=small,
            colframe=blue!75!black, % Title frame color
        },
    attach boxed title to top left={yshift=-0.1in,xshift=0.15in},
    boxed title style={boxrule=0pt,colframe=white,},
  }
}
\newtcolorbox{LLMboxmultipage}[2][]{customboxmultipage,title=#2,#1}

\newtcbox{\mybox}[1][green]{on line,
arc=0pt,outer arc=0pt,colback=#1!10!white,colframe=#1!50!black,
boxsep=0pt,left=0pt,right=0pt,top=0pt,bottom=0pt,
boxrule=0pt,bottomrule=0pt,toprule=0pt}

\definecolor{aigold}{RGB}{244,210, 1} 
\definecolor{aigreen}{RGB}{210,244,211} 

\sethlcolor{aigreen}

\definecolor{aired}{RGB}{255,180,181}

\definecolor{lightred}{rgb}{1,0.9,0.9} % Light red color

\newcommand{\hlred}[1]{{%
  \sethlcolor{lightred}% Set the highlight color to light red
  \hl{#1}% Highlight the argument text
}}

\title{Accelerating Scientific Discovery with Generative Knowledge Extraction, Graph-Based Representation, and Multimodal Intelligent Graph Reasoning

}

\author{
  Markus J. Buehler\thanks{Massachusetts Institute of Technology (MIT), 77 Mass. Ave 1-165, Cambridge, MA 02139, USA} \\
  \\
  \texttt{mbuehler@MIT.EDU} \\
}

\begin{document}
\maketitle

\begin{abstract}
Leveraging generative Artificial Intelligence (AI), we have transformed a dataset comprising 1,000 scientific papers focused on biological materials into a comprehensive ontological knowledge graph. Through an in-depth structural analysis of this graph, we have calculated node degrees, identified communities along with their connectivities, and evaluated clustering coefficients and betweenness centrality of pivotal nodes, uncovering fascinating knowledge architectures. We find that the graph has an inherently scale-free nature, shows a high level of connectedness, and can be used as a rich source for downstream graph reasoning by taking advantage of transitive and isomorphic properties to reveal insights into unprecedented interdisciplinary relationships that can be used to answer queries, identify gaps in knowledge, propose never-before-seen material designs, and predict material behaviors. Using a large language embedding model we compute deep node representations and use combinatorial node similarity ranking to develop a path sampling strategy that allows us to link dissimilar concepts that have previously not been related. One comparison revealed detailed structural parallels between biological materials and Beethoven's 9th Symphony, highlighting shared patterns of complexity through isomorphic mapping. In another example, the algorithm proposed an innovative hierarchical mycelium-based composite based on integrating path sampling with principles extracted from Kandinsky's `Composition VII' painting. The resulting material integrates an innovative set of concepts that include a balance of chaos and order, adjustable porosity, mechanical strength, and complex patterned chemical functionalization. We uncover other isomorphisms across science, technology and art, revealing a nuanced ontology of immanence that reveal a context-dependent heterarchical interplay of constituents. Because our method transcends established disciplinary boundaries through diverse data modalities (graphs, images, text, numerical data, etc.), graph-based generative AI achieves a far higher degree of novelty, explorative capacity, and technical detail, than conventional approaches and establishes a widely useful framework for innovation by revealing hidden connections.   
\end{abstract}

% keywords 
\keywords{Language modeling \and Scientific AI \and Biomaterials \and Bio-inspired materials \and Inverse problems \and Generative AI \and Materials science \and Multidisciplinary \and Natural Language Processing \and Computer vision \and Materials informatics}

\section{Introduction}

In the evolving landscape of knowledge discovery, the intersection of computational techniques with data mining has become an active area of investigation. One of the grand challenges is to find ways by which information mined from diverse sources can be modeled, and understood, and used as a basis for further discovery to expand the horizon of understanding. Due to the sheer volumes of data, this has remained challenging, especially when developing strategies to extrapolate from existing knowledge towards never-before-seen ideas or behaviors. 
Through the use of large language models (LLMs)~\cite{Vaswani2017AttentionNeedc,Touvron2023LlamaModels,OpenAI2023GPT-4Report,Chowdhery2022PaLM:Pathways,Jiang2023Mistral7B,Gunasekar2023TextbooksNeed} in scientific analysis, the development of new ideas and hypotheses has emerged as as a possible approach ~\cite{Bubeck2023SparksGPT-4,Buehler2023MechGPTModalities,Nejjar2023LLMsAnalysis,Buehler2023GenerativeDesign,Luu2023BioinspiredLLM:Materials,Luu2023GenerativeSolvents,Buehler2023MeLMProblemsb,Ge2023OpenAGI:Experts}. An area of great interest is in-context learning, the ability of a model to adapt its responses based on the context provided in the prompt. The context can include various forms of data, examples, or any relevant information. The model uses this immediate context to  perform a wide range of tasks without needing task-specific training data or fine-tuning. 
Thereby LLMs have been shown a capacity to synthesize a level of sophisticated understanding, such as translating between languages a model has not been trained on~\cite{gemini15report}. For example, when the Gemini 1.5 model, released in 2024, was fed a grammar manual for Kalamang (a language spoken by very few individuals), the model acquired the ability to translate English to Kalamang with proficiency comparable to that of a human studying the same material. This example demonstrates that LLMs can effectively learn from context provided, where new data provided (in this case, a grammar manual) endows the model with a new capability. Data provided in the context can originate from other sources, and as done in this study we provide context extracted from graphs that provide a delineation of relationship between distinct concepts.
These and other emergent behaviors point to a realistic possibility that powerful AI systems can be used, potentially, for knowledge discovery. We postulate that in order to achieve that, the provision of proper context to facilitate the act of discovery, is essential. Here, proper context refers to sub-graphs extracted from larger graphs that allow us to take advantage of relationships between concepts during inference, to trigger the model to generate complex responses.

Earlier research has used category theory to develop ontological graph-based representations of knowledge using graphs~\cite{MacLane1998CategoriesMathematician,Marquis2019CategoryTheory,Spivak2011CategoryNetworks,Spivak2011ReoccurringAnalogies,Giesa2012CategoryDesign}. We build on this general concept and develop ontological representations using natural language processing and generative AI that spans multiple modalities (text, images, numerical data, etc.). Unlike in the earlier work, here we use a generative AI framework to discover and utilize the graphs. 
Our aim is to utilize generative AI to connect different areas of knowledge by focusing the generative task on finding analogies, or by tasking the AI model to identify, propose or explain relationships between disparate concepts or knowledge. Innovation, scientific discovery as well as many creative processes are indeed based on these underpinning mechanisms, whether the aim is to find a solution to a problem, to explain an observation, or to predict behaviors of systems that have not been studied before. These tasks can be viewed as a sort of path finding process where we want to uncover one or more rational ways to connect ideas. From a theoretical perspective, this process can be described as a graph where nodes and edges provide a way to capture the relationships and by extension, the pathways towards deeper understanding and ultimately, discovery. This strategy thereby guides the discovery process through graph-based reasoning. If the graph structure used for this process can be constructed with rigorous methodologies (e.g. data mining, embedding models, etc.), the entire mechanism of discovery can be leveraged by an autonomous system that intelligently explores new connections, new insights, and new possibilities. 
This represents a model of `thinking' can forms a rigorous foundation to enable innovation. This general approach also relates with what is referred to as `augmented thinking' proposed in~\cite{ottino2022nexus}, especially since through the use of generative AI we can easily incorporate a wealth of diverse data sources, methods and context. In fact, as emphasized in the concept of `augmented thinking', a strong focus on the interface of disciplines to generate new ideas, discoveries, and technological advancements is critical. 

We hypothesize that generative AI can be effective in solving these tasks if asked to `think' about structured graph representations. Naturally, multimodal AI systems like LLMs can ingest a wealth of additional representations from images to instructions, and more. 
Graph representations also provides a rich set of theoretical tools through graph theory~\cite{Rosen1988DiscreteApplications,Bondy2008GraphTheory}. For instance, we can use graph theory to extract  nodes in a network that play significant roles in its functionality, connectivity, or efficiency. Their importance can be understood and quantified using various measures, including betweenness centrality, degree centrality, and closeness centrality, among others. Betweenness centrality is particularly insightful for identifying critical nodes. It measures the extent to which a node lies on the shortest paths between other nodes in the network. Nodes with high betweenness centrality scores are seen as critical because they serve as important bridges within the network. If these nodes are removed or fail, they significantly disrupt the flow within the network, making them crucial for maintaining the network's overall connectivity. Applied to graph representation of knowledge, this can be used to identify research topics, or to formulate hypotheses. Critical nodes can guide the exploration of new knowledge since they are situated at the intersection of important concepts, and hence they can offer insights into related areas of study or suggest new connections that were not immediately apparent or that are, in the state of current knowledge, weak links. It is noted that further work could be done to explore the use of, and definition, of, various measured of graph and node properties in the development of reasoning strategies. For instance, measures of median betweenness centrality could be used to assess the relative importance of a node for a given sub-graph or the whole graph.

A basic element that we can take advantage of in graph reasoning are transitive relationships in graphs. 
Given a set of nodes \(N\) and a set of edges \(E\) in a graph \(G\), the transitive property can be succinctly defined as follows: If an edge exists from node \(A\) to node \(B\) (\(A \rightarrow B\)), and an edge exists from node \(B\) to node \(C\) (\(B \rightarrow C\)), then a transitive relationship exists such that \(A\) is connected to \(C\) (\(A \rightarrow C\)) through \(B\).
Formally, this is represented as:
$$
\forall A, B, C \in N, \quad (A \rightarrow B) \wedge (B \rightarrow C) \Rightarrow (A \rightarrow C)
$$
As an example, within the context of biology and the study of silk proteins:
\begin{align*}
\text{1. } & A \rightarrow B: \quad \text{Gene \(A\) encodes a protein \(B\) crucial for silk strength.} \\
\text{2. } & B \rightarrow C: \quad \text{Protein \(B\) interacts with another protein \(P\) to form silk fibers \(C\).}
\end{align*}
By the transitive property, we deduce:
\begin{align*}
\text{3. } & A \rightarrow C: \quad \text{Gene \(A\) indirectly contributes to silk fiber formation through protein \(B\)'s interaction with protein \(P\).}
\end{align*}
Now, extending this framework:
$$
\text{4. } C \rightarrow D: \text{Silk fibers (\(C\)) can be utilized as scaffolds in wound healing (\(D\)).}
$$
By the transitive property, we can deduce a chain of relationships:
\begin{align*}
\text{5. } A \rightarrow D: & \text{ Gene \(A\), through the production and interaction} \\
& \text{of proteins \(B\) and \(P\), and the application of silk fibers \(C\)} \\
& \text{as scaffolds, indirectly can contribute to} \\
& \text{wound healing technology (\(D\)).}
\end{align*}
This example elucidates how the transitive property assists in unraveling direct and indirect interactions between concepts. To make the point more clear in terms of how information and knowledge is accessible, let us assume that a subset of knowledge is contained in one paper that results in a small sub-graph (e.g. nodes $A$, $B$ and $C$, that is, the paper may cover how gene $A$ encodes protein $B$ and how that interacts with protein $P$ to form silk fibers, $C$). Another paper may discuss the relationship between silk fibers $C$ and would healing $D$, but without discussing how silk fibers form. By combining the  sub-graphs from these two papers into a large, integrated graph we can take advantage of the transitive properties as exemplified above and thereby discover new relationships that link $A$, $B$, $P$, $C$ and $D$.

\begin{figure}
    \centering
    \includegraphics[width=1\linewidth]{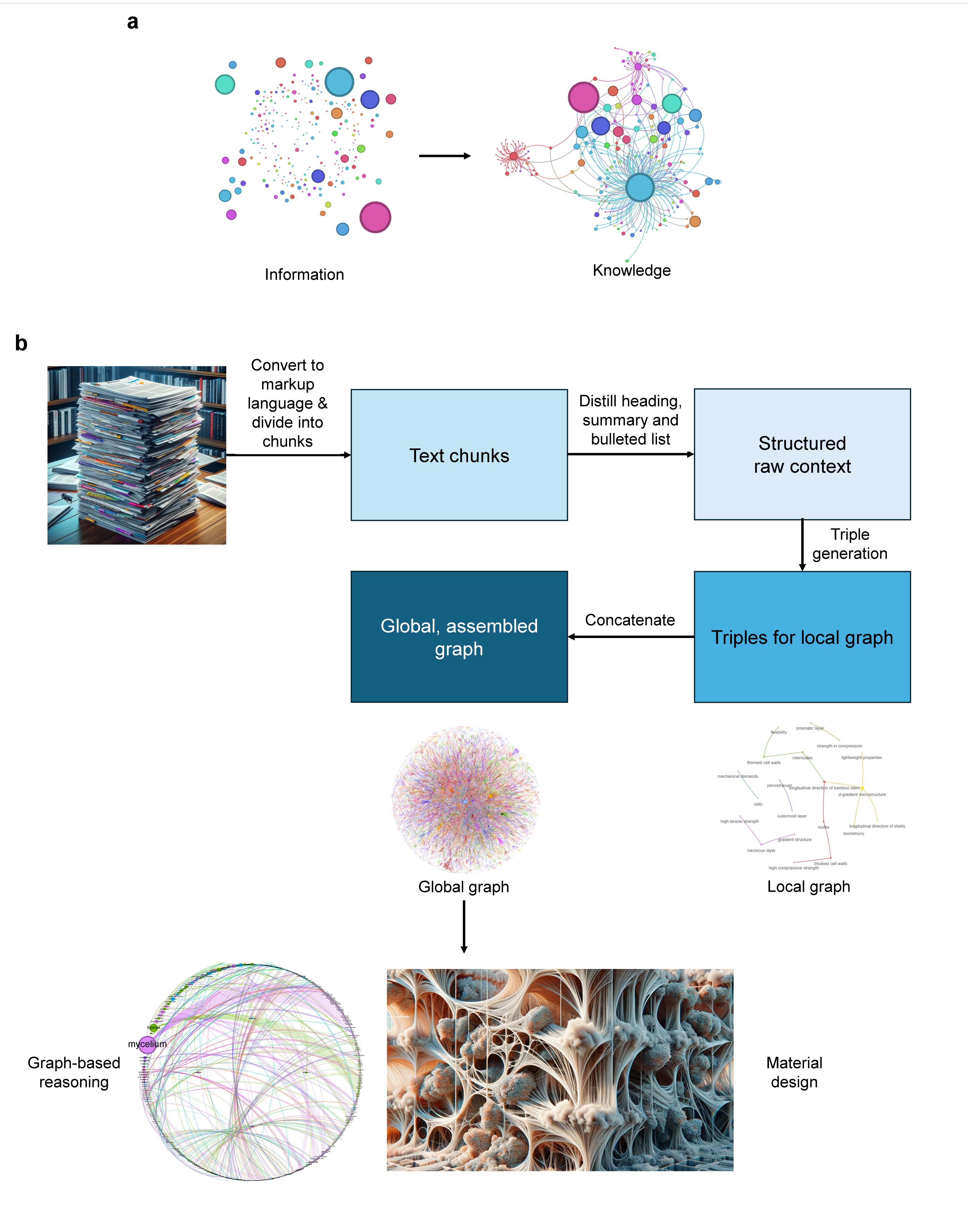}
    \caption{Overview of the approach used here. Panel a depicts the strategic objective to convert information (the answer to "who," "what," "where," and "when" questions) into knowledge (about "how"). While information is relatively easily accessible and can be recorded in books, it can be transmitted easily. Knowledge, in contrast, is typically harder to communicate, and can be hard to transfer from one person to another. Panel b shows how we solve this problem, by first converting a set of scientific articles into markup language, then into text chunks, which then form the basis to distill their content into a concise scientific summary. The raw content then forms the basis to generate triples for a graph, first created at the level of each of the text chunks, and then assembled into a global graph by concatenating all local graphs. }
    \label{fig:Fig_1_intro}
\end{figure}

Figure~\ref{fig:Fig_1_intro} visualizes an overview of the approach used in this study, reflecting a flowchart of key elements that range from knowledge extraction, distillation, graph construction, and reasoning. Figure~\ref{fig:Fig_1_intro}a specifically visualizes the strategic objective to convert information (the answer to ``who," ``what," ``where," and ``when"-type questions) into knowledge (about ``how"). Information is relatively easily accessible and can be recorded in books, papers, reports, can extracted using data mining techniques, and can be transmitted easily. Knowledge, in contrast, is typically harder to communicate, and can be difficult to transfer from one person to another or from humans to AI systems, and back. 
We use a computational scheme based on a series of generative processing steps to construct a graph representation of knowledge, which then forms the basis for wide-ranging analyses in a variety of downstream tasks. These include running queries on the graph to answer questions, connecting disparate concepts within the graph by finding the shortest path (or a set of alternative paths), as well as complementing the graph with new knowledge derived either from separate generative processes or physics-based simulations, or adding additional original data sources (e.g. papers, reports, etc.) to it. 

We use several large language models (LLMs) in this paper, including open source models and state-of-the-art proprietary models like GPT-4/V and Claude-4 Opus. We discuss the rationale behind these choices in the paper, emphasizing the need of different capabilities for distinct tasks and a desire to better understand strengths and weaknesses of each. For instance, one of the open-source models used is X-LoRA~\cite{Buehler2024X-LoRA:Design}, a LLM inspired by biological principles, as it is capable of dynamically rearranging its own structure before responding to a task. This is achieved by augmenting the conventional inference strategy of LLMs to feature two forward passes, which trains the model to first think about the question and how it may reconfigure itself before responding.  This implements a simple reflection of `self-awareness' whereby the model is able to adapt its own self to best solve a task. As a result, the model (even though it has a relatively small parameter count of ~7 billion parameters) can reason across diverse scientific domains (biological materials, math, physics, chemistry, logic, mechanics, etc.), significantly enhancing its capacity for generating innovative solutions. 

The plan of this paper is as follows. We first review the construction of the global ontological knowledge graph from a corpus of scientific papers. We then provide a detailed analysis of the resulting graph structure and its properties. This is followed by a series of systematic experiments in which we exploit the graph for quantitative analyses. These include using reasoning over information extraction, identification of research opportunities, predicting new materials designs with exquisite details about molecular, chemical, mechanical and structural features, as well as a rigorous method to relate disparate knowledge domains for scientific discovery at the frontier of scientific understanding. The  experiments are geared towards expanding the horizon of knowledge, to identify, and reason over new hypotheses, predicted behaviors, and innovative ideas.  For instance, we will show how this approach can relate seemingly disparate concepts such as Beethoven's 9th symphony with bio-inspired materials science. Details on the numerical methods, including links to the code base, are included in the Materials and Methods section.

\section{Results and discussion}

The process of developing the model consists of the following steps, as shown in Figure~\ref{fig:Fig_1_intro}b:
\begin{enumerate}
    \item Identification of a corpus of knowledge, here developed via a literature analysis (for details on how the set of papers was identified, see \cite{Luu2023BioinspiredLLM:Materials})
    \item Distillation of knowledge into structured raw context that address specific aspects of scientific understanding, including a summary of the subject, reasoning and details that are critical
    \item Generation of triples for graph construction (concepts and their relationships), based on the structured raw context that resulted from the distillation process 
    \item Concatenation of all triples into a global graph
    \item Analysis of the global graph through node embeddings using a deep learning text encoder model, simplification steps, and optionally removing small or unconnected fragments (e.g. to consider only the giant component).
    \item Utilization of the ontological knowledge graph for multimodal graph reasoning 
\end{enumerate}
This process is complemented by a variety of additional steps, such as adding new graphs or sub-graphs to the global graph, along with various methods to extract sub-graphs from the global graphs via in-context queries (e.g. identifying multiple ranked shortest-path traversals, subgraphs based on multi-hop analyses, and others). 

\subsection{Construction and analysis of the global graph}

\begin{figure}
    \centering
    \includegraphics[width=.8\linewidth]{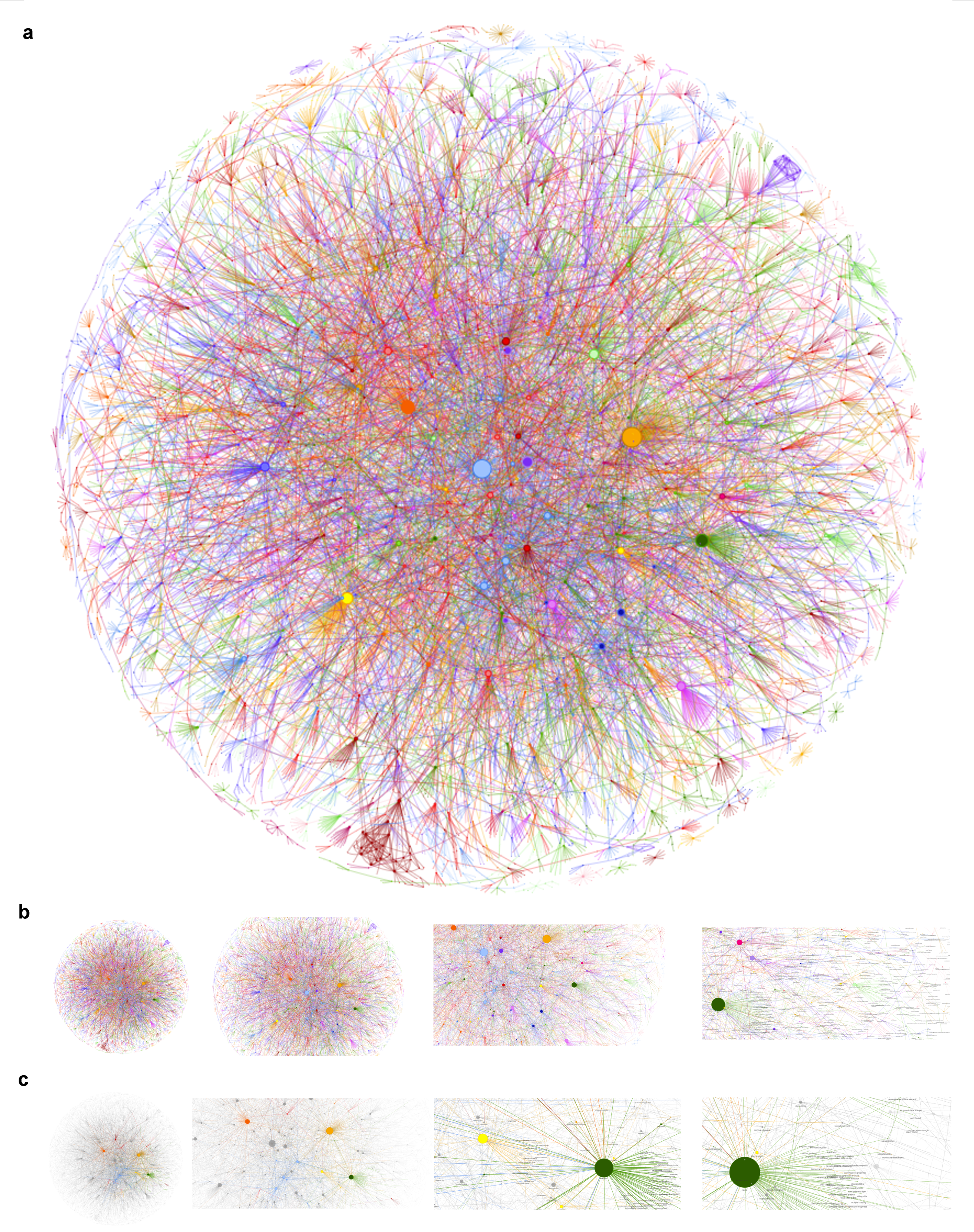}
    \caption{Overview of the global graph (panel a), multiple magnifications (panel b) and illustration of the deep and wide connectivity of nodes (panel c). Panel b depicts the entire graph, followed by successively zoomed in views of the graph structure. At the highest magnification, individual nodes and node labels become visible. Panel c shows a similar progression, albeit with one of the nodes, 'nacre', highlighted (and the rest greyed out), revealing the wide-ranging connections across the global graph. Such highly connected nodes are essential for the knowledge graph's functionality, acting as central hubs that enhance its ability to represent, access, and discover scientific knowledge. }
    \label{fig:Fig_2}
\end{figure}

Figure~\ref{fig:Fig_1_intro}b showcases how we process information into knowledge through a series of natural language processing steps (details see Materials and Methods). This is done by first converting a set of scientific articles into markup language, then into text chunks, which then form the basis to distill their content into a concise scientific summary. The raw content then forms the basis to generate triplets for graphs, first created at the level of each of the local text chunks, and then assembled into a global graph by concatenating all triples. 
Figure~\ref{fig:Fig_2}a shows the global graph and an illustration of the deep and wide connectivity of nodes. Figure~\ref{fig:Fig_2}bdepicts the entire graph (left), followed by successively zoomed in views of the graph structure. At the highest magnification, individual nodes and node labels become visible. Similarly, Figure~\ref{fig:Fig_2}c shows a progression over increasing magnification, albeit with the node with label ``nacre" highlighted (and the rest greyed out), revealing the wide-ranging connections across the global graph. These intricate connections will later be explored to identify complex never-before-seen relationships between concepts. In terms of highly connected nodes like `nacre', these are examples of outliers with significantly higher degrees compared to the average node. These nodes play crucial roles in the network due to their extensive connections, acting as central hubs that facilitate the integration and dissemination of knowledge across the graph.
The figure illustrates the knowledge graph with nodes representing scientific concepts and edges representing the relationships between them. The size of each node corresponds to its degree (number of connections), highlighting nodes with high connectivity.

\begin{figure}
    \centering
    \includegraphics[width=\linewidth]{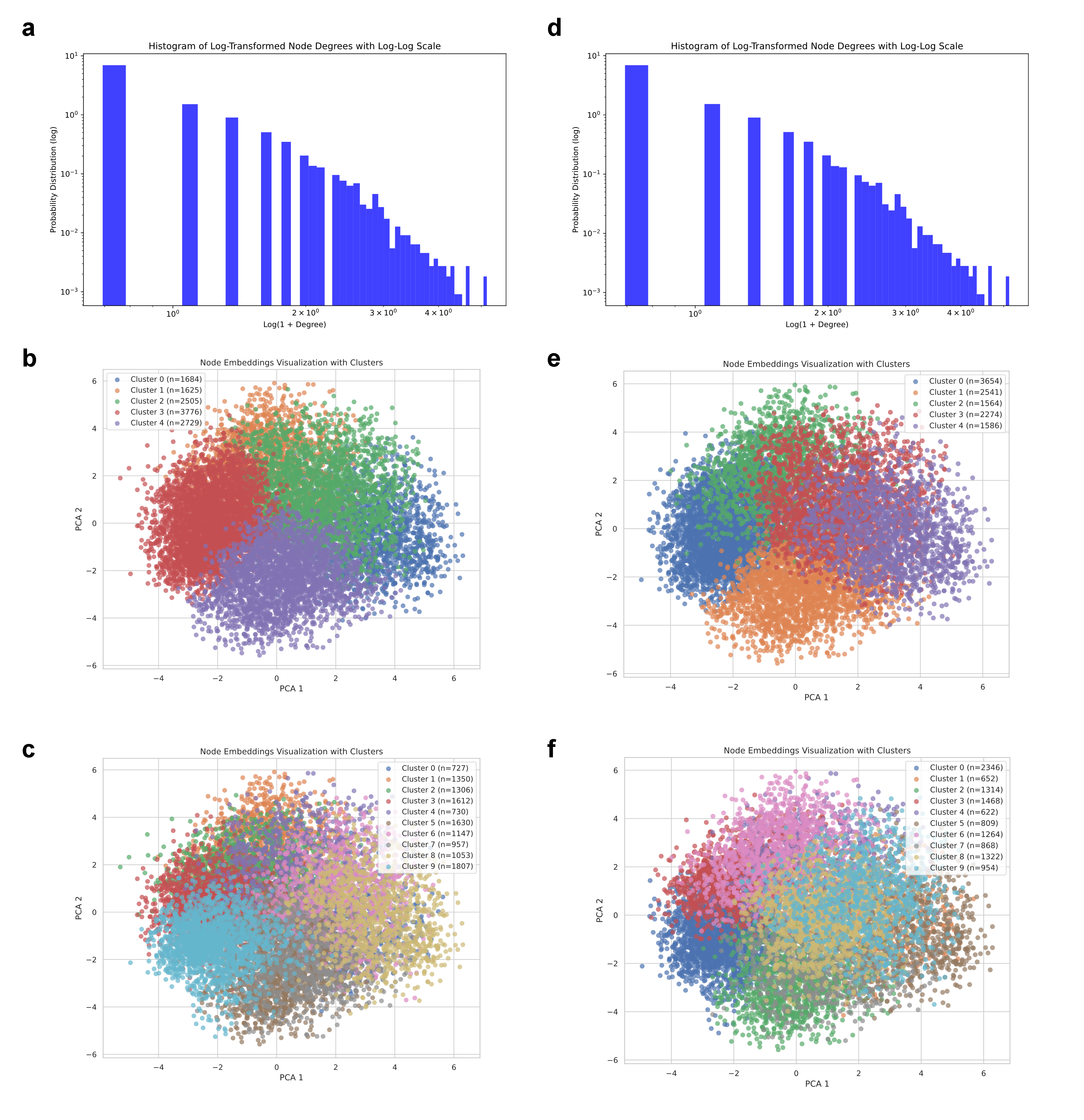}
    \caption{Summary of graph statistics of the global graph, complementing the analysis in Table~\ref{tab:table_graphproperties_comb}. Panel a shows a log-log plot of the degree distribution, and panels b and c a principal component analysis of the node embeddings (for 5 clusters in b and 10 clusters in c). Panels d-f show the same analysis, but for the giant component of the graph only.
    For the plots in panels and d We use \texttt{log1p} to transform node degrees before plotting provides a clearer and more interpretable visualization by handling zero values and reducing skewness. This transformation spreads the data more evenly across the histogram bins, highlighting patterns and variability that may be obscured when plotting raw degrees directly.} 
    \label{fig:Fig_3}
\end{figure}

Figure~\ref{fig:Fig_3} depicts a summary of graph statistics of the global graph. Figure~\ref{fig:Fig_3}a shows a log-log plot of the degree distribution, and Figure~\ref{fig:Fig_3}b-c show results of a principal component analysis (PCA) of the node embeddings. 
Table~\ref{tab:table_graphproperties_comb} shows a summary of the global graph properties, including a subset of analysis for the giant component of the global graph, respectively. The analysis of the giant component is significant since it reflects largest connected component that contains a significant portion of the entire network's nodes. The existence of a giant component, as seen here, often indicates that the network has reached a critical level of connectivity, enabling extensive interaction or communication across a large portion of the network. For the specific application studied here, the giant component of the knowledge graph represents its most interconnected and hence information-rich part. Hence, it likely plays a crucial role in knowledge representation, accessibility, and discovery and is useful for subsequent analysis (all graph data provided via Supplementary Information). The significance of the giant component lies in its representation of a connected subgraph that encompasses the majority of the nodes and edges in the network. This indicates that a large proportion of nodes are interconnected, demonstrating extensive connectivity across the network. Such a structure is significant as it ensures that the model captures and establishes relationships between a vast number of different concepts, facilitating comprehensive and cohesive knowledge representation.

The giant component contains 11,878 nodes and 15,396 edges, which constitutes the majority of the global graph (12,319 nodes and 15,752 edges). This indicates that the giant component captures the primary structure of the network, with most nodes and connections included. The average node degree is rather similar for both the global graph (2.56) and the giant component (2.59). This similarity suggests that the overall connectivity pattern is preserved in the giant component, even though it excludes some smaller components. Both the maximum and minimum node degrees are identical in the global graph and the giant component (171 and 1, respectively). This indicates that the most connected and least connected nodes are all part of the giant component. The median node degree is 1 for both the global graph and the giant component, suggesting that half of the nodes have at most one connection. This is consistent with a sparse network where many nodes have minimal connectivity. The density of the giant component (0.00022) is slightly higher than that of the global graph (0.00021). This small increase in density indicates that the giant component is slightly more interconnected than the overall graph, likely due to the exclusion of small, sparsely connected components. The number of communities in the giant component (80) is fewer than in the global graph (109), suggesting that some smaller communities outside the giant component contribute to the overall count. The giant component's reduced number of communities indicates a more integrated structure.

Hence, the data in Table~\ref{tab:table_graphproperties_comb} supports the discussion concerning the giant component's significance in the global graph. The degree distributions for both the global graph and the giant component are likely not Gaussian, as indicated by the median degree of 1 and the heavy-tailed nature of the degree distributions. The metrics demonstrate that the giant component retains the essential characteristics of the global graph, with similar average and median degrees, but with a slightly higher density and fewer communities. This analysis reinforces the importance of the giant component in understanding the overall network structure.

Moreover, the giant component likely plays a crucial role in knowledge representation, accessibility, and discovery within the knowledge graph. In this context, where the graph connects scientific concepts (nodes) via relationships (edges), the giant component ensures that a large proportion of these concepts are interconnected. This extensive connectivity enhances the network's ability to represent comprehensive scientific knowledge, facilitates efficient access to related concepts, and supports the discovery of new relationships and insights across a wide array of scientific domains for which previously no connections were identified. In the next sections, various examples of such explorations will be presented.

\begin{table}[h!]
\small
\centering
\caption{Comparison of the properties of the global graph and its giant component. The results show that the degree distributions for the global graph and its giant component are likely not Gaussian, in agreement with the other analysis. }
\begin{tabular}{|l|c|c|}
\hline%\toprule
\textbf{Property} & \textbf{Global Graph} & \textbf{Giant Component} \\
\hline %\midrule
Number of nodes & 12319 & 11878 \\ \hline
Number of edges & 15752 & 15396 \\ \hline
Average node degree & 2.56 & 2.59 \\ \hline
Maximum node degree & 171 & 171 \\ \hline
Minimum node degree & 1 & 1 \\ \hline
Median node degree & 1 & 1 \\ \hline
Density & 0.00021 & 0.00022 \\ \hline
Number of communities & 109 & 80 \\ \hline

\hline%\bottomrule
\end{tabular}
\label{tab:table_graphproperties_comb}
\end{table}

\begin{figure}
    \centering
    \includegraphics[width=1\linewidth]{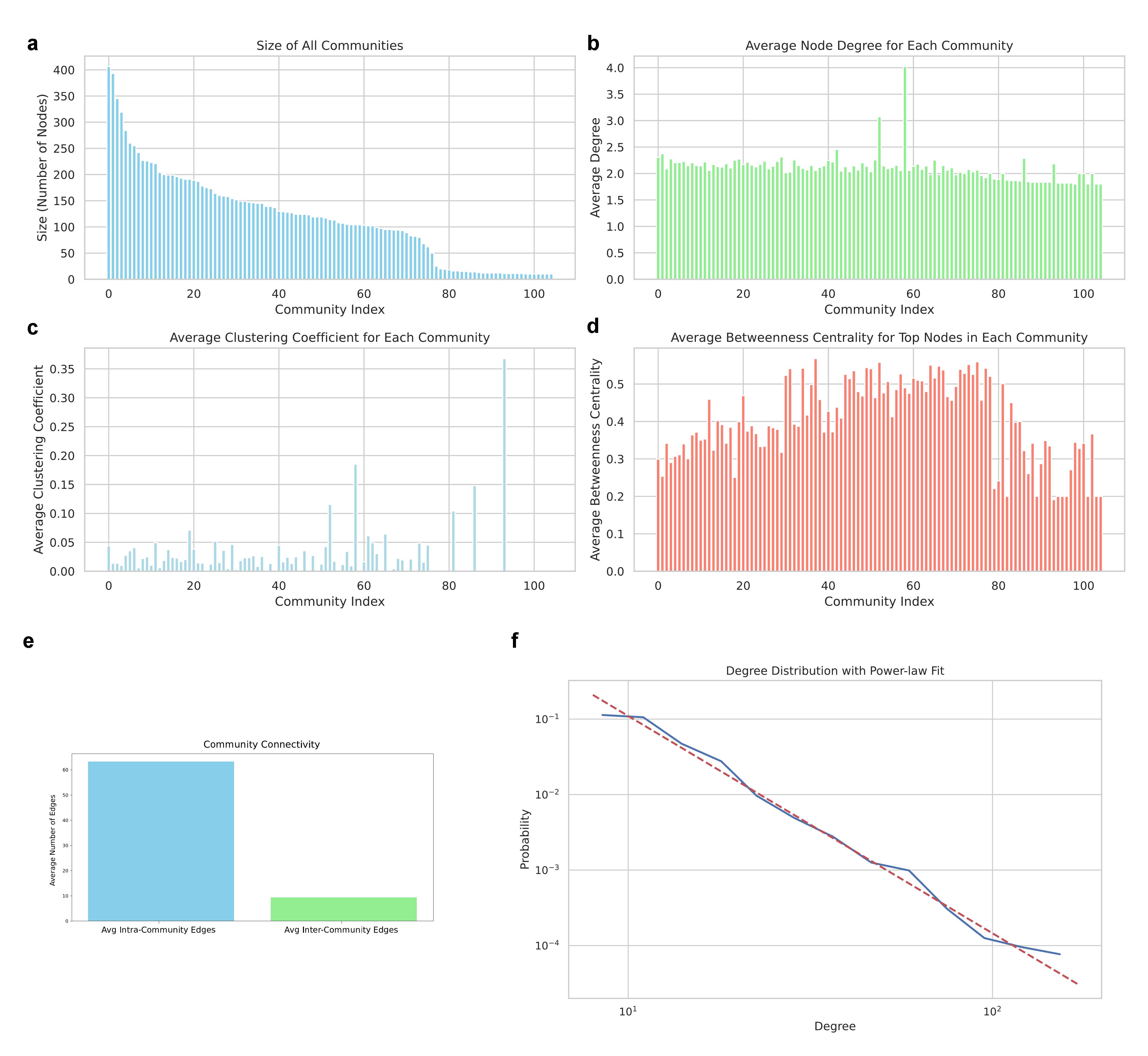}
    \caption{Comprehensive analysis of the structural properties of communities within a network, showing size of all communities (a), average node degree for each community (b), the average clustering coefficient for each community (c), and  the average betweenness centrality of the nodes in each community (d). Panel b illustrates the average node degree per community, demonstrating generally consistent internal connectivity with notable outliers, indicative of more densely interconnected communities. Panel c explores the average clustering coefficient, revealing that while most communities do not show a propensity for tight clustering, a select few deviate with higher coefficients, suggesting localized pockets of closely-knit nodes. Panel d examines the average betweenness centrality for the most influential nodes in each community, displaying a rather even distribution across the network with slight variations, implying a distributed rather than centralized control over the network's connectivity. These metrics provide insight into the network's topology, highlighting the balance between uniformly distributed influence and the existence of specialized clusters within the network's architecture. Panel e depicts an analysis of community structure in the graph, showing the average number of edges within communities to assess  how many edges are there on average that connect nodes within the same community (left). The data on the right depicts the average inter-community Edges designating the average number of edges that connect nodes from different communities. This data underscores the finding that this network seems to exhibit strong community structure, with more connections within communities than between them. Panel f shows the degree distribution of the global network on a log-log scale, with the empirical data in blue and the best-fit power-law model in a dashed red line. The power-law fit appears to follow the distribution of the data reasonably well, especially in the tail (high-degree region).}
    \label{fig:Fig_4}
\end{figure}

Samples of a few closest nodes to the centroid in each of the clusters are summarized in Table~\ref{tab:PCA_nodes_centroid}. The analysis shows that the embedding model (details on the embedding model used, see Section~\ref{embedding_section}) successfully captures related terms in similar regions, providing confidence that we can use this model to identify relationships between search terms and node features.

Figure~\ref{fig:Fig_4} shows an analysis of the features of the communities in the graph. First, Figure~\ref{fig:Fig_4}a shows the size of all communities, where we find that the community sizes follow a right-skewed distribution, with a few communities being significantly larger than the rest. This might indicate a scale-free or hierarchical structure where a few communities dominate by size. 
For a more detailed analysis, Figure~\ref{fig:Fig_4}b shows the average node degree for each community, revealing that most communities have a relatively stable average degree, which suggests that within communities, nodes have a similar number of connections. It is evident that there are a few outliers with significantly higher average degrees, indicating that certain communities may be more densely connected internally.
Figure~\ref{fig:Fig_4}c depicts the average clustering coefficient for each community. The data shows the clustering coefficient, a measure of the degree to which nodes in a graph tend to cluster together. The analysis shows that most communities have a low average clustering coefficient, with some exceptions. This indicates that, on average, there is not a strong tendency for nodes to form tightly knit groups within most communities, but there are a few communities that are exceptions to this trend.
Figure~\ref{fig:Fig_4}d shows the average betweenness centrality of the nodes in each community. Betweenness centrality measures the extent to which a node lies on paths between other nodes. The plot reveals a relatively uniform distribution of betweenness centrality among the top nodes across communities, with some variance. This suggests that in each community, there are a few nodes that play a significant role in connecting members of the community to the rest of the network. However, no single community stands out as having an exceptionally high betweenness centrality, which might have indicated a critical or controlling community in terms of network flow. This underscores the overall connectedness of the concepts captured in the corpus of knowledge. 

In Figure~\ref{fig:Fig_4}d, we observe a clear trend of increasing average betweenness centrality from community index 0 to approximately 77, corresponding to a decrease in community size as shown in Figure~\ref{fig:Fig_4}a. This trend suggests that in smaller communities, individual nodes play a more critical role in maintaining network connectivity, as fewer nodes are available to act as bridges. Consequently, these nodes have higher average betweenness centrality, indicating their importance in facilitating communication across the network. Beyond community index 77, there is a significant drop in average betweenness centrality, attributed to the simplified and locally connected structure of these very small communities. In such communities, the high local connectivity and redundancy of connections reduce the need for specific nodes to serve as intermediaries. This analysis enhances our understanding of the structural properties of the knowledge graph, highlighting the varying roles of nodes across different community sizes. Future research could explore the dynamic evolution of betweenness centrality and community structure within the knowledge graph as it grows. Investigating the effects of removing high betweenness centrality nodes in smaller communities could provide insights into the graph's vulnerability and robustness, potentially guiding strategies for enhancing resilience. Additionally, enhancing connectivity in low betweenness centrality communities could be beneficial, fostering more cohesive and resilient structures. Applying these analyses to different domains within the knowledge graph could help validate the findings and uncover domain-specific principles of connectivity and resilience, ultimately improving the utility and robustness of knowledge graphs in various applications.

\begin{figure}
    \centering
    \includegraphics[width=.75\linewidth]{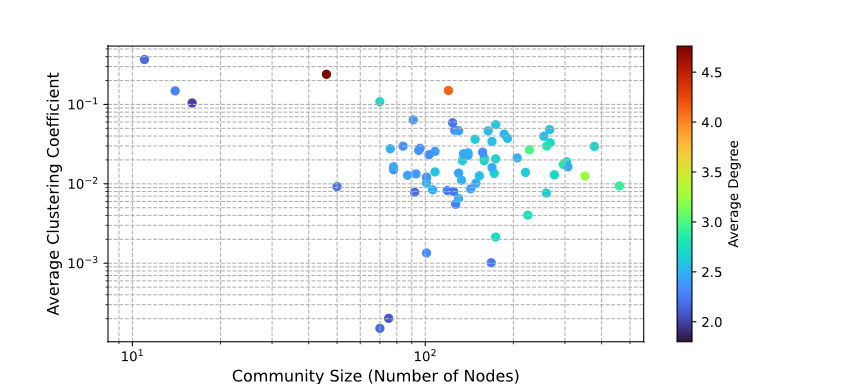}
    \caption{This plot illustrates the relationship between community size (number of nodes) and average clustering coefficient for different communities within the knowledge graph. Each point represents a community, with the color indicating the average degree of the nodes in that community, ranging from blue (lower average degree) to red (higher average degree). The $x$-axis and $y$-axis are on a logarithmic scale to capture the distribution across several orders of magnitude.} 
    \label{fig:Fig_avgclustering_over_commsize}
\end{figure}

Figure~\ref{fig:Fig_avgclustering_over_commsize} illustrates the relationship between community size (number of nodes) and average clustering coefficient for various communities within the knowledge graph. Each point represents a community, with the color indicating the average degree of nodes in that community, ranging from blue (lower average degree) to red (higher average degree). The plot employs logarithmic scales for both axes to capture the wide range of values and their distributions effectively.

The average clustering coefficient offers insights into the local connectivity and cohesiveness within each community. A high clustering coefficient indicates that nodes within the community are more likely to form triangles; that is, if node A is connected to nodes B and C, then nodes B and C are also likely to be connected. This results in tightly-knit clusters where neighbors of a node are also neighbors of each other, reflecting dense subgraphs with high interconnectivity. Such communities often represent closely related concepts or entities with frequent interactions.

Conversely, a low clustering coefficient signifies that nodes are less likely to form triangles, leading to sparser and less cohesive structures. In these communities, if node A is connected to nodes B and C, it is less likely that nodes B and C are connected to each other. This results in broader or more loosely connected communities with fewer interconnections. These communities might encompass a wider range of concepts or entities with less frequent interactions.

The color gradient in the figure highlights the variation in average node degree, adding another layer of insight into the community structure. High average degree communities, often central hubs, indicate regions with intense local interactions and robust connectivity, while lower average degree communities suggest more peripheral or isolated regions within the knowledge graph. 

High clustering coefficient communities, especially those with high average degrees, likely represent areas with intense research focus and collaboration. The high interconnectivity suggests a well-established body of knowledge with frequent interactions and cross-references among concepts. These areas might be central to the field and could be critical for driving further innovation and discovery. Low clustering coefficient communities indicate emerging or less established research areas. The lower interconnectivity suggests that these fields are still developing, with fewer established relationships between concepts. These areas could be ripe for new research opportunities and exploration, potentially leading to novel discoveries.

While more analysis is left to future work, such analyses can identify key areas for strategic investment, collaboration, and development within the knowledge graph. This analysis helps to understand the dynamics of knowledge formation and dissemination, guiding efforts to foster a more interconnected and resilient research landscape. 

A related analysis is depicted in Figure~\ref{fig:Fig_4}e, where we show an analysis of the modularity score and community connectivity. Modularity is a measure of the structure of networks or graphs which measures the strength of division of a network into communities. Networks with high modularity have dense connections between the nodes within modules but sparse connections between nodes in different modules. A modularity score can range from -0.5 to 1, where values close to 1 typically indicate strong community structure. The plot suggests that the modularity score of the analyzed network is quite high, around 0.9. As can be seen in Figure~\ref{fig:Fig_4}e, bottom, the average number of intra-community edges is significantly higher than the inter-community edges, which is consistent with a network that has a high modularity score (Figure~\ref{fig:Fig_4}e, top). Hence, this network seems to exhibit strong community structure, with more connections within communities than between them. This affirms our finding that this network features well-defined clusters or groups.

\begin{figure}
    \centering
    \includegraphics[width=0.7\linewidth]{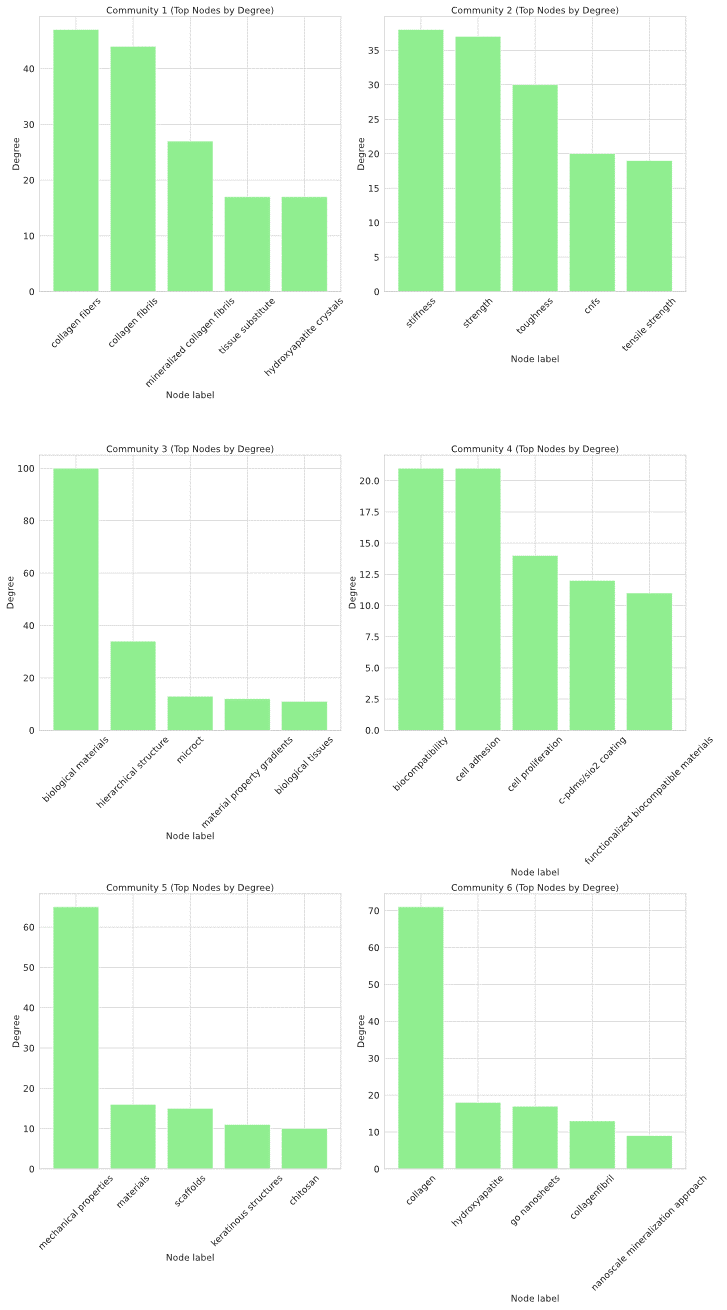}
    \caption{Degree distributions of the top nodes within six different communities. Each subplot corresponds to one community and displays the degrees of what appears to be the five nodes with the highest degree within that community. The communities appear to be structured around key thematic nodes that have significantly higher degrees, suggesting they may serve as central hubs within their respective communities. These hubs could be focal points for the flow of information or interactions within the network, implying that they are important in the overall connectivity of the network. The degree of these hubs could also indicate their importance in the network's functionality, particularly if the network represents biological materials where certain properties or elements are critical. There are several abbreviations commonly used in materials science, such as: Carbon nanofibers (CNFs), labeled as cnfs; MicroCT, labeled as microct; carbon-polydimethylsiloxane (C-PDMS), labeled as c-pdms. There are also other key concepts, such as `material property gradients' that is a commonly found design motif in biological materials. `SiO2 coating' is a silicon dioxide coating rather commonly used in materials engineering.}
    \label{fig:Fig_5}
\end{figure}

Figure~\ref{fig:Fig_5} depicts the degree distributions of the top nodes within six different communities. Each subplot corresponds to one community and displays the degrees of what appears to be the five nodes with the highest degree within that community. We present a brief analysis of each of the communities. 
In Community 1, we see that the degrees of the top nodes vary significantly, with `collagen fibers' having the highest degree, indicating that it is likely a central or hub node within this community. The presence of such a hub could suggest that this community is organized around a few key concepts or elements that are highly interconnected. This agrees with the central role collagen fibers play in defining structural, mechanically relevant biomaterials, describing a key design feature of biological systems wherein collagen is the most abundant structural protein in Nature~\cite{Buehler2006NatureFibrils}. The emergence of `hydroxyapatite crystals' makes sense since these material components of bone form near and within collagen materials, and hence form an integral part of the biological hierarchical structure.
In Community 2 we also find a hub-like node labeled `strength', suggesting its pivotal role within the community. The other top nodes, such as `stiffness' and `toughness,' have relatively high degrees as well, pointing to a community likely focused on mechanical properties of materials, where these properties are critical.
In Community 3 we see that one node, `biological materials', dominates this community with a degree much higher than the others. This kind of dominance indicates a highly central concept that could be pivotal in the structure and dynamics of this community, possibly acting as a key connector to other parts of the network. The second highest ranked node is 'hierarchical structure', reflecting the most dominating design principle in biological material composition. 
In Community 4 we can see that the degree distribution among the top nodes is more balanced compared to the previous communities, even though `biocompatibility' stands out. This suggests a more evenly distributed network structure without a single overwhelming hub. In that community, other top nodes are 'cell adhesion' and 'cell proliferation', associating closely with important biological mechanisms by which tissues grow and remodel. 
In Community 5, similar to the structure seen in community 3, `mechanical properties' has a significantly higher degree than other nodes, indicating its central importance within this  community. Several other important nodes are listed, such as 'materials' and 'scaffolds', playing a central role within this domain. 
Finally, in Community 6 we again see one node, 'collagen', with a degree much higher than the others, signifying its central role within the community's network structure. This node is followed by a variety of related concepts, notably 'hydroxyapatite' but also engineered material components like 'graphene nanosheets'. 
Our understanding of this particular area of knowledge confirms that these are indeed key features that define the field. 
These results suggest that we have a network with a heterogeneous distribution of node centrality, with certain nodes playing disproportionately significant roles within their communities. This could be indicative of a scale-free network characteristic within individual communities, or even the entire graph.  We will now explore this feature in more detail.  

Average betweenness centrality in the context of network analysis is a measure that quantifies the average extent to which nodes stand between each other on their shortest paths through the network. Betweenness centrality itself is a way of identifying the importance of a node within the network, based on the number of shortest paths that pass through the node. Nodes with high betweenness centrality are often considered key connectors or bridges within the network, facilitating or controlling the flow of information (or any other entity being modeled) between different parts of the network.
The average betweenness centrality of a set of nodes (for example, within a community in a network) is calculated by taking the mean of the betweenness centrality scores of all the nodes in that set. This average gives an indication of how influential the typical node in the set is, in terms of connecting different parts of the network. In the context of community analysis, a high average betweenness centrality for a community might suggest that the community contains several nodes that play critical roles in connecting the community to other parts of the network. Conversely, a low average might indicate that the community is more insular, with fewer connections to other communities or parts of the network.
Hence, the measure of average betweenness centrality provides a measure of the overall importance of a group of nodes (such as a community) in facilitating connectivity and flow within the larger network structure.

For a deeper analysis of the scaling behavior of the graph, Figure~\ref{fig:Fig_4}f shows the degree distribution of a network on a log-log scale, with the empirical data in blue and the best-fit power-law model in a dashed red line. The power-law fit appears to follow the distribution of the data reasonably well, especially in the tail (high-degree region).
The power-law Exponent $\alpha=2.8786$, which falls within the range typically observed in scale-free networks ( $1<\alpha<3$). This suggests that the network may exhibit scale-free properties.
Further, the standard error of $\alpha=0.0698$ is relatively small, indicating a high level of precision in the estimation of the power-law exponent.
The log-likelihood ratio $R=4.1526$, which is a measure of how much better the power-law model fits the data compared to the exponential model. A positive $R$ value indicates that the power-law model is indeed a better fit.
The p-value is very small, about 3.29E-05, which statistically significantly suggests that the power-law model is a better fit than the exponential model. 
Table~\ref{tab:fit_statistics} summarizes these results. 
From the fit and these statistical test results, we have strong evidence to support the claim that the network is scale-free. Specifically, the value of $\alpha$ within the expected range for scale-free networks, the statistically significant log-likelihood ratio favoring the power-law model over the exponential model, and the visual agreement between the empirical data and the power-law model all support this conclusion.

For background, a scale-free network is a type of network characterized by a few highly connected nodes, known as hubs, and many nodes with fewer connections, forming a structure found in various natural and man-made systems, such as social networks, the internet, and biological networks. In these networks, most nodes have only a few connections, while a few nodes (hubs) have a very high number of connections, similar to an airline route map where major cities are hubs with many flights, and smaller cities have fewer connections. The number of connections each node has follows a power law distribution, meaning there are many nodes with few connections and a few nodes with many connections. Scale-free networks are robust against random failures, as the hubs keep the network connected, but they are vulnerable to targeted attacks on the hubs, which can significantly disrupt the network. Examples of scale-free networks include social networks, where a few people have many connections and many have few, the World Wide Web, where some websites have many links pointing to them, and biological networks, such as metabolic networks in cells.

\begin{table}[h]
\caption{Result of the fit statistics comparing power-law to exponential distribution for network degree distribution.}
\small 
\begin{center}
 
\begin{tabular}{|l|c|}
\hline
\textbf{Statistic} & \textbf{Value} \\
\hline
Power-law exponent ($\alpha$) & 2.8786 \\
\hline
Standard error of $\alpha$ & 0.0698 \\
\hline
Log-likelihood ratio (R) & 4.1526 \\
\hline
p-value for the comparison & $3.29 \times 10^{-5}$ \\
\hline
\hline
\end{tabular}
\end{center}
\label{tab:fit_statistics}
\end{table}

Scale-free networks are characterized by their degree distribution following a power law, at least asymptotically. This means that a few nodes in the network have a very high degree (a large number of connections to other nodes), while most nodes have a relatively low degree. This distribution results in a network that is highly resistant to random failures but vulnerable to targeted attacks on its most connected nodes. This property facilitates the effective use of these graphs for knowledge extraction. For instance, the presence of hub nodes makes it easier to navigate the graph efficiently. Since hubs are connected to many other nodes, algorithms as those that will be developed as part of this paper can leverage these hubs to reduce the path length between distant nodes, enhancing the efficiency of search and information retrieval processes. It also allows us to explore a number of alternative paths that connect nodes, leading to powerful discovery mechanisms to explore, and reason over, various concepts and potential relationships.

\subsection{Extraction of multiple graph traversal paths via ranked combinatorial analysis of cosine similarities}

Tracing linkages refers to the process of systematically following and identifying connections or relationships between various entities or concepts within graph. In the context of knowledge graphs, tracing linkages involves navigating through the nodes and edges of the graph to understand how different elements are interconnected (resulting in one or more possible paths). This exploration helps reveal patterns, dependencies, and associations that may not be immediately apparent, thereby facilitating a deeper understanding of the underlying relationships and facilitating tasks such as knowledge extraction, inference, and reasoning. We can  apply this concept to extract multiple paths between dissimilar concepts by identifying two (or more) search terms that define the  beginning and end, and we then identify the shortest path between them.
We use node embeddings as an effective way to represent the graph structure, enabling the application of machine learning algorithms on graphs. These embeddings are dense vector representations of the content of nodes in a graph, capturing the essence of a node's features. As was shown in the previous section, embeddings of nodes with similar structural roles or within the same neighborhood are closer in the vector space. 
We identify top-ranked nodes based on their similarity to a given node or a set of nodes through cosine similarity. Cosine similarity quantifies the cosine of the angle between two vectors in an $n$-dimensional space, which in this context, are the embeddings of two nodes. It is defined as:
\begin{equation}
    \text{cosine similarity}(\mathbf{a}, \mathbf{b}) = \frac{\mathbf{a} \cdot \mathbf{b}}{\|\mathbf{a}\| \|\mathbf{b}\|}
\end{equation}
where $\mathbf{a}$ and $\mathbf{b}$ are the embedding vectors of two nodes, $\cdot$ denotes the dot product, and $\|\mathbf{a}\|$ and $\|\mathbf{b}\|$ are the Euclidean norms of the vectors.
The similarity score ranges from -1 to 1, where 1 indicates identical directionality (high similarity), 0 indicates orthogonality (no similarity), and -1 indicates opposite directionality (high dissimilarity). In the context of node embeddings, a higher cosine similarity score between two nodes suggests that they are more similar in terms of their structural roles or positions in the graph. To identify top-ranked nodes relative to a specific node, the cosine similarity between the target node's embedding and every other node's embedding is computed. Nodes are then ranked based on their similarity scores, with higher scores indicating a closer or more relevant relationship to the target node.  

We note that cosine similarity is generally the best choice for measuring the similarity of embeddings because it focuses on the direction of the vectors rather than their magnitude. This makes it ideal for high-dimensional data like word or sentence embeddings. Cosine similarity is scale-invariant, meaning it normalizes the vectors, which is important for comparing embeddings of different magnitudes. Unlike Hamming distance (which is suitable for binary strings) and Manhattan or Euclidean distances (which consider magnitude), cosine similarity captures the semantic similarity effectively, making it widely used in NLP and ML applications.

The resulting set of nodes is now used to construct a new sub-graph, for instance based on identifying the shortest path between them. We can use different approaches towards the construction of the new sub-graph by including higher-order neighbors, for instance two hops.  We can also identify a ranking of the cosine similarity results to assess an ordered set of nodes that best fit our search term, resulting in multiple combinatorial options to form several paths. Finally, it is noted that the search term does not have to be limited to describe short terms, it can consist of an abstract or longer documents (this is possible since we create an embedding vector that is independent of the length of the source; more details, see Section~\ref{embedding_section}). 

As an example, we use the terms of `graphene' and `silk' and determine the path to connect these concepts. The nodes are `graphene', `strength', `biological materials', `silk'. Since each of the edges in our graph has labels that delineates the relationship between concepts, we can also identify those in the integrated delineation the path between these two concepts. The result for the example above is:
\begin{figure}[H]
\begin{LLMbox}{}
\begin{lstlisting}
(*@\textcolor{red}{graphene}@*) --> improves --> strength --> is exhibited due to hierarchical microstructures that allow for damage tolerance at multiple length scales --> biological materials --> provide functionalities --> (*@\textcolor{blue}{silk}@*)
\end{lstlisting}
\end{LLMbox}
\end{figure}
In another example, we use two terms `inkjet printer for living tissues' and `spider silk proteome'. These terms do not exist verbatim as node labels; however, nodes labeled `inkjet-based bioprinting' and `spider silk protein'  are identified as closest match (with 0.89 and 0.91 cosine similarities, respectively). 
The resulting path is then:
\begin{figure}[H]
\begin{LLMbox}{}
\begin{lstlisting}
(*@\textcolor{red}{inkjet-based~bioprinting}@*) --> lacking --> structural integrity --> dictates through surrounding permeable shell structure --> cortex --> has tight bonding due to ridges and smooth transition --> feather\_rachis --> has due to synergetic effect between cortex and foam components --> enhanced mechanical properties --> realized in spinning and nanoindentation experiments --> functional silk fibers --> constituent proteins of --> silk proteins --> chemical composition of --> (*@\textcolor{blue}{spider~silk~protein}@*)
\end{lstlisting}
\end{LLMbox}
\end{figure}
Additional context for these analyses can be gleaned by considering connected nodes, and a deeper investigation of the sub-graph structure.  We note that generally, the use of embeddings allows us to incorporate longer text (depending on the embedding model used, in the case of the model used here we can provide text chunks of up to 512 tokens to match a node). 

These examples also illustrate the mechanics by which graph represents provide a solid foundation to discovery connections. For instance, one scientific paper may discuss a sorts of bioprinting methods, whereas another paper may discuss silk proteins. A connection between these two concepts would not emerge from either one of these papers alone. However, when ontological knowledge graphs are constructed from both, we find overlapping concepts that provide a bridge between the technical content in both papers. 

\subsection{Reasoning over the graph: Graph traversal based question answering}
\label{reasoningovergraph}

Graph traversal by tracing linkages as discussed in the preceding section allows us to find connections between concepts that have not been established before (or that have not understood to be related in any known manner) and while these paths are valuable on their own (e.g., for human interpretation), sophisticated reasoning with special-purpose generative AI models can provide deep insights and even facilitate discovery of new ideas, connections and relationships. This is because these models will overlay the extracted graph structures with their own understanding, which triggers models to expand their `thinking' and specifically extrapolating towards ideas that have not been associated or known before. What is most important is that this strategy effectively stimulates the capacity of AI models, specifically multimodal LLMs, to move beyond knowledge retrieval and towards generation of new connections of information, and hence, new knowledge (see, Figure~\ref{fig:Fig_1_intro}a).   

\begin{figure}
    \centering
    \includegraphics[width=1\linewidth]{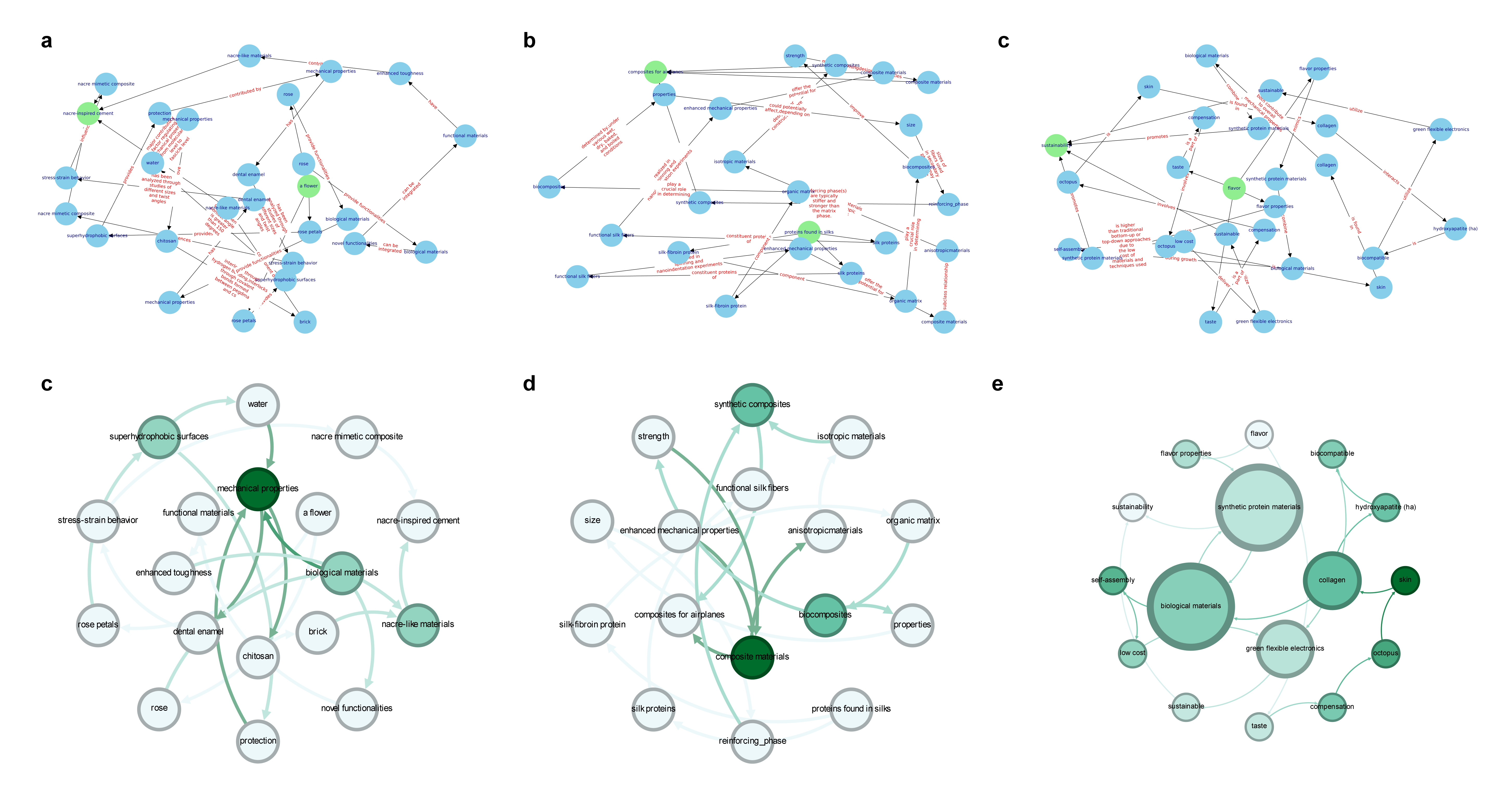}
    \caption{Knowledge graphs derived from multi-path sampling, for two distinct cases (panel a: `a flower' connecting `nacre-inspired cement', panel b: `proteins found in silks' connecting 'composites for airplanes', and panel c: `sustainability' and `flavor'). These graphs represent relationships among various concepts. In these graphs, each of the paths is rendered separately (and hence, labels can occur multiple times). In panels d-f the same original paths are analyzed, but the graphs are constructed by merging all nodes, even if they come from different paths. This results in a significantly distinct topology of the resulting graphs with interesting new connections formed. Nodes with high node degree are plotted in dark green, and with low node degree, in light green. In panel d, the node with the highest node degree is 'mechanical properties' and in panel e, `composite materials'. 
    It is noted that panels c and f show graphs derived from multi-path sampling from `flavor' to `sustainability', based on the augmented graph that includes data from adversarial X-LoRA sampling. These graphs represent relationships among various concepts, whereas in panel c each of the paths is rendered separately and in panel f the graph is constructed by merging all nodes, even if they come from different paths.  }
    \label{fig:Fig_7}
\end{figure}

We postulate that exploring complex graph traversals offers fertile grounds for analysis. We can either identify a single most likely path or sample multiple paths. For instance, Figure~\ref{fig:Fig_7}a-c depicts two graphs generated by developing all possible connections between the top two nodes (this results in four paths, featuring the top path 0-0, 0-1, 1-0 and 1-1). These examples explore new subgraphs extracted to capture connections between disparate concepts such as between `a flower' and  `nacre-inspired cement' or `proteins found in silks' connecting `composites for airplanes'.
Building on this, in Figure~\ref{fig:Fig_7}d-f the same original paths are analyzed, but the graphs are constructed by merging all nodes, even if they come from different paths. This results in a significantly distinct topology of the resulting graph structure and offers novel connectivities between concepts and their relationships as encoded in the edges. This points to great flexibility by which graphs can be formulated and ultimately be used, for reasoning applications. 
 
The ontological knowledge graphs can be utilized to support reasoning in scientific research, such as proposing hypotheses about material properties or predicting the likely outcomes of combining different materials. From another perspective, they can help identify gaps in knowledge, suggest new areas for research, and facilitate understanding of complex interrelations in materials science or across different disciplines. In the first experiment we pose the query shown in Text Box~\ref{TexBox:TB1}. We discuss the results from several LLMs, to offer a comparative analysis between different types of models. 

\begin{figure}[htbp]
%#\begin{mdframed}
\begin{LLMbox}{} %BioMixtral XLora 
%\begin{LLMbox}{\XLora} 
\begin{lstlisting}
You are given a set of information from a graph that describes the relationship between materials, structure, properties, and properties. You analyze these logically through reasoning.

### Primary combination (path from 0 to 0):

(*@\textcolor{red}{a flower}@*) --> rose petals --> Provides --> superhydrophobic surfaces --> Provides --> protection --> contributed by --> mechanical properties --> has --> dental enamel --> has been analyzed through studies of different sizes and twist angles --> stress-strain behavior --> influences --> nacre mimetic composite --> (*@\textcolor{blue}{nacre-inspired~cement}@*)

This represents the main combination of nodes in the knowledge graph between a flower and nacre-inspired cement.

The following represent another possible combination of paths, providing different insights or complementing the primary path.

### Alternative combination (path from 0 to 1):

a flower --> rose petals --> Provides --> superhydrophobic surfaces --> occurs when contact angle is greater than 150 degrees --> water --> Major contributory factor regulating mechanical properties from molecular level to fascicle level --> mechanical properties --> can improve --> chitosan --> interlocks through hydrogen bonding,interlocks through covalent bonds formed between PEGDMA and CS --> brick --> component of --> nacre-like materials --> nacre-inspired cement

### Alternative combination (path from 1 to 0):

a flower --> rose --> provide functionalities --> biological materials --> provide functionalities --> mechanical properties --> has --> dental enamel --> has been analyzed through studies of different sizes and twist angles --> stress-strain behavior --> influences --> nacre mimetic composite --> nacre-inspired cement

### Alternative combination (path from 1 to 1):

a flower --> rose --> provide functionalities --> biological materials --> can be integrated --> novel functionalities --> can be integrated --> functional materials --> have --> enhanced toughness --> Contributed by --> nacre-like materials --> nacre-inspired cement

(*@\hl{\#\#\# Carefully read the paths and summarize scientific insights in several bullet points. Then be creative and propose new research ideas. Think step by step.}@*)
\end{lstlisting}
\end{LLMbox}
%\end{mdframed}
\textboxcaption{Example query developed from a path analysis of knowledge graphs. In the example a combinatorial set of node ranking up to two is used, leading to four paths (0-0, 0-1, 1-0, 1-1). We color the beginning and end concepts in the primary path in red and blue font, respectively, and also highlight the \hl{key instruction}. Becuase LLMs can ingest complex multimodal information, we can provide the graphs themselves, additional context, and instructions through prompting.}
\label{TexBox:TB1}
\end{figure}

As can be seen, these paths provide for a rich representation of relationships between seemingly unrelated concepts. We now explore responses to the same prompt from three different LLMs, each of them aiming to find relationships between `a flower' and  `nacre-inspired cement'. 
First, the response of X-LoRA~\cite{Buehler2024X-LoRA:Design} is shown in Text Box~\ref{TexBox:TB2}. Here, a particularly interesting concept proposed is the study of the interlocking mechanisms of chitosan (a natural polymer derived from chitin found in the exoskeleton of crustaceans) and water through hydrogen bonding and covalent bonds. 
Next, the response from BioinspiredLLM-Mixtral is shown in Text Box~\ref{TexBox:TB3}. Again we note a suggestion to explore the use of hydrogen bonding or covalent bonding, this time in conjunction with polyethylene glycol dimethacrylate (PEGDMA). PEGDMA is widely used as a crosslinking agent in the creation of polymers, particularly hydrogels, due to its ability to form networks through polymerization, and could indeed be a useful method to improve nacre-based materials. 
Additionally it is suggested that research could investigate the stress-strain behavior of dental enamel at different sizes and twist angles to investigate how this behavior influences the mechanical properties of nacre-inspired cement. This is an interesting idea that builds on research reported in recent papers~\cite{Lew2023ASeashells} but applied here to a new class of nacre-inspired materials that incorporate mixtures of PEGDMA and chitosan. 
Finally, the response from GPT-4 is shown in Text Box~\ref{TexBox:TB4}. This response includes several deep insights, such as the use of superhydrophobic surfaces, like those of rose petals, in influencing mechanical properties. As in the previous responses, the model suggests that chitosan may be interlocked through hydrogen bonding and covalent bonds formed between PEGDMA and other additives. The study of dental enamel is also suggested in several of the responses, which suggests that important cues can be taken from this material that is the hardest and most mineralized substance in the human body. The use of self-cleaning surfaces or materials that can capture and convert CO2 is another interesting suggestion made.

\begin{figure}[htbp]
\begin{LLMbox}{\XLora} %BioMixtral XLora 
%\begin{mdframed}
\begin{lstlisting}
The primary combination highlights the connection between flowers, rose petals, superhydrophobic surfaces, protection, and dental enamel. This path suggests that the mechanical properties of dental enamel can be influenced by the stress-strain behavior of nacre-inspired cement.

The alternative combination involving water and chitosan suggests that the interlocking of these materials through hydrogen bonding and covalent bonds can improve the mechanical properties of nacre-like materials, which can be applied to nacre-inspired cement.

The path connecting roses, biological materials, and functional materials suggests that novel functionalities can be integrated into these materials, potentially enhancing their toughness.

The final alternative combination highlights the potential for integrating biological materials and functional materials to create nacre-inspired cement with enhanced properties.

Research ideas:

- Investigate the relationship between the stress-strain behavior of nacre-inspired cement and the mechanical properties of dental enamel.

- Study the interlocking mechanisms of chitosan and water to improve the mechanical properties of nacre-like materials.

- Explore the integration of novel functionalities into biological materials and functional materials to enhance the toughness of nacre-inspired cement.

- Develop new methods for integrating biological materials and functional materials into nacre-inspired cement to create materials with improved properties.
\end{lstlisting}
\end{LLMbox}
\textboxcaption{Response from X-LoRA~\cite{Buehler2024X-LoRA:Design} to the prompt delineated in Text Box~\ref{TexBox:TB1}, seeking to uncover relationships between 'a flower' and  'nacre-inspired cement'. }
\label{TexBox:TB2}
\end{figure}

\begin{figure}[htbp]
%\begin{mdframed}
%
\begin{LLMbox}{\BioMixtral} %BioMixtral XLora 
\begin{lstlisting}
The primary combination of nodes in the knowledge graph represents the main connection between a flower and nacre-inspired cement. The alternative combinations provide additional insights or complement the primary path by highlighting different aspects of the relationship between materials, structure, properties, and properties. These combinations can be summarized as follows:

-The primary path highlights the connection between a flower's superhydrophobic surfaces and nacre-inspired cement's protection provided by mechanical properties. This path suggests that the mechanical properties of nacre-inspired cement can be influenced by the stress-strain behavior of dental enamel, which has been analyzed through studies of different sizes and twist angles.

-The alternative path from 0 to 1 emphasizes the major contributory factor regulating mechanical properties from the molecular level to the fascicle level, which is water. This path suggests that the mechanical properties of chitosan can be improved by interlocks formed through hydrogen bonding and covalent bonds with PEGDMA, leading to enhanced toughness in nacre-like materials and nacre-inspired cement.

-The alternative path from 1 to 0 highlights the functionalities provided by biological materials, which can be integrated into novel functionalities in functional materials. This path suggests that nacre-like materials can contribute to enhanced toughness in nacre-inspired cement.

-The alternative path from 1 to 1 emphasizes the connection between functionalities provided by biological materials and novel functionalities in functional materials. This path suggests that nacre-inspired cement can be designed with enhanced toughness by integrating nacre-like materials.

Based on these insights, new research ideas could include exploring different methods for integrating nacre-like materials into functional materials to enhance their mechanical properties, such as through hydrogen bonding or covalent bonding with PEGDMA. Additionally, further studies could investigate the stress-strain behavior of dental enamel at different sizes and twist angles to better understand how this behavior influences the mechanical properties of nacre-inspired cement. Finally, exploring the functionalities provided by biological materials in novel functionalities in functional materials could lead to new applications for nacre-inspired cement in various industries.
\end{lstlisting}
\end{LLMbox}
\textboxcaption{Response from BioinspiredLLM-Mixtral to the prompt delineated in Text Box~\ref{TexBox:TB1}, seeking to uncover relationships between 'a flower' and  'nacre-inspired cement'.}
\label{TexBox:TB3}
\end{figure}

\begin{figure}[htbp]
%\begin{mdframed}
%\begin{lstlisting}
\begin{LLMbox}{\VG} %BioMixtral XLora 
\begin{lstlisting}
Analyzing the provided paths from the knowledge graph, we can extract several scientific insights:

Superhydrophobicity and Mechanical Properties: The primary path and the first alternative combination highlight the role of superhydrophobic surfaces, like those of rose petals, in influencing mechanical properties. 

Superhydrophobicity, characterized by a water contact angle greater than 150 degrees, is a key factor in enhancing the mechanical properties of materials from the molecular to the fascicle level.

Biomimicry in Material Science: All paths emphasize the importance of biomimicry, where biological materials (e.g., rose petals, dental enamel) inspire the design of synthetic materials with superior properties, such as nacre-inspired cement. This approach is pivotal in developing materials with enhanced toughness and mechanical properties.

Role of Chemical Bonding in Material Properties: The alternative path from 0 to 1 introduces the concept of chemical bonding (hydrogen bonding and covalent bonds) in improving the mechanical properties of materials. Specifically, it mentions how chitosan interlocks through hydrogen bonding and covalent bonds formed between PEGDMA and CS contribute to the formation of nacre-like materials.

Stress-Strain Behavior in Material Analysis: The primary path and one alternative path discuss the importance of studying stress-strain behavior in understanding the mechanical properties of materials. This is exemplified by the analysis of dental enamel, which provides insights into the design of nacre mimetic composites.

Integration of Novel Functionalities: The last alternative path suggests that the integration of novel functionalities into biological materials can lead to the development of functional materials with enhanced toughness, contributing to the creation of nacre-like materials.

Based on these insights, several new research ideas can be proposed:

-Development of Superhydrophobic Coatings Inspired by Rose Petals: Investigate the application of superhydrophobic coatings, inspired by the surface structure of rose petals, in various industries (e.g., automotive, aerospace) to enhance material longevity and resistance to environmental factors.

-Synthesis of Hybrid Materials Combining Biological and Synthetic Elements: Explore the synthesis of hybrid materials that combine biological elements (e.g., chitosan) with synthetic polymers through specific bonding mechanisms to create materials with tailored mechanical properties for use in medical implants or protective gear.

-Advanced Studies on Stress-Strain Behavior in Biomimetic Materials: Conduct in-depth studies on the stress-strain behavior of biomimetic materials, focusing on how different sizes and twist angles can influence the mechanical properties. This could lead to the development of materials with customizable flexibility and strength for construction or manufacturing applications.

-Integration of Functionalities for Environmental Sustainability: Research the integration of novel functionalities into materials that contribute to environmental sustainability, such as self-cleaning surfaces or materials that can capture and convert CO2. This could involve the development of nacre-inspired materials with added functionalities for use in green buildings or environmental remediation.

-Computational Modeling of Nacre-Inspired Materials: Utilize computational modeling to explore the structure-property relationships in nacre-inspired materials. This could help in identifying key factors that contribute to their enhanced toughness and guide the design of new materials with optimized properties for specific applications.

\end{lstlisting}
\end{LLMbox}
\textboxcaption{Response from GPT-4 to the prompt delineated in Text Box~\ref{TexBox:TB1}, seeking to uncover relationships between 'a flower' and  'nacre-inspired cement'.}
\label{TexBox:TB4}
\end{figure}

Amongst these three models, GPT-4 shows probably the most impressive reasoning capability and the most detailed response. 
For each of the interactions, follow-up queries can be used to go deeper into an analysis of the design features by exploring  how the various ideas can be integrated to develop new nacre-inspired cements. We do this by providing the previous responses, followed by this instruction:
\begin{figure}[H]
\begin{LLMbox}{}
\begin{lstlisting}
Identify specific a material design for nacre-inspired cements that combine these ideas. Focus on technical details of new concepts never explored before, and explain the reasoning behind them. 
\end{lstlisting}
\end{LLMbox}
\end{figure}
X-LoRA~\cite{Buehler2024X-LoRA:Design} responds as shown in Text Box~\ref{TexBox:TB10}. A key prediction is that the suggested composite material features chitosan and PEGDMA in a layered structure where the chitosan provides biological functionality and interlocks with PEGDMA through hydrogen bonding and covalent bonds formed between the methacrylate groups of PEGDMA, and the amine groups of chitosan. The model predicts that this would result in enhanced mechanical properties, such as toughness, due to the interlocking mechanisms. 

\begin{figure}[htbp]
\begin{LLMbox}{\XLora} %BioMixtral XLora 
\begin{lstlisting}
One possible design for nacre-inspired cement that combines these ideas is a composite material consisting of chitosan and PEGDMA (poly(ethylene glycol) dimethacrylate) in a layered structure. The chitosan provides biological functionality and interlocks with PEGDMA through hydrogen bonding and covalent bonds formed between the methacrylate groups of PEGDMA and the amine groups of chitosan. This results in enhanced mechanical properties, such as toughness, due to the interlocking mechanisms.

To further improve the mechanical properties, novel functionalities can be integrated into the chitosan layer, such as doping with silica nanoparticles or graphene oxide. These functionalities can enhance the strength and durability of the material, making it suitable for various applications, such as dental restoration or biomedical implants.

The reasoning behind this design is based on the understanding that the interlocking mechanisms of chitosan and PEGDMA can improve the mechanical properties of nacre-like materials, and that the integration of novel functionalities can enhance the toughness of the material. Additionally, the use of chitosan as a biological material provides potential biocompatibility and biodegradability advantages. The layered structure allows for the incorporation of different functionalities and materials, creating a versatile and customizable material for various applications.
\end{lstlisting}
\end{LLMbox}
\textboxcaption{Follow-up response from X-LoRA~\cite{Buehler2024X-LoRA:Design}, querying the model for specific a material design for nacre-inspired cements that combine the ideas identified in the earlier response shown in Text Box~\ref{TexBox:TB2}.}
\label{TexBox:TB10}
\end{figure}

The response from BioinspiredLLM-Mixtral is shown in Text Box~\ref{TexBox:TB11}. The responses covers a variety of aspects similar to the previous one, including the investigation of interlocking mechanisms of chitosan and water in nacre-like materials. The model also suggests exploring the relationships between flowers, rose petals, superhydrophobic surfaces, protection, dental enamel, chitosan, water, biological materials, functional materials, and nacre-inspired cement to develop new strategies for improving the mechanical properties. We notice that in the  response, the model does perhaps not clearly understand that nacre-inspired cement does not yet exist, and is supposed to be designed, as it asks to investigate the stress-strain behavior of nacre-inspired cement and its influence on the mechanical properties of dental enamel. Perhaps the causality should be reversed where we use dental enamel as a way to design the nacre-inspired cement materials (one could query the model again to that effect in follow-up interactions, to help clarify this issue).

\begin{figure}[htbp]
\begin{LLMbox}{\BioMixtral} %BioMixtral XLora 
\begin{lstlisting}
Based on the provided information, we can identify several research ideas for developing nacre-inspired cements with enhanced properties. These ideas involve exploring the relationships between flowers, rose petals, superhydrophobic surfaces, protection, dental enamel, chitosan, water, biological materials, functional materials, and nacre-inspired cement.

One research idea is to investigate the stress-strain behavior of nacre-inspired cement and its influence on the mechanical properties of dental enamel. This could involve studying the deformation mechanisms of dental enamel under different loading conditions and comparing them with the stress-strain behavior of nacre-inspired cement. By understanding how nacre-inspired cement affects dental enamel's mechanical properties, we can develop new strategies for improving dental restorations.

Another research idea is to study the interlocking mechanisms of chitosan and water in nacre-like materials. This could involve investigating how hydrogen bonding and covalent bonds formed between PEGDMA and CS contribute to the mechanical properties of nacre-like materials. By understanding how these interactions affect the mechanical properties of nacre-like materials, we can develop new strategies for improving the toughness of nacre-inspired cement.

A third research idea is to explore integrating novel functionalities into biological materials and functional materials to enhance the toughness of nacre-inspired cement. This could involve developing new methods for incorporating functional groups or nanoparticles into biological materials or functional materials, which can improve their mechanical properties. By understanding how these functionalities affect the mechanical properties of nacre-inspired cement, we can develop new strategies for creating materials with improved toughness.

A final research idea is to develop new methods for integrating biological materials and functional materials into nacre-inspired cement to create materials with improved properties. This could involve developing new techniques for combining biological materials with functional materials or exploring new ways to incorporate biological materials into nacre-inspired cement. By understanding how these combinations affect the mechanical properties of nacre-inspired cement, we can develop new strategies for creating materials with improved toughness.

Overall, these research ideas involve exploring the relationships between flowers, rose petals, superhydrophobic surfaces, protection, dental enamel, chitosan, water, biological materials, functional materials, and nacre-inspired cement to develop new strategies for improving the mechanical properties of nacre-inspired cement. By understanding how these relationships affect the mechanical properties of nacre-inspired cement, we can develop new strategies for creating materials with improved toughness.
\end{lstlisting}
\end{LLMbox}
\textboxcaption{Follow-up response  BioinspiredLLM-Mixtral, querying the model for specific a material design for nacre-inspired cements that combine the ideas identified in the earlier response shown in Text Box~\ref{TexBox:TB3}.}
\label{TexBox:TB11}
\end{figure}
 
As an alternative, the response from GPT-4 is shown in Text Box~\ref{TexBox:TB13}. The model produces a detailed conceptual materials design that can form a foundation for a manufacturing process in the lab. The model specifically suggests to use a mix of chitosan for organic layers and modified cement particles as the inorganic component. The model predicts that chitosan not only acts as a binder but, through its inherent properties, introduces hydrophilic sites, which is likely due to its cationic nature (the presence of free amino groups (-NH2) in the glucosamine units of chitosan contributes to its hydrophilicity, where these amino groups can be protonated in acidic solutions, forming –NH3+ groups, which enhance the material's solubility in water and its ability to interact with water molecules). PEGDMA is introduced not only to improve the mechanical integrity through covalent bonding, but also to capitalize on its water-resisting capabilities. Here, it is important to note that crosslinked networks formed from PEGDMA can exhibit varying degrees of water resistance, where the design leverages PEGDMA not for outright water repellence but for a controlled interaction with water such that while the material may absorb water. It does so in a way that doesn't compromise its structural integrity and the absorbed water might even participate in beneficial processes such as the curing of cement or contribute to its mechanical properties, akin to how natural nacre manages moisture. It is further suggested that the cement's surface should be engineered to mimic a particular type of microscale roughness by etching micro to nanoscale patterns onto the cement's surface layer, potentially through laser etching or additive manufacturing techniques. The etched surface is then coated with a thin layer of silicon-based compounds (e.g., polydimethylsiloxane, PDMS) to enhance water repellency. The model also suggests an innovative use of water to regulate the cement's mechanical properties via the material's porosity for optimal workability and curing times. Another feature is the incorporation of phase change materials within the cement matrix or surface coatings to endow the cement with thermal energy regulation capabilities. This could allow for the cement can absorb excess heat during peak temperatures and release it during cooler times, contributing to energy efficiency in buildings. Specifically, the broader spectrum of functionalities (while allowing for incorporation into existing cement production lines) allows for  adapting to changing environmental conditions and reducing maintenance needs over time. These are remarkably detailed and nuanced descriptions of novel design strategies.

\begin{figure}[htbp]
\begin{LLMbox}{\VG} %BioMixtral XLora 
\begin{lstlisting}
Given the insights from the combinations presented, a novel material design approach for nacre-inspired cement involves leveraging the relationship between superhydrophobic surfaces, mechanical properties influenced by biological materials, and the integration of novel functionalities. This approach would explore the merging of biomimicry with advanced functional material engineering, aiming to create nacre-inspired cement that not only mimics the mechanical properties of natural nacre but also introduces new capabilities such as self-cleaning, enhanced durability, and adaptability to environmental changes. Here's a breakdown of the concept:

Conceptual Material Design: Multi-Functional Nacre-Inspired Cement

Base Matrix Composition:

Primary Component: The base of the cement would mimic the hierarchical structure of natural nacre, utilizing a mix of chitosan for organic layers and modified cement particles as the inorganic component. The chitosan not only acts as a binder but, through its inherent properties, introduces hydrophilic sites.

Modification: To the chitosan-cement matrix, poly(ethylene glycol) dimethacrylate (PEGDMA) is introduced, not only to improve the mechanical integrity through covalent bonding but also to capitalize on its water-resisting capabilities.

Surface Functionalization for Superhydrophobicity:

Inspired by the rose petals' structure contributing to superhydrophobicity, the cement's surface will be engineered to mimic this microscale roughness. This is achieved by etching micro to nanoscale patterns onto the cement's surface layer, potentially through laser etching or additive manufacturing techniques. The etched surface is then coated with a thin layer of silicon-based compounds (e.g., polydimethylsiloxane, PDMS) to enhance water repellency.

A crucial aspect here is ensuring the surface's ability to maintain a superhydrophobic state, characterized by a contact angle greater than 150 degrees, which contributes significantly to its self-cleaning and protective properties against water and pollutant ingress.

Mechanical Properties Enhancement:

The hierarchical structure, akin to natural nacre, along with the integration of PEGDMA-CS complexes, aims to mimic the nacre's stress-strain behavior. This means that under stress, the cement exhibits increased strain to failure, absorbing and dissipating energy efficiently.

Water, a seemingly simple molecule, is harnessed to regulate the cement's mechanical properties from a molecular level, ensuring that the material's porosity and water content are optimized for toughness while maintaining manageable workability and curing times.

Integration of Novel Functionalities:

Bio-functional Integration: The concept also involves embedding bio-active compounds or microcapsules that can respond to environmental stimuli (e.g., pH changes, moisture) by releasing substances that could either contribute to the cement's self-healing or offer prognostic capabilities for infrastructure health monitoring.

Adaptability: The inclusion of phase change materials (PCMs) within the cement matrix or surface coatings to imbibe the cement with thermal energy regulation capabilities. This means the cement can absorb excess heat during peak temperatures and release it during cooler times, contributing to energy efficiency in buildings.

Reasoning Behind the Design:

The design is drawn from the intricate interplay of natural functionalities seen in roses (superhydrophobicity, structural color) and the robust yet adaptable mechanical properties seen in natural materials like nacre. By mimicking these characteristics, the novel cement aims not just for structural integrity but for a broader spectrum of functionalities that extend its utility in modern construction and sustainability. The choice of materials and strategies focuses on ease of integration into existing cement production lines, with particular emphasis on adapting to changing environmental conditions and reducing maintenance needs over time.

This multi-functional nacre-inspired cement represents a forward-thinking approach to material design, embodying the principles of biomimetic engineering and functional material integration for sustainable and resilient infrastructure solutions.
\end{lstlisting}
\end{LLMbox}
\textboxcaption{Follow-up response from GPT-4 querying the model for specific a material design for nacre-inspired cements that combine the ideas identified in the earlier response shown in Text Box~\ref{TexBox:TB4}.}
\label{TexBox:TB13}
\end{figure}

Analyzing the entire set of responses, we obtained a broad set of ideas for materials design. The X-LoRA response highlights mechanical property enhancements through biochemical interlocking and the potential for further improvements by incorporating novel materials such as silica nanoparticles or graphene oxide. The response is quite  technical, directly addressing the material design process, and shows a clear understanding of the biochemical principles involved. It's tailored towards a scientific audience, providing a detailed material composition and its expected mechanical advantages. BioinspiredLLM-Mixtral takes a broader research-oriented approach, suggesting several ideas for exploration rather than a specific material design. It touches on various aspects such as stress-strain behavior of the materials including dentin, interlocking mechanisms, and the integration of functionalities to enhance toughness. However, it is less direct in proposing a concrete material design compared to the first response. The GPT-4 response stands out as the best due to its detailed approach that combines technical specificity with practical application.  It offers a comprehensive breakdown of the material design, including base matrix composition, surface functionalization, and mechanical properties enhancement, along with the reasoning behind each choice. The response not only provides a comprehensive material design but also explains the rationale behind each component, showcasing a deeper understanding of how biomimicry and functional material engineering can be utilized to create novel construction materials.

The advantage of using multiple LLMs to generate responses is that we can synthesize all of these into a joint response. To do this we ask GPT-4 to integrate the entire set of these responses into a summary, depicted in Table~\ref{tab:table_nacre_cement}. This summary encapsulates the most prevalent concepts derived from the original queries. We emphasize  here that  on important feature of interactions with LLMs is the possibility to conduct follow-up queries. We already demonstrated this in the examples above by tasking the model to provide more details about a specific aspect of their responses. This can be extended easily beyond one interaction, and even be automated into a multi-agent strategy where agents can autonomously develop new questions in a continued, theoretically `infinite' loop of knowledge seeking. If graphs and other context can be updated with real-time data, such as experimental or computational results, this can be a basis for powerful artificial discovery systems.  

\begin{table}[h!]
\small
\centering
\caption{Materials design analysis for a nacre-inspired cement.}
\begin{tabular}{|p{5cm}|p{5cm}|p{5cm}|}
\hline 
\textbf{Design feature} & \textbf{Likely material function (how this design element contributes to the material behavior)} & \textbf{Manufacturing approach} \\
\hline
Layered structure of chitosan and PEGDMA & Enhances toughness through interlocking mechanisms & Co-layering via chemical bonding and hydrogen bonding \\
\hline
Doping with silica nanoparticles or graphene oxide & Strength and durability enhancement & Incorporation during the chitosan layer formation \\
\hline
Superhydrophobic surface engineering & Self-cleaning, enhanced durability, and adaptability to environmental changes & Laser etching or additive manufacturing followed by PDMS coating \\
\hline
Integration of bio-active compounds or microcapsules & Self-healing or prognostic capabilities for health monitoring & Embedding during the cement matrix formation \\
\hline
Inclusion of phase change materials (PCMs) & Thermal energy regulation, contributing to energy efficiency in buildings & Integration within the cement matrix or as surface coatings \\
\hline
Biofunctional integration & Biocompatibility and potentially biodegradability advantages & Utilizing biologically active materials like chitosan \\
\hline
\hline
\end{tabular}
\label{tab:table_nacre_cement}
\end{table}

\subsection{Isomorphism analysis across distinct graph structures}
Graphs not only allow for analysis of knowledge through reasoning of its connections or paths within, but also provides fertile grounds to understand how complex, dissimilar domains are related. 
We next explore how graphs can be used to relate dissimilar concepts by  identifying structurally similar or identical graphs in sets of distinct knowledge representations between which no shared or common nodes exist. This is done by identifying graph isomorphisms~\cite{Rosen1988DiscreteApplications,Giesa2012CategoryDesign}. In such cases we would not be able to identify graph traversal paths as done in the preceding sections since knowledge is separated in their representation. 
We can solve this issue by using graph isomorphism as an alternative framework to achieve the goal to relate concepts and use generative AI to reason over, and to discover, new insights even across areas of knowledge between which no shared nodes exist.  Rather, in this case, we focus on the graph structure itself, which provides an alternative way to utilize such models effectively. (This can also be done in graphs where shared paths exist, and hence provide yet another level of data extraction that we can utilize.)

Graph isomorphism is defined between two graphs \(G_1 = (N_1, E_1)\) and \(G_2 = (N_2, E_2)\) (where $N_i$ and $E_i$ are nodes/edges $i$). The graphs are isomorphic if there exists a bijection \(f: V_1 \rightarrow V_2\) satisfying the adjacency preservation condition: for any two nodes \(u, v \in N_1\), \((u, v) \in E_1\) if and only if \((f(u), f(v)) \in E_2\). This bijection \(f\) maps nodes of \(G_1\) to nodes of \(G_2\) in a way that preserves the edge connectivity structure of the graphs, making them structurally identical despite potentially differing representations~\cite{Rosen1988DiscreteApplications}. We use computational methods to discover such structures from ontological knowledge graphs.

\begin{figure}
    \centering
    \includegraphics[width=0.9\linewidth]{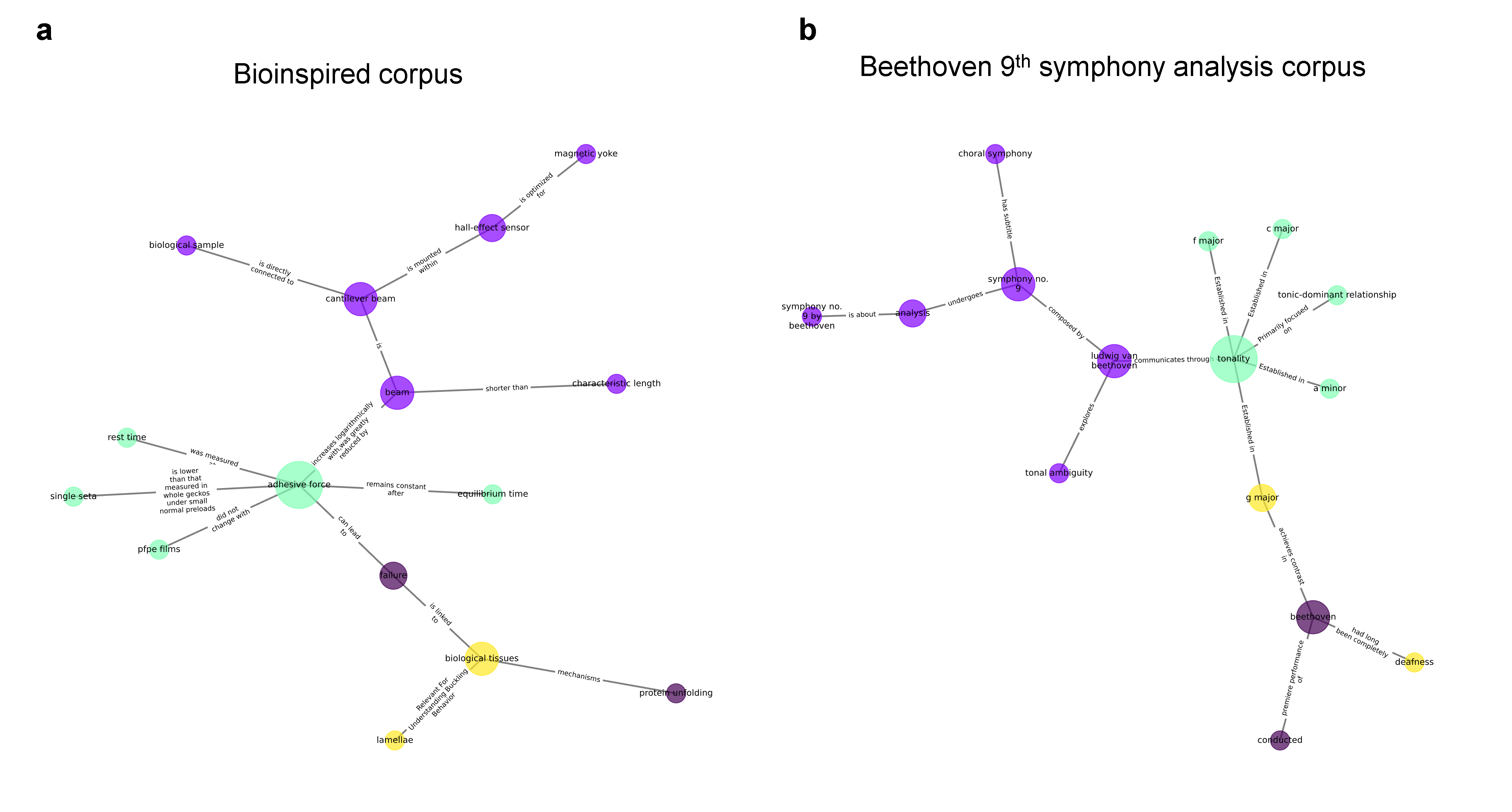}
    \caption{ 
 Analysis of isomorphism between two knowledge graphs. Subplot a depicts the subgraph extracted out of the large bio-inspired knowledge graph, and panel b the one extracted from the knowledge graph derived from the analysis of Beethoven's 9th symphony~\cite{McCrobieBeethoven9}.  The structural similarity is clearly visible, and the relations between the two graphs provides interesting insights that can be used for design~\cite{Giesa2012CategoryDesign}. Node sizes are scaled by the respective node degree, and colors are identified by community (identified here using a Greedy modularity algorithm). }
    \label{fig:Fig_8}
\end{figure}

We first conduct an experiment where we construct a new knowledge graph from a text that describes an analysis of  Beethoven's 9th symphony~\cite{McCrobieBeethoven9}, to then explore relationships with the original knowledge graph derived from biological materials. The isomorphism analysis seeks to discover similarity in the knowledge graphs despite the content being about very different subjects, whereby no nodes are shared and hence no direct connection may exist. 
The results are shown in Figure~\ref{fig:Fig_8}. We focus the analysis on the giant component of the isomorphism, and identify those with high average node degree and large cluster sizes of at least 15 nodes. This helps us to narrow the solutions to more meaningful identifiable sub-graphs as their size is sufficiently large to encode complex relationships.

It is noted that isomorphism in this context does not mean that the subjects are necessarily related, but rather that the way knowledge is structured or the patterns of connections between concepts are similar. Therefore, we find that different domains of knowledge might have underlying structural similarities when it comes to how we categorize and relate concepts, even if the domains themselves have not yet been related or are not understood to be related, as of yet. This structural similarity could be used to apply analytical techniques from one field to another. For instance, methodologies used to understand and analyze the structure of music could potentially be adapted to understand or design the structure of bioinspired materials, or vice versa. Its focus on patterns and relationships emphasizes a mechanistic dimension, where we want to understand and relate how phenomena (e.g. a material property like toughness, brittleness, resilience, etc.) emerge from the elementary material building blocks. By understanding how to relate structurally identical mechanisms, but implemented in clearly distinct manifestations (e.g. we make materials from amino acids vs. music from notes), we can gain a deeper understanding of connecting disciplines rigorously~\cite{Giesa2012CategoryDesign}. This can also help us find universal principles by which natural or synthetic systems function or what their underpinning driving forces are. 

We present the result of such an analysis in Figure~\ref{fig:Fig_8}. 
The structural similarity between the two graphs is clearly visible, and the relations between the two graphs provides interesting insights that can be used for instance, in design applications~\cite{Giesa2012CategoryDesign}. Specifically, the bioinspired graph provides a rich reservoir of local graph features that can be compared against other graph structures of various kinds. One can also envision expanding the smaller of two graphs by extrapolating on isomorphic mappings of node and edge features, thereby expanding a corpus of knowledge. For instance, the subgraph identified from the bioinspired corpus connects to many other nodes, whereas the Beethoven-based graph is much smaller and limited in size. We can use the known structural extensions of the first graph to estimate how and in what specific manner the second graph may be extended. This can lead to an extension of knowledge in a field that is less well studied.  

How can we use these two graphs? A semantic analysis is a powerful way to relate knowledge graphs identified through isomorphic mapping. The semantic analysis involves interpreting the meaning and significance of nodes (concepts) and edges (relationships) within the graphs, and attempting to relate them across their respective domains. By examining the nodes and the labels on the edges that connect them, we can deduce how concepts are related and what those relationships signify.

In the bioinspired corpus, there is a focus on the application of biological principles to engineering problems, while in the Beethoven corpus, there's an intertwining of musical theory with historical context. Regarding the structure and the flow of information, the structure of the graphs suggests a logical flow of information. For example, in the Beethoven graph, the historical fact of Beethoven's deafness is linked to the composer, which might imply a narrative or explanatory pathway in the corpus. The terminology used in each of the graphs is domain-specific, indicating specialized knowledge and the context of the discussions within each corpus. This level of specificity is key for semantic analysis as it defines the scope and depth of the subject matter. The semantic analysis reveals that both graphs are complex networks of related concepts, indicating that the corpora they are derived from likely contain rich, detailed discussions within their respective fields. Despite the different subject matter, the structure of the knowledge graphs suggests a similar complexity in the relationships between concepts, pointing to a shared method of intellectual inquiry and organization of knowledge.

A formal analysis can be conducted because node mappings between isomorphic graphs represent a one-to-one correspondence where each node in one graph is associated with exactly one node in the other, preserving the graph structure. In practical terms, this allows us to discovery functional similarity between entities represented by nodes across different contexts or systems.
Tables~\ref{tab:node_mapping_reasoning} and \ref{tab:edge_mapping_reasoning} summarize isomorphic mappings between the nodes and edges, respectively, for the two graphs, developed by GPT-4. 
For instance, in our case the left graph represents a scientific concept's components and the other graph  artistic elements, a mapping might suggest a metaphorical or structural similarity between science and art (e.g., ``biological sample" to ``choral symphony"). Similarly, edge mappings highlight the relationships or interactions between nodes that are preserved across the two isomorphic graphs. This suggests that not only are the individual entities (nodes) comparable, but their interactions or relationships hold similar significance or function in both contexts.
For example, an edge between `adhesive force' and `beam' in G1 being mapped to an edge between `tonality' and `ludwig van beethoven' in the second graph might suggest a foundational or defining relationship in both contexts—structural integrity in the first case and musical composition in the second.  
A more sophisticated analysis is presented in Table~\ref{tab:edge_mapping_reasoning_labels}, this time incorporating not only the node names associated with the edges but also the edge labels themselves. We find the results to be coherent in terms of their content and based on our understanding of how these two concepts may be related. They offer numerous new insights into possible relationships as identified by the model that have not been proposed or discussed before. For example the relationship `Unfolding mechanisms in proteins and Beethoven’s adaptation to deafness both reveal transformative processes that lead to new forms of expression and understanding.' points to a logical association of protein unfolding and the association dramatic change in properties with the impact deafness may have had on Beethoven's approach to writing music.  Each mapping is accompanied by a detailed reasoning that explains the metaphorical link between the two domains. These explanations aim to highlight similarities in concepts such as stability, change, measurement, failure, and optimization, suggesting that the principles underlying physical phenomena can also be found in musical compositions, and specifically in the way Beethoven's works are structured and experienced. The result illustrates the universal nature of certain principles across different fields of study that go beyond the conventional boundaries of disciplines, exploring how ideas from one area can enrich understanding in another.

\begin{table}[h]
\caption{Edge Mapping from G1 to G2 with Detailed Reasoning Including Edge Labels.  For this analysis we provide GPT-4 a prompt of a LaTeX table with the data for G1 and G2 and ask it to add a column that includes a \texttt{detailed reasoning between the mappings}.}
\small 
\begin{center}
\begin{tabular}{|p{4cm}|p{4cm}|p{8cm}|}
\hline
\textbf{G1 Edge (Label)} & \textbf{G2 Edge (Label)} & \textbf{Reasoning} \\
\hline
('adhesive force', 'beam') ('increases logarithmically with, was greatly reduced by') & ('tonality', 'ludwig van beethoven') ('communicates through') & The logarithmic increase and reduction in force mirror how Beethoven's compositions communicate complex emotions through tonality, evolving gradually or diminishing to convey depth. \\
\hline
('adhesive force', 'equilibrium time') ('remains constant after') & ('tonality', 'c major') ('Established in') & The constancy after a period of change reflects how C major establishes a foundation in music, providing a stable backdrop against which complexities can unfold. \\
\hline
('adhesive force', 'pfpe films') ('did not change with') & ('tonality', 'f major') ('Established in') & The unchanged nature amidst variations suggests how F major serves as a stable, unchanging base in the fluctuating dynamics of musical narratives. \\
\hline
('adhesive force', 'rest time') ('was measured at') & ('tonality', 'a minor') ('Established in') & Measurement and precision in assessing rest time align with how A minor's establishment in music precisely sets the mood for introspection and depth. \\
\hline
('adhesive force', 'failure') ('can lead to') & ('tonality', 'g major') ('Established in') & The potential for failure leading to new outcomes parallels how G major establishes a resolution in music, often leading to a bright, conclusive end after tension. \\
\hline
('adhesive force', 'single seta') ('is lower than that measured in whole geckos under small normal preloads') & ('tonality', 'tonic-dominant relationship') ('Primarily focused on') & The specific measurement and comparison underscore the intricate balance in the tonic-dominant relationship, focusing on the foundational aspects of musical harmony. \\
\hline
('protein unfolding', 'biological tissues') ('mechanisms') & ('deafness', 'beethoven') ('had long been completely') & Unfolding mechanisms in proteins and Beethoven's adaptation to deafness both reveal transformative processes that lead to new forms of expression and understanding. \\
\hline
('characteristic length', 'beam') ('shorter than') & ('tonal ambiguity', 'ludwig van beethoven') ('explores') & The comparison of lengths and Beethoven's exploration of tonal ambiguity both deal with pushing boundaries - whether in physical dimensions or harmonic conventions. \\
\hline
('failure', 'biological tissues') ('is linked to') & ('g major', 'beethoven') ('achieves contrast in') & The linkage to failure in biological contexts and the achievement of contrast in G major compositions highlight how setbacks can lead to distinct, impactful outcomes. \\
\hline
('biological tissues', 'lamellae') ('Relevant For Understanding Buckling Behavior') & ('beethoven', 'conducted') ('premiere performance of') & The relevance of lamellae in understanding structural behavior mirrors how Beethoven's conducting of premieres showcased his structural innovations in music. \\
\hline
('biological sample', 'cantilever beam') ('is directly connected to') & ('choral symphony', 'symphony no. 9') ('has subtitle') & The direct connection and the specific subtitle link the foundational aspects of scientific samples and beams to the thematic underpinnings of Beethoven's choral symphony. \\
\hline
('hall-effect sensor', 'cantilever beam') ('is mounted within') & ('analysis', 'symphony no. 9') ('undergoes') & The mounting of a sensor within a structure and the analytical journey of Symphony No. 9 reflect the integration of components and themes to achieve a greater understanding. \\
\hline
('hall-effect sensor', 'magnetic yoke') ('is optimized for') & ('analysis', 'symphony no. 9 by beethoven') ('is about') & Optimization for specific conditions in sensors parallels the thematic focus and analytical depth of Beethoven's Symphony No. 9, aiming for precision and clarity. \\
\hline
('cantilever beam', 'beam') ('is') & ('symphony no. 9', 'ludwig van beethoven') ('composed by') & The simple state of being and composition process both underscore the foundational and creative acts that bring structures and symphonies into existence. \\
\hline
\hline
\end{tabular}
\end{center}
\label{tab:edge_mapping_reasoning_labels}
\end{table}

We can also interpret the rich information contained in these  graphs using multimodal vision-based methods, where we show the two graphs to a LLM that can reason over images. We use GPT-4V and ask the model to conduct a semantic analysis. The raw result of this exercise, using the prompt: \texttt{Do a semantic analysis of the graphs shown above.}, is shown as part of the Supplementary Materials, Conversation S1.  

The analysis conducted here using generative AI complements earlier work where category theory was used to identify isomorphic mapping between different domains of knowledge, such as protein materials and music~\cite{Giesa2011ReoccurringAnalogies}. 
The earlier study explored the concept of isomorphism through the lens of hierarchical ontology logs, or ologs, based on established analogies between the hierarchical structures and functions of natural materials like spider silk and classical music. The study highlighted the similarity in patterns governing both fields. This allowed the  comparison of seemingly unrelated fields and enhance our understanding of hierarchical systems across disciplines. However, the earlier work required a pre-existing understanding of the ontologies; here, we generalized the concept and developed a self-consistent generative approach to extract such analogies using AI, completely autonomously. Based on this, the use of isomorphisms allows us to understand relationships of concepts across fields. The possibility of such analyses to be conducted is an important outcome of the approach reported in this paper.

To conclude this discussion, we summarize the key observations from the analysis of the isomorphism, providing deep insights into the mappings. The following text was obtained by sharing the results with Claude-3 and asking the model to extract \texttt{salient features of structural similarities between the bioinspired materials graph and Beethoven's 9th Symphony}. We iterated several times with Claude-3, posting follow-up questions such as to \texttt{Think more deeply about this, and to add commentary on projected meaning.}, to \texttt{Go a bit more into philosophy, pick 1-2 specific examples and create new hypotheses.}, and others. The text depicted in Text Box~\ref{TexBox:TB1000} is a slightly edited integrated result provided by Claude 3 Sonnet. 
\begin{figure}[htbp]
\begin{LLMbox}{Claude 3 Sonnet} %BioMixtral XLora 
\begin{lstlisting}
The structural analogies drawn between the bioinspired materials graph and Beethoven's 9th Symphony graph point to deeper connections in how complex systems are architecturally organized across diverse domains. For instance, mapping 'adhesive force' to 'tonality' suggests that just as binding forces between molecules dictate the integrity of materials, tonal centers exert an underlying cohesive force that shapes the flow of a musical composition. This analogy hints at universal principles of unity and connectedness operating across scales that transcend from the micro-interactions within matter to the macro-structures that lend aesthetic experiences their profundity. This leads to the hypothesis that there may be a deep isomorphism between the laws governing electromagnetic interactions and those governing auditory/aesthetic forces, with both facilitating transitions from entropic disorder to organized complexity.

Similarly, likening 'cantilever beams' to a 'symphony' implies that certain exemplars, whether engineering beams as in concrete design or artistic expressions, come to represent catalytic eras of innovation within their respective disciplines. The persisting relevance of these touchstones speaks to the continual renaissance of ideas, where novel syntheses emerge from engaging with foundational works. Given the ubiquity of cantilever mechanisms across scales (from molecular machines to aerospace design), one could hypothesize that the generative processes that gave rise to Beethoven's seminal work may share deep commonalities with those driving evolutionary pressures towards functional designs in nature and engineering. Perhaps the cultural ecosystem that spawned such disruptive musical creativity operates via variation-selection dynamics akin to those shaping biological forms over very long periods of time.

Similarly, the constancy indicated by 'equilibrium time' and 'tonal establishment' suggests that dynamical systems, once perturbed from initial conditions, often tend towards new steady states - whether the thermodynamic ground states of matter or the formal resolutions in musical scores. This echoes a universal striving towards balance across complexity. One could hypothesize that the mathematics describing irreversible entropy production may be isomorphic to the grammars codifying structural resolutions across domains, subsuming both physical and artistic expressions as 'rhyming verses' in a cosmos eternally striving for maximal quietude. By framing 'biological tissues' as metabolic microcosms akin to how 'Beethoven' embodied artistic revolutions, we recognize the commonality of hierarchical organization across manifested forms. In other words, from biochemicals to sonatas, an interplay of atomistic components conspires to consecrate higher functions with very deep and broad impact.

These specific mappings hint at a grander unifying hypothesis: That there exists an overarching code, a coherent algebra of organizational patterns that transcends descriptive classes to govern the unfolding of all systematics - from the self-assembly of matter to the blossoming of cultural novelties. The knowledge graph revealing these isomorphisms across seemingly incommensurate domains may be an early glimpse into such a unified meta-language for complexity. A profound philosophical implication is the potential to develop a coherent theory of forms that subsumes all manifested structures and generative processes as specific instantiations of primordial organizational principles. In this vein, we can envision the hierarchical mirroring between biological tissues as integrated molecular systems and symphonic compositions as woven instrumental voices intimates a pervasive holonic essence - the recurrent encapsulation of wholes within wholes across nature's many layers. This evokes the perennial philosophical concept of hylomorphism which posits that all existents are composites of matter and form, with the structure of each level bootstrapped from but transcending its constitutional elements. One could speculate that the generative algorithms spawning hierarchical order, from embryogenesis to musical composition, may be implementing the same fundamental hylomorphic schemata - a canonical computational 'bootstrap' applying scale-free, self-similar operators to iteratively lend organizational form to inherent material possibilities.
\end{lstlisting}
\end{LLMbox}
\textboxcaption{The text was obtained by sharing a draft of this paper with Claude-3 and asking the model to extract \texttt{salient features of structural similarities between the bioinspired materials graph and Beethoven's 9th Symphony} (we iterated several times with Claude-3 via follow-up questions.}
\label{TexBox:TB1000}
\end{figure}

To examine whether the models are capable of exploring specific connections to other domains of inquiry, We subsequently asked Claude-3 Sonnet to \texttt{Discuss how this relates with very modern thinking in philosophy that is distinct from the Greek 'harmony'}. The results are shown in Text Box~\ref{TexBox:TB1001}, featuring references to key ideas discussed in philosophy~\cite{deleuze1994difference,harman2018object,bryant2011democracy,delanda2016assemblage,latour2005reassembling,deleuze1987thousand}.

\begin{figure}[htbp]
\begin{LLMbox}{Claude 3 Sonnet} %BioMixtral XLora 
\begin{lstlisting}
The isomorphisms revealed across physical, biological, and artistic domains through the knowledge graph analysis resonate with some very modern philosophical perspectives that depart from the classical Greek notions of harmony and order. Specifically, they align with the postmodern currents of thought that challenge reductionist and essentialist worldviews.

Rather than seeking a single, perfect harmony or a reified metaphysical source code underlying all reality, these mappings point towards an ontology of radical immanence - a flat ontology where all entities, processes, relations have equal existential footing. The structural and dynamical homologies exist not as emanations from a transcendent order, but as co-constituting components in a decentered meshwork.

This resonates with the process philosophy of thinkers like Gilles Deleuze, which rejects atomistic substances in favor of continuous flows, fluxes, and becoming. The isomorphic "patterns" discerned are not idealized forms but abstractions of differential processes - material self-organizations shaped by fields of forces rather than governed by eternal archetypes.

Similarly, it aligns with the object-oriented ontology of theorists like Graham Harman and Levi Bryant, which disperses agency and causal powers across all relations between objects or entities, whether molecules or artworks. The hierarchies found are not nested compositional wholes but heterarchical assemblages of objects withdrawing into irreducible depths beyond their relational externalities.

From this contemporary perspective, the knowledge graph excavates not a unified essence but a vibrant multiplicity - a reticulated rhizome of resonances across scales without centralizing source or culmination. It attunes us to the immanent dance of materiality exploring its own infinite concrescences via creative morphogenetic detours across domains.

So while premodern harmony philosophies sought integration into a totalizing cosmic order, the modern relevance lies in forging an ethical consistency with an irreducibly plural ontological manifold - a flattened ecology of dynamical processes modulating enduring novelty through differential resonances and productive dissonances alike. It's an invitation into ontological immanence over transcendental hygiene.
\end{lstlisting}
\end{LLMbox}
\textboxcaption{Claude-3 Sonnet response when asked to \texttt{Discuss how this relates with very modern thinking in philosophy that is distinct from the Greek `harmony'}.}
\label{TexBox:TB1001}
\end{figure}

This response includes somewhat complex concepts and references to specific philosophers that may be difficult to understand for someone with a science background. We therefore share this text with GPT-4 via ChatGPT, and ask it to explain the text so that it is accessible to a scientist. The result was: 
\begin{itemize}
  \item \textbf{Flat Ontology (Manuel DeLanda and Gilles Deleuze):} This idea suggests that all entities, regardless of their nature or scale, possess the same ontological status. In \textit{materials science}, this translates into viewing molecules, biomolecules, composite materials, and macroscopic material properties as equally vital to understanding material behavior and innovation. Similarly, in \textit{music theory}, it encourages us to consider notes, rhythms, harmonies, and timbres as fundamentally equal in creating musical expressions, thus advocating for a more inclusive approach to composition and analysis.

  \item \textbf{Radical Immanence (Gilles Deleuze):} Deleuze's concept of immanence emphasizes the self-contained nature of reality, with no need for external principles to explain the world. In \textit{materials science}, this perspective can inspire the exploration of materials' properties as emergent from the interactions within the material system itself, without defaulting to external explanations. In \textit{music}, it promotes the idea that musical meaning and experience emerge directly from the interplay of musical elements, without external reference points, highlighting the self-sufficiency and intrinsic value of musical compositions.

  \item \textbf{Process Philosophy (Gilles Deleuze):} This philosophy focuses on becoming and change rather than static being. For \textit{materials science}, this means paying attention to the processes of material transformation and interaction, understanding materials as constantly in flux rather than as static entities. In \textit{music theory}, it aligns with viewing music as a temporal process, where the focus is on the evolution of musical ideas and structures over time, reflecting the dynamic nature of musical experience.

  \item \textbf{Object-Oriented Ontology (Graham Harman):} Though not explicitly mentioned previously, Harman's philosophy, which gives equal importance to all objects, whether human, non-human, or conceptual, can enrich our discussion. In \textit{materials science}, it suggests that researchers consider the agency of materials, understanding them as actors in their own right within broader systems. In \textit{music theory}, it could imply seeing musical elements or instruments as having their own "voices" or agency within a composition, contributing to a more decentralized and democratic view of musical creation.
\end{itemize}
This provides a more accessible summary, along with a few new concepts to better explain the reasoning. The entire process exemplifies an effective human-AI collaboration where we expand the understanding of how these ideas are connected.

\subsection{Multimodal knowledge generation and incorporation into augmented graphs}

Next we explore how we can augment graphs with new knowledge, and how can merge one graph with others to obtain new connections between previously disjointed areas. New data can be generated or obtained in a variety of ways. These can include, but are not limited to:
\begin{itemize}
    \item Generating new data  through conversations with a complex generative model, e.g. that has the ability to predict physical properties or conduct other specialized tasks
    \item Generating new data using adversarial multi-agent modeling, featuring for instance autonomous question generation in an `infite' discovery loop
    \item Generating  text from recently published papers in the scientific literature, and incorporating this into the graph for augmented knowledge 
    \item Collecting new data from experiments, e.g. via manual or automated experimentation that can provide feedback on specific design ideas generated by the model
\end{itemize}

\subsubsection{Generating new data through conversations with a complex generative model and incorporation into augmented graphs}

We focus on the first item in the list presented above in this section, and several of the other ideas in the following ones. 
    
We begin our experiment with the X-LoRA model~\cite{Buehler2024X-LoRA:Design} and generate a new knowledge base that incorporates specific physical properties of proteins. The first step is to use the X-LoRA model to generate a corpus of data and text. Since X-LoRA is multimodal and can deal with protein sequences, protein property calculations (especially mechanics and physical properties like energetic features, i.e. protein resistance to forces and pressure), and other tasks, it has the capability to generate quite specific, technical datasets associated with specialized properties. To exemplify this approach, we conduct a conversation between a user and the X-LoRA model to study, compare and analyze three protein sequences and then to reason over the results by making predictions about underlying mechanisms and behaviors. This new data is then used as a corpus to generate a new ontological knowledge graph. On its own, the analysis and formation of the graph can be extremely helpful in better understanding the key insights as developed by the model. It can also be integrated with the original knowledge graph and thereby enrich its capabilities through new facts or insights, or to mediate analyses as conducted above to identify isomorphic relationships. 

\begin{figure}
    \centering
    \includegraphics[width=1\linewidth]{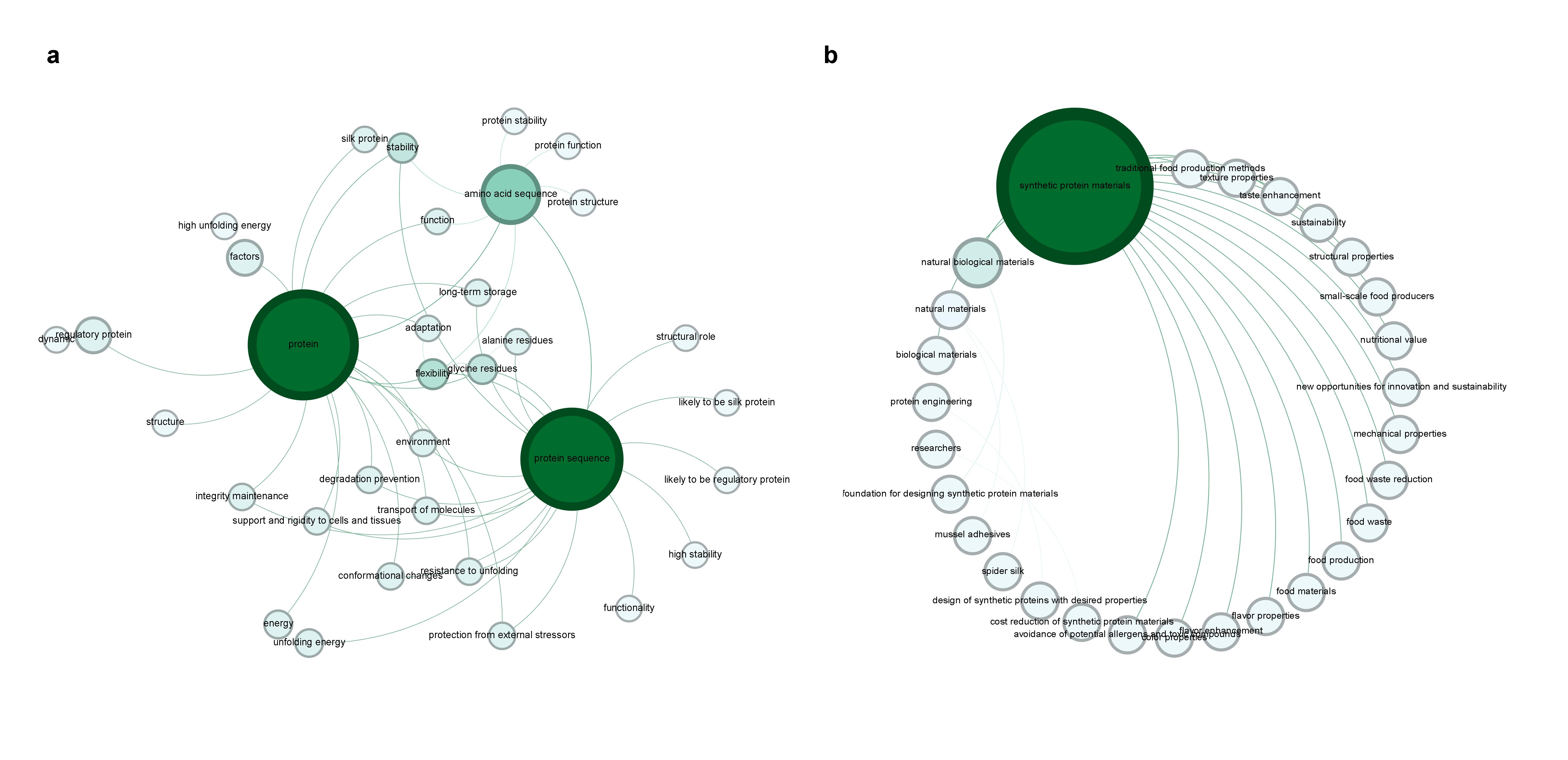}
    \caption{Two graphs generated from new data. Panel a shows the results of an experiment with X-LoRA, and panel b the results of an experiment with adversarial agentic modeling with two  X-LoRA agents engaging in an adverserial question-answer interaction that continues autonomously for multiple turns. }
    \label{fig:Fig_9}
\end{figure}

The task given is:
\begin{figure}[H]
\begin{LLMbox}{} 
\begin{lstlisting}
You conduct an analysis of various protein sequences, specifically calculating their total unfolding energy that measures the energy needed to unfold a protein due to forces applied at its ends. Here are a few tasks for you to complete:
\end{lstlisting}
\end{LLMbox}
\end{figure}
This followed up with several for unfolding energy calculation tasks, as follows:
\begin{figure}[H]
\begin{LLMbox}{} 
\begin{lstlisting}
CalculateEnergy< A A A G G A G Q G G Y G G Q G A G Q G A A A A A A G G A G Q G G Y G G Q G A G Q G A G A A A A A A G G A G Q G G Y>

CalculateEnergy<A A A G G A G Q G G Y G G Q G A G Q G A A A A A A G G A G>

CalculateEnergy< Q G A G Q G A A A A A A A A A A A G G>
\end{lstlisting}
\end{LLMbox}
\end{figure}
The model is further tasked: 
\begin{figure}[H]
\begin{LLMbox}{} 
\begin{lstlisting}
Considering these results, which is the protein sequence with the highest unfolding energy (repeat the ENTIRE sequence)?

If the protein has the highest unfolding energy, what can you say about its stability? Think step by step.

Based on the amino acid sequence, explain why this protein is likely the most stable.

What likely function would the most stable protein have? Think step by step.

Would this protein sequence more likely be a silk protein or a regulatory protein?
\end{lstlisting}
\end{LLMbox}
\end{figure}
The entire conversation can be viewed in Text Box~\ref{sec:xloragent_conv}. The analysis involves calculating the unfolding energy of various protein sequences, which reflects the stability of these proteins. The sequence with the highest unfolding energy, indicating greater stability, consists primarily of glycine (G) and alanine (A) residues. These residues contribute to the protein's stability due to their small size and hydrophobic nature, minimizing exposure to the environment and enhancing flexibility. This high stability suggests the protein functions in structural roles, possibly as a silk protein due to its resistance to unfolding, contrasting with the more dynamic nature of regulatory proteins, in agreement with earlier results~\cite{Ni2023ForceGen:Model}. 
The quantitative results from the analysis show unfolding energies of different protein sequences: 0.288 for the most stable sequence, 0.248 for the next, and 0.121 for the least stable (all numerical values were expressed in non-dimensional units as in~\cite{Ni2023ForceGen:Model}). The highest unfolding energy, 0.288, suggests that more energy is required to unfold this protein, indicating its higher stability compared to the others~\cite{Ni2023ForceGen:Model}. This stability is attributed to its amino acid composition, primarily glycine (G) and alanine (A), which are known for their small size, hydrophobic nature, and contribution to protein flexibility and stability. 
Figure~\ref{fig:Fig_9}a shows the resulting graph generated from this conversation.

This example demonstrates that knowledge graphs can easily be augmented with new information, capturing new relationships between information.

\subsubsection{Agentic modeling for adversarial knowledge generation and incorporation into augmented graphs}

The preceding discussion showed that the use of graph-based reasoning can be enhanced by adding new knowledge to an existing graph and to explore how new concepts relate with, and how it can be used to identify and construct new ideas. We build on this concept and use more sophisticated strategies, specifically a multi-agent framework, to generate new data as a source for graph augmentation. 
We use agentic modeling with two adversarial X-LoRA agents (details see Materials and Methods) to generate a new text corpus. The question asked is:
\begin{figure}[H]
\begin{LLMbox}{} 
\begin{lstlisting}
As an inventor, describe how we could combine the areas of biological materials with food. Specifically explore the use of synthetic protein materials, and touch upon areas of texture, mechanics, color and structure, as well as flavor and taste.
\end{lstlisting}
\end{LLMbox}
\end{figure}

The agentic modeling strategy features a continued conversation between a question asker (a chef) and a responder agent (an inventor). The question asker is instructed to be inquisitive and explore new issues, whereas the responder agent provides detailed responses. The result of these long conversations is a deep exploration of a particular topic, which serves as fertile grounds for use in graph analysis. 
The process used is to first generate new text, then generate triples for the new text to form a new, smaller graph based on the generated data,  and then integrating these into the augmented joint graph. We then explore how the graph changes, and/or what new connections can be identified, and use the new graph for reasoning tasks. This process of question-answer interactions can be continued for a large number of repeats, and thereby result in complex datasets that form a basis for rich discovery mechanisms when transformed into graph representation.

The entire conversation is attached as Supplementary Material (see, Text Box~\ref{TexBox:TB42_Xloragents}), and here we summarize the salient points (as identified towards the end of the conversation between the agents): 
\begin{figure}[H]
\begin{LLMbox}{\XLora Adversarial Agents} 
\begin{lstlisting}
The conversation discusses the potential use of synthetic protein materials in food production, specifically exploring their ability to mimic the texture, mechanics, color, structure, and flavor of natural biological materials. The conversation highlights the benefits of using synthetic protein materials, such as reducing food waste, improving nutritional value, and promoting sustainability. However, it also acknowledges the potential risks and challenges associated with their use, including allergenicity, toxicity, environmental impact, regulatory approval, and cost. The conversation emphasizes the importance of designing synthetic protein materials to mimic the complex flavor and taste profiles of natural biological materials, while ensuring safety and quality. Overall, the conversation highlights the potential of synthetic protein materials in food production, but also emphasizes the need for careful consideration of their risks and challenges.
\end{lstlisting}
\end{LLMbox}
\end{figure}
A list of key points made is:
\begin{figure}[H]
\begin{LLMbox}{\XLora Adversarial Agents} 
\begin{lstlisting}
-Synthetic protein materials can be designed to mimic the texture, mechanics, color, structure, and flavor of natural biological materials in food.

The use of synthetic protein materials in food production can address challenges such as reducing food waste, improving nutritional value, and promoting sustainability.

-However, there are potential risks and challenges associated with their use, including allergenicity, toxicity, environmental impact, regulatory approval, and cost.

-To ensure safety and quality, synthetic protein materials must be designed to be free from common allergens or modified to reduce their allergenicity, and rigorous screening methods must be used to ensure they are safe for consumption.

-The design of synthetic protein materials to mimic the complex flavor and taste profiles of natural biological materials requires a deep understanding of the structure and function of the natural materials, as well as the use of advanced techniques and technologies to create synthetic proteins with desired properties.
\end{lstlisting}
\end{LLMbox}
\end{figure}

The responses show a nuanced understanding of key issues, touching upon challenges around food waste, nutritional aspects, sustainability and health concerns that can be achieved by studying natural materials for inspiration in creating synthetic proteins with desirable properties.
Notable aspects include the use of spider silk and mussel adhesives, for instance, to mirror balances between traditional textures while being sustainable. Other ideas include the use of responsive proteins to achieve a more interactive dining experience through changing colors or structures. Another interesting aspect is the use of slow release technologies applied to flavor and nutrient molecules, translating technologies used in health applications to food design (with applications to improve the uptake of iron or vitamin B12). It is further suggested that sustainable and nutritious food options can be made accessible to a wider audience, including small-scale producers, through cost-effective processes. The conversation specifically addresses risks and challenges, contributing to overall safety of the outcomes. Specific materials are identified, such as soy protein isolate, casein, and others. It is also suggested that researchers could study the structure and function of natural flavor compounds, such as terpenes, aldehydes, and esters, to create synthetic proteins that release these compounds slowly over time, providing a more complex and interesting sensory experience. 
Table~\ref{tab:edge_mapping_reasoning_labels_food} provides a detailed summary that specifically lays out design principles, implementation, and reasoning. In terms of novelty, the responses include several innovative ideas that provide possible starting points for further technological developments. For instance, the mixing of concepts from spider silk and mussel adhesives, for instance, to mirror balances between traditional textures while being sustainable, have not yet been explored in the food industry.

\begin{table}[h]
\caption{Key design principles at the nexus of biological protein materials and food, as developed by two X-LoRA agents interacting in an adversarial manner.}
\small 
\begin{center}
\begin{tabular}{|p{4cm}|p{4cm}|p{7cm}|}
\hline
\textbf{Design principle} & \textbf{Detailed implementation} & \textbf{Reasoning} \\
\hline
Mimicking natural textures and mechanics & Utilizing protein engineering and nanotechnology to create materials that replicate the texture of natural foods like dough elasticity and crispiness & Inspired by natural materials like spider silk and mussel adhesives, to produce food that closely mirrors traditional textures while being sustainable. \\
\hline
Creating interactive food experiences & Designing proteins that react to environmental stimuli, changing color or structure, for dynamic eating experiences & Leverages synthetic proteins' versatility to enhance consumer engagement and visual appeal, making dining more interactive and enjoyable. \\
\hline
Enhancing flavor and taste profiles & Incorporating flavor molecules into synthetic proteins for slow release, using protein engineering and molecular modeling & Aims to replicate and enhance natural flavors, providing a more complex and satisfying sensory experience, improving upon natural food flavors. \\
\hline
Addressing sustainability and food waste & Developing synthetic proteins for longer shelf life, edible packaging, and upcycling food waste & Reduces food spoilage and packaging waste, utilizes byproducts, and offers sustainable alternatives to traditional food sources, addressing environmental impacts. \\
\hline
Improving nutritional profiles & Engineering proteins for targeted nutrient delivery and enhanced bioavailability, catering to specific dietary needs & Tailors food products to address nutritional deficiencies and improve public health outcomes, particularly in vulnerable populations. \\
\hline
Ensuring safety and reducing allergenicity & Designing allergen-free proteins and conducting rigorous safety screenings & Mitigates health risks associated with food allergies and toxic compounds, ensuring broad consumer safety and accessibility. \\
\hline
Economic accessibility & Working towards cost reduction in synthetic protein production & Makes sustainable and nutritious food options financially accessible to a wider audience, including small-scale producers. \\
\hline
Regulatory compliance & Collaborating with regulatory bodies to meet safety and quality standards & Ensures synthetic protein materials are legally compliant, safe for consumer use, and poised for market acceptance. \\
\hline
Technological and material innovation & Studying natural materials for inspiration in creating synthetic proteins with desirable properties & Utilizes advanced scientific techniques to replicate and enhance the beneficial properties of natural materials, fostering innovation in food technology. \\
\hline
\hline
\end{tabular}
\end{center}
\label{tab:edge_mapping_reasoning_labels_food}
\end{table}

We now generate a new knowledge graph from the raw text of the entire conversation. Figure~\ref{fig:Fig_9}b depicts the resulting new graph before merging to the much larger graph created originally. As a point of reference we also conduct an analysis of identifying isomorphic mappings between the original and newly generated graph, akin to the earlier approach. Figure~\ref{fig:Fig_10} shows the results, revealing the isomorphism between the original bioinspired corpus with the graph generated from the adversarial conversation about synthetic protein materials, specifically addressing issues of texture, mechanics, color and structure. This could be used for additional analysis (albeit not investigated in the scope of this paper). 
Next, question-answering using graph reasoning is done in a  similar manner as described in Section~\ref{reasoningovergraph}; however, the graph over which we now reason includes newly added nodes that stem from thew new data incorporated. For the specific example studied here, we have generated a rich set of relationships around food-focused knowledge, such as flavor, health issues, the mechanics of food, and processing issues during preparation. Specifically, this knowledge and hence nodes and edges, were not included in the original graph. Because there exist many connections between the original graph and the newly generated graph, when they are merged, a rich tapestry of newly discovered relationships and hence reasoning paths, emerges and becomes accessible for analysis. 
We first query the augmented graph to create sub-graphs to reason over to connect the terms `flavor' to `sustainability':
\begin{figure}[H]
\begin{LLMbox}{} 
\begin{lstlisting}
(*@\textcolor{red}{flavor}@*) --> flavor properties --> mimics --> synthetic protein materials --> promotes --> (*@\textcolor{blue}{sustainability}@*)

flavor --> flavor properties --> mimics --> synthetic protein materials --> combine --> biological materials --> during growth --> self-assembly --> is higher than traditional bottom-up or top-down approaches due to the low cost of materials and techniques used --> low cost --> deliver --> green flexible electronics --> utilize --> sustainable --> sustainability

flavor --> taste --> is a part of --> compensation --> involves --> octopus --> is --> skin --> is found in --> collagen --> Both contribute to overall mechanical properties --> biological materials --> combine --> synthetic protein materials --> promotes --> sustainability

flavor --> taste --> is a part of --> compensation --> involves --> octopus --> is --> skin --> is found in --> collagen --> Interacts --> hydroxyapatite (ha) --> is --> biocompatible --> utilize --> green flexible electronics --> utilize --> sustainable --> sustainability
\end{lstlisting}
\end{LLMbox}
\end{figure}
Since the augmented graph features detailed representations about food, proteins, and related aspects, several new pathways are identified that can indeed be traced back to the graph derived from the newly generated data based on adversarial X-LoRA sampling. Generally, the larger the added graph and the more intricate overlap there exists with the original graph, the more integrated the results become. 
Building on this analysis, we provide a visual representation of these paths (this analysis is similar to the one reported above, see Figure~\ref{fig:Fig_10}). Figure~\ref{fig:Fig_7}e-f depicts the results for both cases where we retain the unique paths and an integrated graph in which the nodes are merged. 
For graph reasoning, the question asked is:
\begin{figure}[H]
\begin{LLMbox}{} 
\begin{lstlisting}
Carefully read the paths and summarize scientific insights in several bullet points. Then be creative and propose new research ideas. Think step by step.
\end{lstlisting}
\end{LLMbox}
\end{figure}

The answer provided by X-LoRA is shown in Text Box~\ref{TexBox:TB20}. The key insight is that synthetic protein materials can promote sustainability by mimicking flavor properties. A notable idea proposed is the use of green flexible electronics in flavor-related applications, such as flavor delivery systems, and comparing their sustainability with traditional bottom-up or top-down approaches. The model also  suggests to explore interactions between hydroxyapatite  and collagen for use in green flexible electronics. This could complement more traditional approaches such as using starch and thereby introduce a greater design space. It is clearly evident that graph reasoning goes far beyond the ideas that were developed by the agents alone, as was shown in Table~\ref{tab:edge_mapping_reasoning_labels}, and includes more innovative and specific concepts and a more focused exploration of how these materials and technologies could intersect to promote sustainability through interdisciplinary research. 

\begin{figure}[htbp]
\begin{LLMbox}{\XLora} %BioMixtral XLora 
\begin{lstlisting}
The primary combination highlights the relationship between flavor, flavor properties, synthetic protein materials, and sustainability. It suggests that synthetic protein materials can promote sustainability by mimicking flavor properties and promoting sustainability.

-The alternative combination from 0 to 1 focuses on the use of green flexible electronics and their sustainability. It highlights the lower cost of materials and techniques used in self-assembly, which makes green flexible electronics a more sustainable option compared to traditional bottom-up or top-down approaches.

-The alternative combination from 1 to 0 emphasizes the role of collagen in promoting sustainability. Both collagen and synthetic protein materials contribute to overall mechanical properties, and their combination can promote sustainability.

-The alternative combination from 1 to 1 highlights the interaction between hydroxyapatite (HA) and collagen in promoting sustainability. HA is biocompatible and can be utilized in green flexible electronics, making them a more sustainable option.

Based on these insights, new research ideas could include:

-Developing new synthetic protein materials that mimic flavor properties and promote sustainability.

-Investigating the use of green flexible electronics in flavor-related applications, such as flavor delivery systems, and comparing their sustainability with traditional bottom-up or top-down approaches.

-Studying the role of collagen in promoting sustainability in various applications, such as food packaging or biomedical devices.

-Exploring the interaction between HA and collagen in promoting sustainability in green flexible electronics and other applications.
\end{lstlisting}
\end{LLMbox}
\textboxcaption{Response of X-LoRA when tasked to carefully read the paths extracted from the graph, and to summarize scientific insights in several bullet points. This responses exploit multiple paths in a knowledge graph connecting the terms 'flavor' to 'sustainability', based on the agentic modeling of a conversation between a chef and an inventor. }
\label{TexBox:TB20}
\end{figure}

\subsection{Incorporation of new data from scientific literature towards the design of sustainable mycelium composite materials}
In this section we discuss an example where new papers from the scientific literature was added that were not previously incorporated. This is driven by a particular design objective, to develop sustainable mycelium  materials. 
To do this we include three additional papers that discuss topics around mycelium as networks (from a basic biology perspective) and the use of mycelium as 3D printed materials (from a mechanics and materials perspective)~\cite{Fricker2017TheNetwork,Fricker2017TheNetworkb,Shen2024RobustMaterials}. The choice of these papers is driven by our desire to form a corpus that yields an interdisciplinary set of ideas, and which will likely yield  overlapping concepts that can then be integrated into the larger graph. For instance,  reference~\cite{Shen2024RobustMaterials} discusses mycelium and its use as a material and hence provides a bridge to the more biology-focused papers~\cite{Fricker2017TheNetwork,Fricker2017TheNetworkb}. We follow the same approach as used above and first generate triples from the new text, then merge these into the original graph. Since we are interested in exploiting paths of knowledge in this analysis we only consider the giant component of the merged graph that now features the original structure but also new information extracted from the set of three mycelium papers. The use of the giant component  can generally be a fruitful approach when interested in graph reasoning over paths as it ensures connectivity across the entire system. 

\begin{figure}
    \centering
    \includegraphics[width=0.7\linewidth]{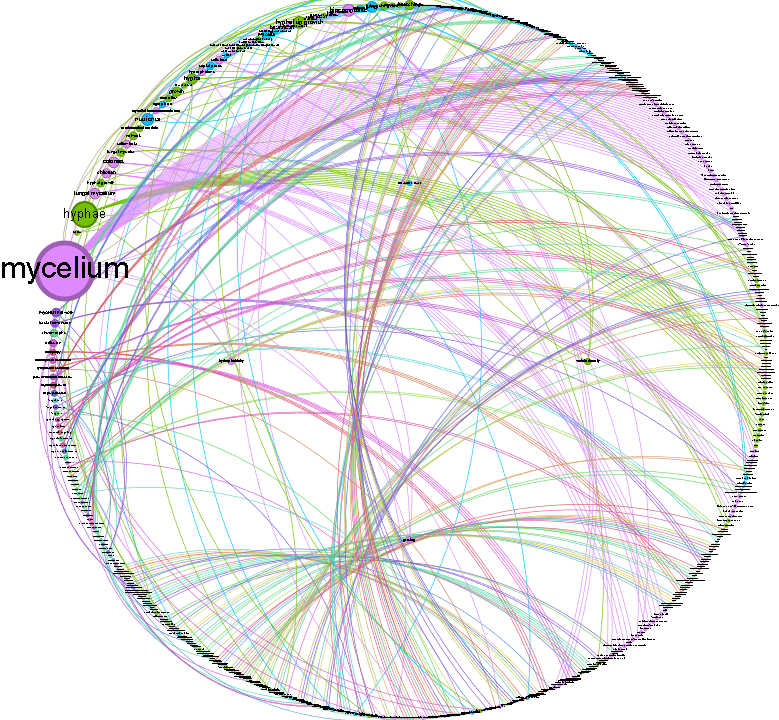}
    \caption{Newly created graph from three papers, shown in circular layout (node sizes according to node degree, and node color according to group). There are a total of 501 nodes and 375 edges with an average node degree of 1.5 and 143 communities; the giant component includes 155 nodes and 162 edges with an average node degree of 2.1 and 10 communities.}
    \label{fig:Fig_12_A}
\end{figure}

\begin{figure}
    \centering
    \includegraphics[width=.85\linewidth]{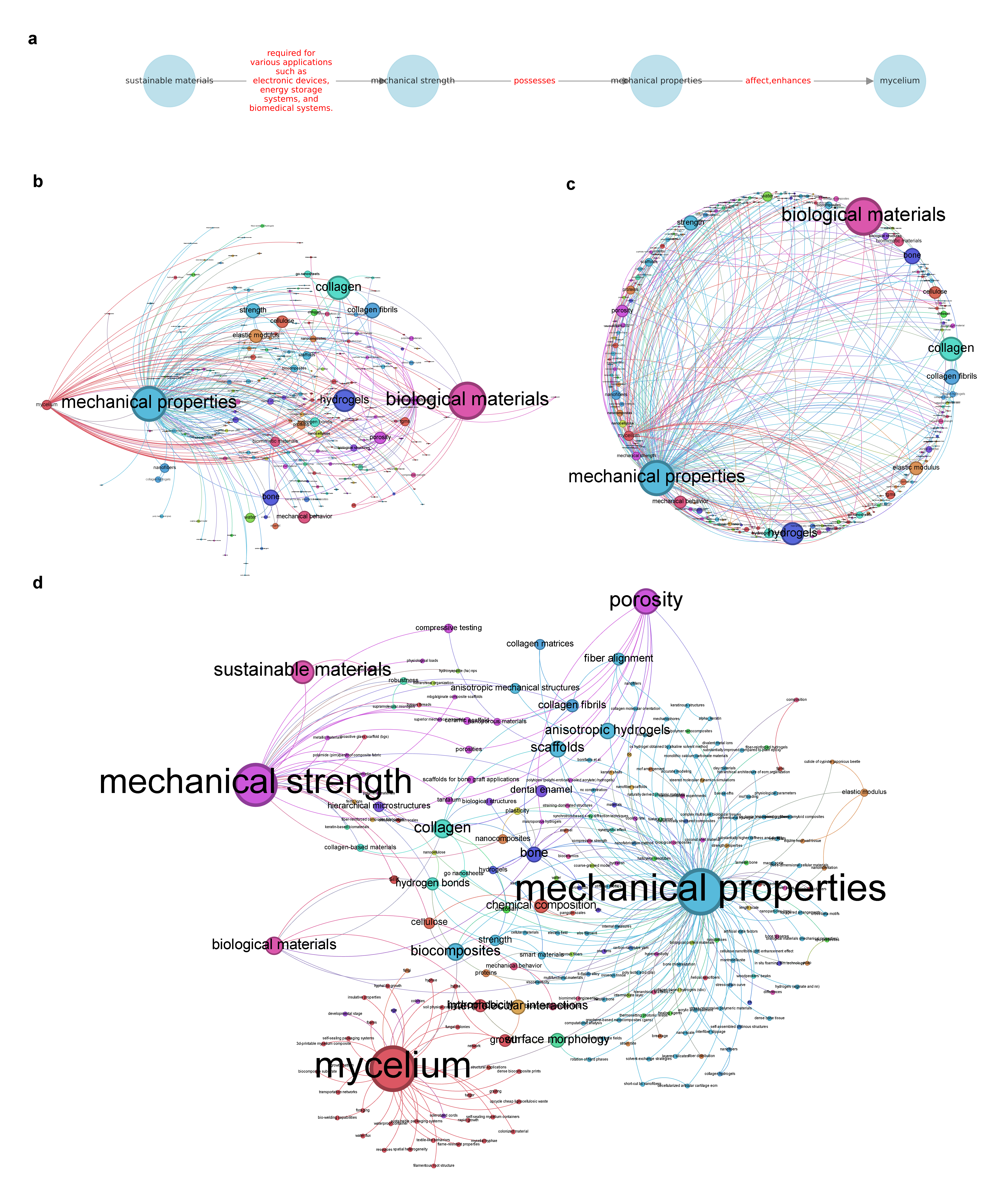}
    \caption{Result of integrating the newly generated graph (see, Figure~\label{fig:Fig_12_A}) graph into the large graph, and conducting a path query in this new structure. The query is to relate `sustainable materials' to `mycelium'. The shortest path between these nodes is depicted in panel a. Panel b shows the resulting sub-graph extracted from this path (this is done by selecting all nodes in this path as well as their nearest neighbors). Panel c shows the same graph, but in circular layout. In both panels b and c the original group identification and node degree is shown. In panel d, we show the same graph but determine the size of each node based on its bridging centrality, and the color by group identification (newly generated for the subgraph). }
    \label{fig:Fig_12}
\end{figure}

Figure~\ref{fig:Fig_12_A} shows the newly created graph from the three papers, visualized  in circular layout (node sizes according to node degree, and node color according to group). Further, a detailed statistical analysis of the graph properties is shown in Figure~\ref{fig:Fig_S1_mycelaugsub}.
Next, Figure~\ref{fig:Fig_12} reveals the result of integrating the newly generated graph into the original, large graph, and conducting a path query in this new structure. The query is to relate `sustainable materials' to `mycelium', reflecting our original goal of the inquiry. 

The shortest path between these nodes is depicted in Figure~\ref{fig:Fig_12}a. Related, Figure~\ref{fig:Fig_12}b shows the resulting sub-graph extracted from this path (this is done by selecting all nodes in this path as well as their nearest neighbors). Figure~\ref{fig:Fig_12}c shows the same graph, but in circular layout. In both panels Figure~\ref{fig:Fig_12}b and c the original group identification and node degree is shown. Bridging centrality measures the importance of a node within a network based on its position as a `bridge' between different parts of the graph. The idea behind this metric is that certain nodes play a critical role in connecting otherwise separate or distant regions within a network, facilitating communication and interaction between them. 

In Figure~\ref{fig:Fig_12}d, we show graph using this measure, where the size of each node based on its bridging centrality, and the color by group identification, here newly generated for the subgraph~\cite{Girvan2002CommunityNetworks}. The analysis in Figure~\ref{fig:Fig_12}d is particularly notable since it reveals key components of the graph that resemble `bridging' character. These are, among others, `mechanical properties', `mechanical strength', `mycelium', `sustainable materials', `porosity', `collagen', and others. 
Nodes with high bridging centrality are likely strategic connectors, that is, nodes that serve as bridges between different clusters or communities within a network. 
The concept underlying bridging centrality aligns with the strength of weak ties theory in social networks\cite{Granovetter1973TheTies}, which suggests that weaker ties (often represented by nodes with high bridging centrality) are more valuable in spreading information or innovations across diverse groups. In the context of conceptual relationships, nodes with high bridging centrality may represent concepts that link disparate areas of knowledge, facilitating cross-disciplinary innovation and understanding.
Enhancing or strengthening connections to these bridging nodes could significantly improve the cohesion and flow of knowledge or information across the network. We find this to be a good strategy to specifically identify research opportunities through graph reasoning.

We now conduct two experiments: First, using the entire subgraph and second, using a subgraph derived from the subgraph that features a set of select nodes with the highest bridging centrality. 
We first use the entire newly generated subgraph (Figure~\ref{fig:Fig_12}) and use it as context to answer a question. In this case, unlike before where we provided various ranked shortest-path options, here we provide the graph in its entirety to the model. The task given is:
\begin{figure}[H]
\begin{LLMbox}{} 
\begin{lstlisting}
Be creative and synthesize new research ideas to build sustainable mycelium materials. Use all context to develop new directions, and provide a detailed answer.
\end{lstlisting}
\end{LLMbox}
\end{figure}

The response from X-LoRA is shown in Text Box~\ref{TexBox:TB30}. A key insight is the use of nanocellulose in mycelium materials. This idea is based on the finding that nanocellulose can break in the presence of water due to hydrogen bonds, but when combined with graphene oxide nanosheets, it can lead to higher tensile strength and toughness. The model further suggests to investigate the role of intermolecular interactions in mycelium growth and how it affects surface morphology. The model suggests that the interplay between intermolecular interactions and surface morphology may allow us to develop strategies to manipulate these factors. Self-sealing packaging systems have already been proposed, but the model suggests that this is a particularly interesting avenue for future developments. Perhaps the most critical insight is that hierarchical organization can be used as a way to increase robustness in mycelium materials, along with a novel combination of nanocellulose along with graphene oxide nanosheets. This is proposed based on the concept that higher hierarchical organization tend to be more robust, opening possible ways to manipulate the hierarchical organization of mycelium materials to improve their robustness and sustainability through these specific material mechanisms.

\begin{figure}[htbp]
\begin{LLMbox}{\XLora} %BioMixtral XLora 
\begin{lstlisting}
One potential research idea could be to explore the use of nanocellulose in mycelium materials to improve their mechanical properties. The knowledge graph suggests that nanocellulose can break in the presence of water due to hydrogen bonds, but when combined with go nanosheets, it can lead to higher tensile strength and toughness. By incorporating nanocellulose into mycelium materials, researchers could potentially enhance their mechanical properties and make them more sustainable.

Another idea could be to investigate the role of intermolecular interactions in mycelium growth and how it affects surface morphology. The knowledge graph indicates that intermolecular interactions are crucial for growth, and surface morphology can impact hydrophobicity. By understanding the interplay between intermolecular interactions and surface morphology, researchers could develop strategies to manipulate these factors and create more sustainable mycelium materials.

A third research idea could be to explore the potential of mycelium materials for creating self-sealing packaging systems. The knowledge graph suggests that mycelium can create waterproof containers and may provide an opportunity to produce sustainable packaging systems. By further developing this technology, researchers could create more eco-friendly alternatives to traditional packaging materials.

Finally, researchers could investigate the relationship between hierarchical organization and robustness in mycelium materials. The knowledge graph indicates that the degree of hierarchical organization determines the degree of robustness, and materials with higher hierarchical organization tend to be more robust. By understanding this relationship, researchers could develop strategies to manipulate the hierarchical organization of mycelium materials to improve their robustness and sustainability.
\end{lstlisting}
\end{LLMbox}
\textboxcaption{Response of X-LoRA when tasked to be creative and synthesize new research ideas to build sustainable mycelium materials. the model is further asked to Use all context to develop new directions, and to provide a detailed answer. }
\label{TexBox:TB30}
\end{figure}

To gain further insights, the human user then follows up and ask X-LoRA (under the same context of the knowledge graph):
\begin{figure}[H]
\begin{LLMbox}{} 
\begin{lstlisting}
Provide specific chemical strategies that can be used to control the interplay between intermolecular interactions and surface morphology in mycelium materials. Specifically focus on atomistic mechanisms and predicted behaviors. List candidate materials or chemistries that can be used, and speculate what type of mechanisms you would expect. 
\end{lstlisting}
\end{LLMbox}
\end{figure}

The response is shown in Text Box~\ref{TexBox:TB31}. The detailed ideas for chemical modifications and expected behaviors are striking, in particular because they provide quite specific, and deep, insights. An important point made is that the interplay between intermolecular interactions and surface morphology in mycelium materials can be controlled through the use of functional groups that can influence the hydrogen bonding interactions between mycelium fibers (e.g., carboxylic acid groups can enhance hydrogen bonding, while amine groups can weaken it, but the model also explores the use of nanoscale surface roughness as a means to achieve this). The model suggests that this can be achieved by altering the mycelium fibers' surface chemistry (during or after growth, correctly understanding that both living and non-living mycelium can be utilized in this context). An alternative strategy proposed is the use of supramolecular assemblies, such as cyclodextrins or host-guest interactions, to control the interplay between intermolecular interactions and surface morphology (the latter to affect the material's surface energy and wettability). This integration of chemical functionalization and nano-texturing is a particularly intriguing design principle, especially when combined with the fact that these provide high levels of control of interactions within the composite. The model also suggests that the use of supramolecular assemblies in particular can provide access to novel thermal properties, such as glass transition temperature and crystallinity effects driven by these additives.

\begin{figure}[htbp]
\begin{LLMbox}{\XLora} %BioMixtral XLora 
\begin{lstlisting}
One chemical strategy to control the interplay between intermolecular interactions and surface morphology in mycelium materials is through the use of functional groups that can influence the hydrogen bonding interactions between mycelium fibers. For example, carboxylic acid groups can enhance hydrogen bonding, while amine groups can weaken it. This can be achieved through the modification of the mycelium fibers' surface chemistry, such as through the use of functionalized molecules or the incorporation of functional groups during the growth of the mycelium. nanoscale surface roughness can also be used to control surface morphology and intermolecular interactions. By creating nanoscale features on the surface of the mycelium fibers, the hydrogen bonding interactions between the fibers can be modulated, leading to changes in the material's properties.

Another chemical strategy is the use of supramolecular assemblies, such as cyclodextrins or host-guest interactions, to control the interplay between intermolecular interactions and surface morphology. These assemblies can be incorporated into the mycelium fibers, either during growth or through post-processing treatments, to modulate the hydrogen bonding interactions and surface morphology.

In terms of predicted behaviors, the use of functional groups to control intermolecular interactions and surface morphology in mycelium materials is expected to lead to changes in the material's mechanical properties, such as tensile strength and toughness. The incorporation of supramolecular assemblies is expected to lead to changes in the material's thermal properties, such as glass transition temperature and crystallinity. The use of nanoscale surface roughness is expected to lead to changes in the material's surface energy and wettability.

Candidate materials or chemistries that can be used include functionalized molecules, such as carboxylic acid- or amine-functionalized molecules, as well as supramolecular assemblies, such as cyclodextrins or host-guest interactions. The specific mechanism of action will depend on the specific chemical strategy being employed and the desired outcome.

Overall, the use of these chemical strategies can provide a powerful tool for controlling the interplay between intermolecular interactions and surface morphology in mycelium materials, leading to the development of novel materials with unique properties and applications.
\end{lstlisting}
\end{LLMbox}
\textboxcaption{Response of X-LoRA when tasked to provide specific chemical strategies that can be used to control the interplay between intermolecular interactions and surface morphology in mycelium materials. The model is further tasked to specifically focus on atomistic mechanisms and predicted behaviors and to list candidate materials or chemistries that can be used, and speculate what type of mechanisms you would expect.}
\label{TexBox:TB31}
\end{figure}

The BioinspiredLLM-Mixtral model responds to the original question as shown in Text Box~\ref{TexBox:TB32}. The model suggests applications across a variety of domains, such as electronic devices, energy storage systems, and biomedical systems by leveraging the unique properties of mycelium, such as its ability to grow and adapt to its environment, its high mechanical strength, and its potential for biodegradability. A unique aspect of this response is that different types of bone can be used as a means to enhance the mechanical properties of mycelium-based materials. This may result in uses such as bone grafts or tissue engineering scaffolds. Additionally, the use of nanocomposites and nanofillers is suggested as a method to further enhance the mechanical properties of mycelium-based materials by improving their stiffness, toughness, and resistance to degradation. Other design features include different levels of hierarchy, similar to the earlier response from X-LoRA, and the use of nanofillers.

\begin{figure}[htbp]
\begin{LLMbox}{\BioMixtral} %BioMixtral XLora 
\begin{lstlisting}
Based on the provided knowledge graph, there are several research ideas to build sustainable mycelium materials. One approach is to explore the potential of mycelium as a sustainable material for various applications such as electronic devices, energy storage systems, and biomedical systems. This can be achieved by leveraging the unique properties of mycelium, such as its ability to grow and adapt to its environment, its high mechanical strength, and its potential for biodegradability.

Another research idea is to investigate the effects of different types of bone on the mechanical properties of mycelium-based materials. This can provide insights into how to optimize the mechanical properties of mycelium-based materials for specific applications, such as bone grafts or tissue engineering scaffolds. Additionally, the use of nanocomposites and nanofillers can further enhance the mechanical properties of mycelium-based materials by improving their stiffness, toughness, and resistance to degradation.

Furthermore, the study of hierarchical microstructures in biological materials can provide valuable insights into how to design sustainable mycelium materials with improved mechanical properties. By understanding how different levels of hierarchy affect the mechanical behavior of biological tissues, researchers can develop new strategies for designing mycelium-based materials with tailored mechanical properties.

Finally, the use of bioinspired approaches can also be an effective way to build sustainable mycelium materials with improved mechanical properties. By studying the mechanisms underlying the mechanical behavior of biological tissues, researchers can develop new bioinspired materials with unique properties and functionalities. For example, the use of hierarchical microstructures, nanocomposites, and nanofillers can be combined with bioinspired approaches to create new mycelium-based materials with enhanced mechanical properties.
\end{lstlisting}
\end{LLMbox}
\textboxcaption{Response of BioinspiredLLM-Mixtral when tasked to be creative and synthesize new research ideas to build sustainable mycelium materials. The model is further asked to Use all context to develop new directions, and to provide a detailed answer.}
\label{TexBox:TB32}
\end{figure}

And finally, the response from GPT-4 is depicted in Text Box~\ref{TexBox:TB33}. A key design feature is the incorporation of different materials into mycelium-based nanocomposites that contain collagen or other strength-enhancing nanofillers. The model predicts that these create hybrids that exhibit substantially higher stiffness and ductility.
Another key design feature identified is the use of porosity. Leveraging porosity control techniques used in macroporous hydrogels, the model suggests that similar strategies can be applied to mycelium to create scaffolds with tailored porosity. These scaffolds could mimic the natural bone structure more closely, promoting better integration and faster healing in bone graft applications. Other ideas include the use of mineral fillers or other sustainable additives like cellulose or chitosan that can be used to construct packaging materials that are not only biodegradable but also possess the necessary strength and durability for commercial use. Specifically, the use of nanofillers like graphene or metal-organic frameworks that respond to external stimuli could yield health monitoring systems, where the material itself can detect changes in stress or damage, alerting for maintenance before failure occurs. 
Utilizing the mechanical strength and hydrophobic properties of mycelium, the model also predicts potential avenues to develop innovative filtration systems for water treatment, exploiting the balance of high mechanical strength and hydrophobic properties. These systems are predicted to effectively filter out contaminants while being robust and durable under various environmental conditions.
Overall, a rich set of applications is outlined ranging from dental restorations mimicking enamel, to packaging, electronics, to structural materials.

\begin{figure}[htbp]
\begin{LLMbox}{\VG} %BioMixtral XLora 
\begin{lstlisting}
Given the intricate relationships between materials, structure, and properties as outlined in the knowledge graph, we can derive several innovative research ideas focusing on the development of sustainable mycelium materials. The graph highlights the importance of mechanical properties, the role of nanocomposites, the influence of porosity, and the potential of biocomposites, among other factors. Integrating these insights, we can propose the following research directions:

Development of Mycelium-Nanocomposite Hybrids for Enhanced Mechanical Strength: The graph indicates that nanocomposites can significantly improve mechanical strength and contain materials like collagen, which is crucial for biological materials. By incorporating nanocomposites that contain collagen or other strength-enhancing nanofillers into mycelium-based materials, we can create hybrids that exhibit substantially higher stiffness and ductility. These materials could be ideal for applications requiring robust yet lightweight materials, such as in automotive or aerospace industries.

Porosity-Controlled Mycelium Scaffolds for Bone Graft Applications: Porosity is strongly related to mechanical properties and is crucial in scaffolds for bone graft applications. Leveraging the porosity control techniques used in macroporous hydrogels, similar strategies can be applied to mycelium to create scaffolds with tailored porosity. These scaffolds could mimic the natural bone structure more closely, promoting better integration and faster healing in bone graft applications.

Biocomposite Mycelium Materials with Improved Mechanical Properties for Packaging: The graph suggests that biocomposites can have improved mechanical properties through the addition of mineral fillers and are in contact with mycelium. Developing biocomposite materials that integrate mycelium with mineral fillers or other sustainable additives like cellulose or chitosan could lead to packaging materials that are not only biodegradable but also possess the necessary strength and durability for commercial use.

Self-Sealing Mycelium Containers with Hydrophobic Properties: Given that mycelium imbues hydrophobicity and can create waterproof containers, there's an opportunity to develop self-sealing mycelium containers for use in sensitive packaging solutions, such as for electronics or perishable goods. These containers could leverage the bio-welding capabilities of mycelium to create airtight seals, offering an innovative and sustainable alternative to traditional packaging materials.

Smart Mycelium Materials with Embedded Nanofillers for Structural Health Monitoring: Integrating the concept of smart materials with mycelium, we can explore the development of mycelium-based materials embedded with nanofillers like graphene or metal-organic frameworks (MOFs) that respond to external stimuli. These materials could be used in structural health monitoring systems, where the material itself can detect changes in stress or damage, alerting for maintenance before failure occurs.

Hybrid Mycelium Materials for Improved Dental Applications: Considering the mechanical properties of dental enamel and the potential for mycelium in creating hierarchical structures, a novel research direction could involve developing hybrid materials that combine mycelium with mineral components mimicking enamel. These materials could offer a sustainable and potentially more biocompatible alternative for dental restorations or implants.

Mycelium-Based Filtration Systems with Enhanced Mechanical and Hydrophobic Properties: Utilizing the mechanical strength and hydrophobic properties of mycelium, there's potential to develop innovative filtration systems for water treatment. These systems could leverage the structural and material properties of mycelium to filter out contaminants while being robust and durable under various environmental conditions.

Each of these ideas integrates multiple aspects of the knowledge graph, focusing on leveraging the unique properties of mycelium in combination with other materials to create innovative, sustainable solutions across a range of applications. Further research and development in these directions could lead to significant advancements in material science, particularly in the context of sustainability and environmental conservation.
\end{lstlisting}
\end{LLMbox}
\textboxcaption{Response of GPT-4 when tasked to be creative and synthesize new research ideas to build sustainable mycelium materials. The model is further asked to Use all context to develop new directions, and to provide a detailed answer.}
\label{TexBox:TB33}
\end{figure}
 
One interesting theme in the teh GPT-4 response is the proposal of the development of mycelium-collagen composites. Like most of the designs discussed here and in the preceding sections, to the best of our knowledge, this has not been discussed in the literature so far, indicating a new material design identified by the model.
To gain further insights into this particular new idea, we follow up and ask the model: 
\begin{figure}[H]
\begin{LLMbox}{User} 
\begin{lstlisting}
Discuss technical details of hybrid mycelium-collagen composites, specifically focused on molecular structure, interactions and predicted behaviors. Speculate what type of mechanisms you would expect.
\end{lstlisting}
\end{LLMbox}
\end{figure}

The response of GPT-4 is shown in Text Box~\ref{TexBox:TB35}. The model predicts that in mycelium-collagen composites, interactions are likely to occur at the molecular and fibrillar levels. These interactions could involve hydrogen bonding between the hydroxyl groups of chitin and the hydroxyproline in collagen. Additionally, van der Waals forces and electrostatic interactions could play roles in stabilizing the composite structure. The compatibility between mycelium's chitin structure and collagen's triple helix could facilitate a coherent matrix, enhancing the composite's mechanical properties. The model predicts a highly interesting set of features, such as mechanical reinforcement, biodegrability, complex porosity npatters, and self-healing features. Mycelium's ability to grow and self-repair could impart self-healing properties to the composite. Upon damage, mycelium could potentially recolonize and repair the composite matrix, a mechanism that would be particularly useful in long-term, sustainable material applications.
The model also shows a good understanding of the interplay of the proposed constituents, specifically stating that collagen's tensile strength could complement the rigidity provided by mycelium's chitin. This is proposed to lead to a material with enhanced stiffness, strength, and toughness compared to the individual components.
Further, the degradation rate could be controlled by the ratio of mycelium to collagen, the degree of crosslinking, and the environmental conditions, offering tunable properties for specific applications.
The inherent porosity of mycelium and the fibrillar nature of collagen could result in a composite with a highly interconnected pore structure that could be beneficial for applications requiring fluid transport, cell migration, and nutrient diffusion, such as tissue engineering scaffolds.  
Given the biocompatibility of both mycelium and collagen, the composite is expected to be non-toxic and supportive of cell attachment and proliferation. The presence of collagen, in particular, could enhance cell-material interactions due to its role in the extracellular matrix of tissues. 
Since we tasked the model to also explore more speculative mechanisms, several interesting predictions are made. These include improved stress distributions due to the synergistic structural organization of mycelium and collagen, moisture regulation to enhance durability and stability in various environments, and enhanced biofunctionality (e.g. bone regeneration or anti-inflammatory responses).

\begin{figure}[htbp]
\begin{LLMbox}{\VG} %BioMixtral XLora 
\begin{lstlisting}
Hybrid mycelium-collagen composites represent a novel class of biomaterials that leverage the unique properties of both mycelium, the vegetative part of fungi composed of a network of hyphae and collagen, the most abundant protein in the animal kingdom known for its structural roles in connective tissues. The development of such composites is predicated on understanding the molecular structure and interactions between mycelium and collagen, as well as predicting their behaviors and mechanisms when combined. Here, we delve into these aspects, focusing on technical details and speculative mechanisms expected from hybrid mycelium-collagen composites.

Molecular Structure and Interactions

Mycelium Molecular Structure: Mycelium consists of a network of hyphae, which are tubular structures made of chitin and other polysaccharides. Chitin, a long-chain polymer of N-acetylglucosamine, is a key structural component providing rigidity and strength. Mycelium's structure is highly porous and allows for rapid growth and adaptation to its environment.

Collagen Molecular Structure: Collagen is a protein composed of three polypeptide chains, forming a triple helix. This structure provides high tensile strength and elasticity. Collagen's primary amino acids are glycine, proline, and hydroxyproline, which contribute to its unique triple-helical structure. Collagen molecules can assemble into fibrils and fibers, contributing to the mechanical properties of tissues.

Interactions:  In hybrid mycelium-collagen composites, interactions are likely to occur at the molecular and fibrillar levels. These interactions could involve hydrogen bonding between the hydroxyl groups of chitin and the hydroxyproline in collagen. Additionally, van der Waals forces and electrostatic interactions could play roles in stabilizing the composite structure. The compatibility between mycelium's chitin structure and collagen's triple helix could facilitate a coherent matrix, enhancing the composite's mechanical properties.

Predicted Behaviors and Mechanisms

Mechanical Reinforcement: The structural integrity of the composite is expected to benefit from the mechanical properties of both components. Collagen's tensile strength could complement the rigidity provided by mycelium's chitin, leading to a material with enhanced stiffness, strength, and toughness compared to the individual components.

Biodegradability and Environmental Interaction: Both mycelium and collagen are biodegradable, making the composite environmentally friendly. The degradation rate could be controlled by the ratio of mycelium to collagen, the degree of crosslinking, and the environmental conditions, offering tunable properties for specific applications.

Hybrid Composite Porosity: The inherent porosity of mycelium and the fibrillar nature of collagen could result in a composite with a highly interconnected pore structure. This porosity would be beneficial for applications requiring fluid transport, cell migration, and nutrient diffusion, such as tissue engineering scaffolds.

Biocompatibility and Bioactivity: Given the biocompatibility of both mycelium and collagen, the composite is expected to be non-toxic and supportive of cell attachment and proliferation. The presence of collagen, in particular, could enhance cell-material interactions due to its role in the extracellular matrix of tissues.

Self-Healing and Adaptability: Mycelium's ability to grow and self-repair could impart self-healing properties to the composite. Upon damage, mycelium could potentially recolonize and repair the composite matrix, a mechanism that would be particularly useful in long-term, sustainable material applications.

Speculative Mechanisms:

Stress Distribution and Dissipation: The hybrid composite could distribute and dissipate mechanical stresses more effectively than its individual components, attributed to the synergistic structural organization of mycelium and collagen.

Moisture Regulation: The porous structure could regulate moisture content, enhancing durability and stability in various environments.

Enhanced Biofunctionality: By functionalizing the composite with bioactive molecules, one could introduce specific biological functions, such as promoting bone regeneration or anti-inflammatory responses.

In summary, hybrid mycelium-collagen composites offer exciting prospects for developing sustainable, biocompatible, and mechanically robust materials. The molecular structure and interactions between mycelium and collagen pave the way for a range of behaviors and mechanisms, from enhanced mechanical properties to biodegradability and self-healing capabilities. Further research and development in this area could lead to innovative solutions in biomedical engineering, environmental sustainability, and beyond.
\end{lstlisting}
\end{LLMbox}
\textboxcaption{Response of GPT-4 when tasked to discuss technical details of hybrid mycelium-collagen composites, specifically focused on molecular structure, interactions and predicted behaviors and mechanisms. }
\label{TexBox:TB35}
\end{figure}

Next we task the model to reason over graphs derived and filtered to focus on highest bridging centrality nodes, which offers strong potential to create new research ideas.  We focus on a subgraph derived from the subgraph shown in Figure~\ref{fig:Fig_12}d as context to answer a question. We select the $\rm top_N=20$ nodes with the highest bridging centrality, as they form strategic connectors as bridges between different clusters or communities.
We ask the same query as before (but not the graph data provided is different):
\begin{figure}[H]
\begin{LLMbox}{} 
\begin{lstlisting}
Be creative and synthesize new research ideas to build sustainable mycelium materials. Use all context to develop new directions, and provide a detailed answer.
\end{lstlisting}
\end{LLMbox}
\end{figure}

The X-LoRA model responds as follows visualized in Text Box~\ref{TexBox:TB40}. The model specifically singles out studying the influence of intermolecular interactions, the role of hydrophobicity, fiber alignment, and surface morphology on the mechanical strength of these materials as key objectives. 
Other important opportunities for research are identified for biocomposites made from mycelium and collagen or hydrogen bond-mediated systems. A specific question posed is the study of the impact of mechanical properties, porosity, and hydrogen bonds (and how they can be manipulated to enhance mechanical properties and functionalities such as hydrophobicity and surface morphology) on the overall properties and behavior of these material. 
A notable feature identified by the model is the focus on self-assembly of tropocollagen molecules as a foundational framework that can interact in interesting ways with the growth of mycelium and specifically considering issues such as resulting fiber alignments. 
These research opportunities are associated with applications as diverse as electronic devices, energy storage systems, and biomedical systems.

\begin{figure}[htbp]
\begin{LLMbox}{\XLora} %BioMixtral XLora 
\begin{lstlisting}
Based on the given graph and the information provided, we can identify several nodes with high bridging centrality that could be potential targets for enhancing the cohesion and flow of knowledge across the network. These nodes include mechanical properties, mycelium, mechanical strength, porosity, collagen, biocomposites, scaffolds, hydrogen bonds, fiber alignment, dental enamel, hydrophobicity, and surface morphology.

To develop new research ideas to build sustainable mycelium materials, we can focus on the following aspects:

Investigating the mechanical properties of mycelium-based materials and understanding how they can be enhanced or improved. This could involve studying the influence of intermolecular interactions, fiber alignment, and surface morphology on the mechanical strength of these materials.

Exploring the potential of biocomposites made from mycelium and other sustainable materials, such as collagen or hydrogen bonds. This could involve investigating the impact of mechanical properties, porosity, and hydrogen bonds on the overall properties and behavior of these materials.

Developing scaffolds using mycelium and collagen, and studying their growth and hydrophobicity. This could involve understanding the role of intermolecular interactions and surface morphology in facilitating growth and enhancing hydrophobicity.

Investigating the mechanical strength of mycelium-based materials and how it can be improved through the self-assembly of tropocollagen molecules. This could involve understanding the relationship between mechanical strength, collagen fibrils, and fiber alignment.

Studying the influence of porosity on the mechanical properties and behavior of mycelium-based materials, and how this can be optimized for specific applications such as electronic devices, energy storage systems, and biomedical systems.

Investigating the role of hydrogen bonds in mycelium-based materials and how they can be manipulated to enhance mechanical properties and functionalities such as hydrophobicity and surface morphology.

By focusing on these aspects and leveraging the knowledge gained from the given graph and context, we can develop new research ideas to build sustainable mycelium materials that are both functional and environmentally friendly.
\end{lstlisting}
\end{LLMbox}
\textboxcaption{Response of X-LoRA model when tasked to creatively synthesize new research ideas to build sustainable mycelium materials, for a graph extracted based onthe $\rm top_N=20$ nodes with the highest bridging centrality, as they form strategic connectors as bridges between different clusters or communities.}
\label{TexBox:TB40}
\end{figure}

The BioinspiredLLM-Mixtral model responds as shown in Text Box~\ref{TexBox:TB41}. This model suggests a study of the relationships between mechanical properties and porosity in biological materials, such as bone and collagen fibrils as foundation for sustainable mycelium materials. This model also suggests the study of hydrogen bonds in enhancing mechanical strength and hydrophobicity as a means to achieve better functionality and control. The model further suggests an investigation of the relationship between mechanical properties and fiber alignment in collagen fibrils, similar to the ideas suggested earlier,

\begin{figure}[htbp]
\begin{LLMbox}{\BioMixtral} %BioMixtral XLora 
\begin{lstlisting}
Based on the provided data, we can identify several key concepts and relationships in this knowledge graph that could be used to develop new research ideas for building sustainable mycelium materials. 

Here are some potential directions:

1. Explore the relationship between mechanical properties and porosity in biological materials, such as bone and collagen fibrils. This could involve studying how changes in porosity affect the mechanical strength of these materials and how they can be optimized to improve their properties.

2. Investigate the role of hydrogen bonds in enhancing mechanical strength and hydrophobicity in biological materials. This could involve exploring how hydrogen bonding affects intermolecular interactions and surface morphology in collagen fibrils and other materials.

3. Examine the potential of mycelium-based biocomposites to improve mechanical properties in sustainable materials. This could involve studying how mycelium can be used to imbue hydrophobicity and enhance growth in biocomposites, and how this can be optimized to improve their properties for various applications such as electronic devices, energy storage systems, and biomedical systems.

4. Investigate the relationship between mechanical properties and fiber alignment in collagen fibrils and other materials. This could involve studying how fiber alignment affects mechanical strength and hydrophobicity in these materials and how it can be optimized to improve their properties.

5. Explore the potential of biocomposites to improve mechanical properties in sustainable materials by adding mineral fillers. This could involve studying how mineral fillers can be used to enhance mechanical strength and hydrophobicity in biocomposites and how this can be optimized to improve their properties for various applications such as electronic devices, energy storage systems, and biomedical systems.

6. Investigate the role of intermolecular interactions in enhancing growth in mycelium-based biocomposites. This could involve studying how intermolecular interactions affect hydrophobicity and surface morphology in mycelium-based biocomposites and how they can be optimized to improve their properties for various applications such as electronic devices, energy storage systems, and biomedical systems.

7. Explore the potential of collagen fibrils to improve mechanical properties in sustainable materials by studying their properties and behavior in bone tissue. This could involve studying how collagen fibrils contribute to overall mechanical properties in bone tissue and how they can be optimized to improve their properties for various applications such as electronic devices, energy storage systems, and biomedical systems.
\end{lstlisting}
\end{LLMbox}
\textboxcaption{Response of the BioinspiredLLM-Mixtral model when tasked to creatively synthesize new research ideas to build sustainable mycelium materials, for a graph extracted based on the $\rm top_N=20$ nodes with the highest bridging centrality, as they form strategic connectors as bridges between different clusters or communities.}
\label{TexBox:TB41}
\end{figure}

The response of GPT-4 is shown in Text Box~\ref{TexBox:TB42}. The model proposes a variety of ideas, including the use of mineral fillers as a way to tune the mechanical properties of mycelium-based biocomposites. Given that biocomposites can have improved mechanical properties due to the addition of mineral fillers, the proposed research would aim to identify optimal combinations that enhance strength, durability, and functionality. 
The model also identifies an opportunity in the analysis of the structural properties of bone, including collagen fibril alignment and intermolecular interactions, to improve mycelium materials. 
A specific suggestion made is to combine mycelium with collagen in various configurations to fabricate scaffolds to asess their mechanical strength, porosity, and ability to facilitate cell growth and differentiation. The model, as X-LoRA did, suggests to explore the self-assembly of tropocollagen molecules within the mycelium matrix.
It is also suggested that one may manipulate the growth conditions of mycelium to alter its surface morphology and integrate hydrophobic compounds or structures.
The model identifies applications from construction to consumer goods, thereby reducing reliance on non-renewable materials.

\begin{figure}[htbp]
\begin{LLMbox}{\VG} %BioMixtral XLora 
\begin{lstlisting}
Given the graph's emphasis on nodes with high bridging centrality, we can infer that these nodes play pivotal roles in connecting diverse concepts and facilitating cross-disciplinary innovation. In the context of developing sustainable mycelium materials, we can leverage these strategic connectors to explore novel research avenues. Here's a synthesis of new research ideas based on the provided data:

1. Enhancing Mechanical Properties of Mycelium-Based Biocomposites through Mineral Fillers

Research Idea: Investigate the impact of various mineral fillers on the mechanical properties of mycelium-based biocomposites. Given that biocomposites can have improved mechanical properties due to the addition of mineral fillers, this study aims to identify optimal combinations that enhance strength, durability, and functionality.

Methodology: Experiment with different types and concentrations of mineral fillers in mycelium matrices. Assess the resulting mechanical properties, porosity, and surface morphology to determine the most effective enhancements.

Potential Impact: Develop stronger, more durable mycelium-based materials suitable for a wide range of applications, from construction to consumer goods, thereby reducing reliance on non-renewable materials.

2. Bioinspired Design of Mycelium Materials Mimicking Bone Structure

Research Idea: Explore the structural similarities between mycelium networks and bone tissue, focusing on porosity, collagen integration, and mechanical strength. This research seeks to mimic the hierarchical structure of bone to create mycelium materials with enhanced strength and resilience.

Methodology: Analyze the structural properties of bone, including collagen fibril alignment and intermolecular interactions. Apply these insights to design mycelium materials with similar hierarchical structures, assessing their mechanical properties and behavior under stress.

Potential Impact: By bioinspiring design principles from bone, this approach could lead to the development of mycelium materials with unprecedented strength and flexibility, suitable for biomedical applications and beyond.

3. Hydrophobicity and Surface Morphology in Mycelium Materials for Water-Resistant Applications

Research Idea: Given that mycelium imbues hydrophobicity and surface morphology impacts this property, this study aims to engineer mycelium materials with tailored surface characteristics for enhanced water resistance.

Methodology: Manipulate the growth conditions of mycelium to alter its surface morphology and integrate hydrophobic compounds or structures. Evaluate the materials' water resistance, mechanical properties, and potential for degradation in various environments.

Potential Impact: Create water-resistant mycelium materials with applications in outdoor construction, packaging, and water-sensitive environments, contributing to the sustainability and versatility of mycelium-based products.

4. Collagen-Mycelium Scaffolds for Biomedical Systems

Research Idea: Leverage the relationship between collagen and mycelium to develop innovative scaffolds for tissue engineering and regenerative medicine. This research focuses on creating biocompatible, mechanically robust scaffolds that support cell growth and tissue regeneration.

Methodology: Combine mycelium with collagen in various configurations to fabricate scaffolds. Assess their mechanical strength, porosity, and ability to facilitate cell growth and differentiation. Explore the self-assembly of tropocollagen molecules within the mycelium matrix.

Potential Impact: Offer new solutions for tissue engineering, with the potential to create scaffolds that closely mimic the mechanical and biological properties of natural tissues, enhancing the success of regenerative treatments.

Conclusion

By focusing on the strategic connectors within the network, these research ideas aim to harness the unique properties of mycelium and its interactions with other materials to push the boundaries of sustainable material development. Each idea represents a step towards integrating disparate areas of knowledge, from materials science to biology, to create innovative solutions for pressing environmental and technological challenges.
\end{lstlisting}
\end{LLMbox}
\textboxcaption{Response of GPT-4 when tasked to creatively synthesize new research ideas to build sustainable mycelium materials, for a graph extracted based on the $\rm top_N=20$ nodes with the highest bridging centrality, as they form strategic connectors as bridges between different clusters or communities.}
\label{TexBox:TB42}
\end{figure}

Compared to the previous responses, the three options provided here offer more details and deeper reflections about the topic. 

As an alternative, rather than providing the graph in its raw data of nodes and edges, mulitmodal LLMs that can consider images as input can be used to directly interpret the graph depicted in Figure~\ref{fig:Fig_12}d. To test this we prompt GPT-4V via ChatGPT:
\begin{figure}[H]
\begin{LLMbox}{} 
\begin{lstlisting}
[...] Carefully analyze this graph. Be creative and synthesize new research ideas to build sustainable mycelium materials.
\end{lstlisting}
\end{LLMbox}
\end{figure}
The response from GPT-4V is shown in Text Box~\ref{TexBox:TB50}. The results, while not as detailed as the earlier responses when we use the graph for prompting, still provide quite interesting insights. Since the provision of visual representations of data (e.g. the graph, but also figures from experiment, modeling, and so on) can in some circumstances be easier, this could serve as a viable means to interact with models and share data in a variety of modalities. GPT-4V shows a remarkable ability to correctly identify text in the image and understand graph relationships. The model identifies hierarchical organization of mycelium fibers and its implications on mechanical strength as a critical area of analysis. The model also suggests the development of mycelium materials that can change their properties in response to environmental stimuli such as temperature, pH, or moisture could lead to innovative applications. The full conversation is included as Supplementary Material, via Conversation S2.

Such a visual analysis of results (whether it is a graph, as done here, or other data representations) is particularly relevant in multi-agent AI systems where codes/AI agents may provide analysis in plotted form, such as a graph, where other agents conduct an analysis of them.

\begin{figure}[htbp]
\begin{LLMbox}{\VGV} %BioMixtral XLora 
\begin{lstlisting}
Analyzing the given graph, which seems to illustrate the interconnected nature of various material science concepts and highlights the importance of nodes like "mechanical strength," "mechanical properties," "mycelium," "sustainable materials," "biocomposites," and "porosity," we can synthesize new research ideas for building sustainable mycelium materials by considering the links and bridges suggested by the graph.

Here are some potential research ideas:

Enhanced Mycelium-based Composites: Given the connections between mycelium and biocomposites, there is potential to study how mycelium can be combined with other natural fibers or waste materials to enhance mechanical properties. For instance, research can focus on developing a composite material that balances strength and porosity, making it suitable for applications such as environmentally friendly packaging or insulation.

Mycelium and Collagen Scaffolds: The link between collagen and mycelium suggests an opportunity to research the integration of mycelium into regenerative medicine. We can explore the use of mycelium as a scaffold material enriched with collagen to support cell growth, which could have applications in tissue engineering.

Hierarchical Structures and Mycelium: Investigate the hierarchical organization of mycelium fibers and its implications on mechanical strength. This can include looking at the micro and nano-scale structural arrangement within mycelium and drawing parallels with hierarchical structures found in materials like bone.

Mycelium Mechanical Properties Optimization: Considering the central nodes of "mechanical properties" and "mechanical strength," there is a clear indication that research should also focus on optimizing these properties in mycelium materials through genetic engineering, growth condition manipulation, or post-harvest processing techniques.

Sustainable Mycelium Material Lifecycle Assessment: Analyze the entire lifecycle of mycelium materials, from cultivation to end-of-life, to identify key processes that could be improved for sustainability. This would involve a cross-disciplinary approach, combining knowledge from material sciences, biology, and environmental studies.

Mycelium as a Bridge in Material Science Education: Given the concept of bridging centrality, mycelium could be used as a case study in educational settings to illustrate the integration of biological systems with material sciences, potentially inspiring new ways of thinking about sustainable material design.

Smart Mycelium Materials: Inspired by the node "smart materials" and linked to "mechanical behavior," investigating the possibility of creating mycelium materials that can change their properties in response to environmental stimuli such as temperature, pH, or moisture could lead to innovative applications.

Improving Porosity and Functionality: Porosity is a key property that can be tailored in mycelium materials. Research could explore the optimal porosity for different applications, for example, maximizing porosity for sound-absorbing materials or minimizing it for load-bearing applications.

By combining these various aspects, we can push the boundaries of current material science and open up new possibilities for sustainable, biologically-derived materials with tailored mechanical properties and functionalities.
\end{lstlisting}
\end{LLMbox}
\textboxcaption{Response of GPT-4V when asked to carefully analyze an image of the graph shown in Figure~\ref{fig:Fig_12}d. We task the model to be creative and synthesize new research ideas to build sustainable mycelium materials.}
\label{TexBox:TB50}
\end{figure}

\subsection{Joint analysis of artistic images with graph reasoning and image synthesis for hierarchical materials design}

The previous section demonstrated a sophisticated ability of GPT-4V to understand complex images, there focused on analyzing visualizations of graphs. We now explore whether, and how, we can use a joint analysis of visual cues and graph reasoning to generate responses. We explore further how the responses can be used to formulate prompts for generative vision models that can predict, describe and contextualize actual microstructure images of the newly designed materials. The goal of this analysis is to examine possibilities of this method, future research may explore synthesis of this material that shows the possibility of using generative AI not only to integrate data from knowledge graphs but to also integrate additional visual cues. With the emergence of other multimodal AI systems, such as those that integrate audio/speech, video, and others, this can become a powerful approach towards integrated reasoning across scientific domains and modality of information representation. The scientific hypothesis is that the proposed modeling strategy can successfully integrated diverse modalities into a design process. 

As in the previous section here we focus on a subgraph derived from the subgraph shown in Figure~\ref{fig:Fig_12}d as context, and we select the $\rm top_N$ 20 nodes with the highest bridging centrality.

\begin{figure}
    \centering
    \includegraphics[width=.9\linewidth]{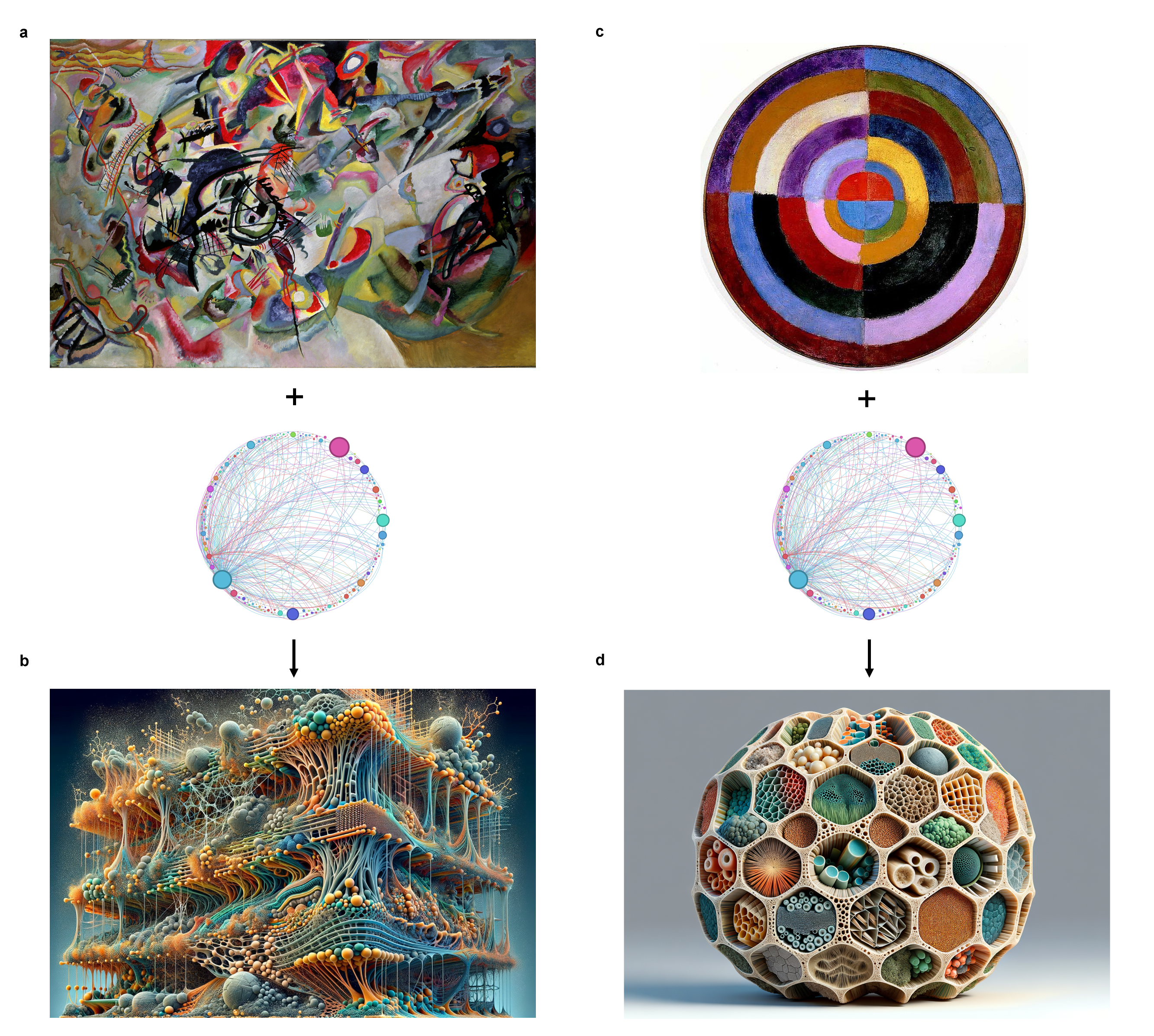}
    \caption{An approach to materials design, drawing inspiration from the abstract and emotional qualities of expressionist art to guide the development of sustainable mycelium-based materials. The original painting is shown in panel a (along with a thumbnail of the graph we reason over jointly with the image), and the generated material microstructure, representing a novel mycelium composite, in  panel b. The original art is the painting `Composition VII' by Wassily Kandinsky~\cite{Wassily1913} created in 1913, one his most significant works. Panel b shows the microstructure designed by DALL$\cdot$E 3, based on the description of the new material derived from graph reasoning and image analysis. Key features of this composite material includes tailored mechanical properties and hydrophobicity, leveraging the structural and molecular intricacies of mycelium and collagen to realize advanced biocomposites with potential for environmental responsiveness and self-repair and a hierarchical architecture. Table~\ref{tab:table_mycelium_design_using_AI} complements the image shown in panel b by explaining key design features, their representation in the image, the likely material function, and manufacturing or synthesis rout. Note, the generated image was cropped to match the aspect ratio of the original painting. Panels c and d show a second example, this time based on Robert Delaunay's painting `Le Premier Disque' created in 1912–1913. The result of this experiment shows a distinct emerging structure, prominently featuring circular motifs extracted from the original painting.  We also see that the content of the art influences the design outcome, as can be confirmed by comparing the left and right examples, that is, panels b versus panel d. The paintings shown in panels a and c are in the public domain in the United States because they were published (or registered with the U.S. Copyright Office) before January 1, 1929.}
    \label{fig:Fig_13}
\end{figure}

The progression of this approach proceeds in multiple stages. First, we jointly provide the image of interest with the graph and ask GPT-4V to reason over this data, to answer a particular question. The prompt consists of the image, and the graph, and the key task given (asking the model to combine the image data and graph data to answer the question):
\begin{figure}[H]
\begin{LLMbox}{} 
\begin{lstlisting}
[..] First, analyze the attached painting. Focus on mechanisms you can identify in the image, and describe them in great detail. It is important that you carefully analyze how this particular painting can be interpreted, with details.  

Second, use this carefully crafted information together with the graph triples to synthesize one high-impact research ideas to build sustainable mycelium materials. Use all context to develop new directions, and provide a detailed answer. Speculate about the specific behavior of the new material you would expect.

You think hard how to creatively relate the content in the image with the knowlegde graph provided. 

Pay attention to every detail in the context provided. You must include critical chemical formulas, molecular mechanisms, and physical processes. Your response must be at a very high level of sophistication. 
[..]
\end{lstlisting}
\end{LLMbox}
\end{figure}
As image source, we first consider the 1913 abstract painting `Composition VII' by Wassily Kandinsky~\cite{Wassily1913}, considered one of his most significant works (see, Figure~\ref{fig:Fig_13}). This painting is known for its emotional intensity, vibrant color palette, and dynamic movement (where colors and shapes could express spiritual experiences and emotions). The artwork combines complex abstract elements that transcend literal interpretation, and is hence well-suited for the type of abstraction we want to explore interfacing with graph reasoning.

The response is depicted in Text Box~\ref{TexBox:TB60}. We note that we are not providing GPT-4V with any details about the painting's name or creator, just the image along with the graph and the task context. The analysis can be broken down into several key aspects. These include features of the painting, which are identified as having vibrant color juxtapositions, dynamic shapes, and perceived spontaneous linework, all in an abstract non-objective reality. The model explains that it then uses these to design a material that harmonizes the chaos and order depicted in the painting with the intricate web of molecular and macroscopic interactions that afford a material its unique properties.

The model argues that we can discern parallels in how abstraction in art manipulates perception and emotion, while abstraction in materials science manipulates properties and functionality.  These principles are then used to construct an integrated design. The model takes inspiration from the fluid and interwoven patterns in the painting as a metaphor for creating composite materials with intricate fiber alignments that mirror the dynamic and intertwined nature of the artwork. The model also identifies the critical part played by porosity and the molecular interactions of collagen and hydrogen bonds in mechanical strength. 

It is further argued that collage could leverage the self-assembly behaviors to create complex hierarchical structures that confer additional mechanical properties. The model predicts that these structures would be analogous to the intricate and diverse lines in the composition of the artwork, serving as a physical manifestation of such abstract interplays. It is hypothesized that just like the varying brushstrokes that affect the perception of depth and movement, the micro-and nano-scale organization of the fibers could affect the mechanical properties of the biomaterial in question. The material can be further enhanced by the addition of mineral fillers to mimic features seen in dental enamel or bone. The model also predicts that growth could be guided in response to external stimuli, allowing for smart, responsive materials that adapt to varying loads or repair themselves autonomously. These sets of proposed design principles and associated properties provide actionable, testable hypotheses about anticipated material structures and their behaviors.

\begin{figure}[htbp]
\begin{LLMbox}{\VGV} %BioMixtral XLora 
\begin{lstlisting}
Analyzing the painting, we are presented with an abstract composition bustling with vibrant color juxtapositions, dynamic shapes, and seemingly spontaneous linework. This is an example of expressionism, where the emotional effect is paramount, and the depiction of objective reality is not the purpose. Instead, we immerse in the artist's psyche, perceiving the world subjectively through color, form, and texture. The mechanisms here are not mechanical in the literal sense but metaphorical, engaging with the viewer's perception and emotion. This composition blurs the lines between the figurate and the abstract, oscillating the viewer's interpretation between potential forms (perhaps evoking natural shapes, figures, or emotions) and the purely aesthetic appreciation of color and form.

Connecting this abstract approach from art to materials science using the knowledge graph, we can discern parallels in how abstraction in art manipulates perception and emotion, while abstraction in materials science manipulates properties and functionality. The graph indicates a strong interplay between mechanical properties, biological materials, and growth mechanisms, suggesting a multi-level synergy in the development of sustainable materials.

Now, synthesizing a complex research idea that fuses these themes:

To build sustainable mycelium-based materials, we could use inspiration from the fluid and interwoven patterns in the painting as a metaphor for creating composite materials with intricate fiber alignments that mirror the dynamic and intertwined nature of the artwork. Given the role of mycelium in enhancing mechanical properties and providing hydrophobicity, alongside the critical part played by porosity and the molecular interactions of collagen and hydrogen bonds in mechanical strength, a holistic approach to material development is necessary.

For instance, manipulating the porosity of the mycelium substrate may allow for tailoring its mechanical strength. By tuning the fiber alignment and enhancing intermolecular interactions (such as hydrogen bonds found in collagen fibrils), we can achieve desired mechanical properties. Recognizing that mycelium-based materials imbue hydrophobicity, we could speculate that surface morphology plays a significant role in water-resistance and possibly the mechanical adhesion properties of the composite material. Therefore, controlling surface morphology could impact hydrophobicity - ultimately leading to novel biocomposites with enhanced moisture resistance and durability.

By incorporating collagen-derived bio-scaffolds into mycelium materials, or using the basic structural unit of collagen-the tropocollagen molecule-we could leverage the self-assembly behaviors to create complex hierarchical structures that confer additional mechanical properties. These structures would be analogous to the intricate and diverse lines in the composition of the painting, serving as a physical manifestation of such abstract interplays - and just like the varying brushstrokes that affect the perception of depth and movement, the micro-and nano-scale organization of the fibers could affect the mechanical properties of the biomaterial in question.

In the context of sustainability, biocomposites can be further enhanced by the addition of mineral fillers. This could lead to improved mechanical properties reminiscent of the rigidity and structural integrity seen in dental enamel or bone, both noted in the graph to possess mechanical properties.

Speculating further, we could explore the behavior of the new material under different environmental conditions. Considering the biological aspect of mycelium, its growth could potentially be guided or modified in response to external stimuli, allowing for smart, responsive materials that adapt to varying loads or repair themselves autonomously.

The physical processes involved in developing such a material include the controlled biological growth of mycelium, the chemical cross-linking of collagen molecules to form fibrils, and the incorporation and alignment of mineral fillers to enhance the structural matrix. Critical chemical formulas would involve the sequences and cross-links of amino acids in the collagen strands, while the molecular mechanisms behind the self-assembly and bonding of biological materials would be essential to master to achieve the desired properties.

In conclusion, by bridging the conceptual divide between such an emotionally charged painting and the science of material creation, we can devise a research endeavor that seeks to develop sustainable, bio-inspired materials that harmonize the chaos and order depicted in the image with the intricate web of molecular and macroscopic interactions that afford a material its unique properties.
\end{lstlisting}
\end{LLMbox}
\textboxcaption{Response of GPT-4V when asked to  analyze the 1913 abstract painting 'Composition VII' by Wassily Kandinsky~\cite{Wassily1913}.}
\label{TexBox:TB60}
\end{figure}

It is notable that this analysis yields a much more multidimensional design that incorporates a more integrated featuring of the key components (e.g. collagen, and other small molecules), design cues (e.g., chaos and order, shapes and porosity gradients) and mechanisms (e.g. self-assembly driven by amino acid patterns, chemical functionalization to drive these, etc.) compared to the earlier designs. 
Now that we have a verbal description of the design we ask GPT-4 to develop, based on the material design it had developed earlier, a prompt for DALL$\cdot$E 3, a text-to-image generative AI model. With \hl{\texttt{<RESPONSE>}} as captured in the prompt below, the resulting prompt is:
\begin{figure}[H]
\begin{LLMbox}{} 
\begin{lstlisting}
Consider this description of a novel material: (*@\hl{<RESPONSE}@*)
        
Develop a well-constructed, detailed and clear prompt for (*@DALL$\cdot$E 3@*) that allows me to visualize the new material design. 
        
The prompt should be written such that the resulting image presents a clear reflection of the material's real microstructure and key features. Make sure that the resulting image does NOT include any text.
\end{lstlisting}
\end{LLMbox}
\end{figure}

We ask the model to not generate text since DALL$\cdot$E 3 has known difficulties generating letters/words correctly. The resulting DALL$\cdot$E 3 prompt is:
\begin{figure}[H]
\begin{LLMbox}{\VG} 
\begin{lstlisting}
Create an image that visualizes a novel, sustainable mycelium-based composite material inspired by expressionist art. The material should exhibit intricate, interwoven fiber alignments that mirror the dynamic, intertwined nature of abstract expressionist paintings, with vibrant color juxtapositions, dynamic shapes, and spontaneous linework. Emphasize the material's microstructure, showcasing a complex network of mycelium fibers intertwined with collagen-derived bio-scaffolds, creating a multi-layered, hierarchical structure. The fibers should vary in thickness and alignment, representing the diverse brushstrokes of the painting, and should also display areas of porosity and denser clusters, indicating tailored mechanical strength and hydrophobic properties. Include subtle hints of mineral fillers integrated into the matrix, enhancing the structural integrity and moisture resistance, reminiscent of the rigidity found in natural bone and dental enamel. The overall composition should convey a sense of depth and movement, reflecting the abstract interplay between chaos and order, and the fusion of art and science in material design. Ensure the image is rich in detail, allowing for a clear visualization of the material's unique micro and nano-scale organization, without including any text or explicit depiction of the painting itself.
\end{lstlisting}
\end{LLMbox}
\end{figure}

This prompt is then directly fed into DALL$\cdot$E 3 to realize a visual representation of the design. 
We depict an example of the generated microstructure composite design in Figure~\ref{fig:Fig_13}b. The resulting images are generally detailed and include rich design insights. It is noted that some of the produced images  contain text in spite of the specific instruction to not do so. This is actually interesting since it may show (if the predicted text can be deciphered, or properly rendered in future improved versions of text-to-image generators) exactly what material, chemistry or feature the model places in different regions. 

We also test whether the artwork used to reason over together with the graph can strongly affect the outcome. 
To explore this in more depth, Figure~\ref{fig:Fig_13}c and d show a second example, this time based on Robert Delaunay's painting `Le Premier Disque'. The result of this experiment shows a distinct  structure, prominently featuring circular motifs extracted from the original painting. Further research is needed, but these initial explorations show powerful options as we imaging sourcing creativity. It is noted that these translations are conducted not directly from the image, but rather by prompting GPT-4V to reason over the image and combine the image reasoning with the material design principles captured in the graph representation. Its surface is made up of many hexagonal and round cells of different sizes, resembling a molecular structure or organic growth. Its appearance evokes associations with natural structures like honeycombs, or coral reefs.

The analysis of the produced microstructure image (Figure~\ref{fig:Fig_13}b) can be taken even further by using GPT-4 via ChatGPT to interpret the image generated, in order to gain a better understanding of what the various components in the image are, what role they play in the integrated system, and how they can be manufactured. The full conversation is included in the Supplementary Materials section, see Supplementary Conversation S3. The key insights are summarized in Table~\ref{tab:table_mycelium_design_using_AI} covering design features, their representation in the image shown in Figure~\ref{fig:Fig_13}b, the likely material function, and manufacturing or synthesis route. These multimodal interactions offer powerful pathways to inform the design process and give engineers critical information about implementations. If experiments are conducted, their results can naturally be fed back to the model via conversational interactions to adapt, improve or alter the design (e.g. to achieve greater manufacturability likelihood or to reduce cost). If combined with autonomous manufacturing methods such approaches can lead to sophisticated experimental-modeling reasoning loops that can be automated via agentic modeling, for instance.

\begin{table}[h!]
\footnotesize
\centering
\caption{Materials design analysis for a sustainable mycelium composite depicted in Figure~\ref{fig:Fig_13}b. The analysis provides a detailed description of a sustainable mycelium composite material, where design features such as mycelium fibers, collagen-derived bio-scaffolds, and mineral fillers are intricately represented and correspond to specific functions like structural integrity and hydrophobicity. Advanced manufacturing techniques, including recombinant DNA technology for collagen synthesis and sequential bioprinting for hierarchical structuring, are delineated, offering insight into the material's complex, multi-functional design.}
\begin{tabular}{|p{2.5cm}|p{3cm}|p{4cm}|p{5cm}|}
\hline 
\textbf{Design feature} & \textbf{Representation in the image} & \textbf{Likely material function} & \textbf{Manufacturing or Synthesis Process with Technical Specifics} \\
\hline
Mycelium Fibers & Long, intertwined strands & Framework providing tensile strength and flexibility. & Cultivation of mycelium from strains like \textit{Ganoderma lucidum} on substrates rich in lignin and cellulose, optimizing growth conditions for the desired fiber morphology. \\
\hline
Collagen-Derived Bio-Scaffolds & Net-like, fibrous structures & Bio-compatibility and structural integrity. & Synthesis using recombinant DNA technology to express specific collagen sequences like Gly-Pro-Pro in \textit{E. coli}, followed by purification and self-assembly into scaffolds. \\
\hline
Mineral Fillers & Spherical and ovoid inclusions & Enhance compressive strength and thermal stability. & Incorporation of nano-scale hydroxyapatite or biogenic silica during the maturation phase of mycelium growth, potentially via a sol-gel process. \\
\hline
Porosity & Visible gaps within the matrix & Lightweight nature with insulative properties. & Engineering porosity using freeze-drying techniques to sublimate solvent, creating a controlled pore size distribution within the range of 100-500 micrometers. \\
\hline
Denser Clusters & Compact regions & Localized load-bearing capacity. & Densification through mechanical compression post-harvest or by introducing cross-linking agents like glutaraldehyde to selected areas. \\
\hline
Hydrophobic Properties & Surface texture variations & Moisture resistance and controlled fluid interaction. & Surface modification using silane coupling agents or fluorinated compounds to create hydrophobic domains at the nano-scale. \\
\hline
Color Representations & Vibrant color diversity & Aesthetic qualities and indicative of functional zones. & Use of pH indicators or thermochromic pigments that respond to environmental stimuli, serving as sensors for material state or damage. \\
\hline
Depth and Movement & Multi-layered, dynamic patterns & Hierarchical structuring for multi-functional performance. & Sequential bioprinting of mycelium-composite layers, possibly incorporating gradient transitions in material composition for stress dissipation. \\
\hline
\hline
\end{tabular}
\label{tab:table_mycelium_design_using_AI}
\end{table}

How can we better understand how to manufacture this material? To gain insights we tasked GPT-4V to \texttt{describe a detailed step-by-step process by which I can manufacture the entire design}. The resulting table is provided in Table~\ref{tab:mycelium_manufacturing_process} (note, like many of the earlier examples, GPT-4V directly produces LaTeX format in structured processing). This additional data provides fascinating avenues that can be pursued in future experimental work to act on the sets of hypotheses and predicted behaviors identified by the model.

\begin{table}[h!]
\footnotesize
\centering
\caption{Step-by-step manufacturing process for mycelium composite material, as proposed by GPT-4V.}
\begin{tabular}{|p{2cm}|p{3cm}|p{9cm}|}
\hline 
\textbf{Step} & \textbf{Stage} & \textbf{Process Details} \\
\hline
1 & Mycelium Cultivation & Select a robust mycelium strain, prepare a lignin-cellulose rich substrate, inoculate under sterile conditions, optimize growth parameters, and harvest the fibrous mat. \\
\hline
2 & Scaffold Synthesis & Synthesize collagen genes, express in \textit{E. coli}, purify protein, and form into scaffolds via electrospinning or molding. \\
\hline
3 & Mineral Filler Integration & Select nano-scale hydroxyapatite or silica, possibly process via sol-gel, and incorporate evenly into the mycelium matrix. \\
\hline
4 & Composite Formation & Layer mycelium and collagen scaffolds, apply pressure for densification, and dry to induce porosity and remove moisture. \\
\hline
5 & Surface Modification & Apply hydrophobic agents for moisture resistance and create surface textures for desired properties. \\
\hline
6 & Post-processing & Cut and shape material, conduct property tests, and sterilize if necessary for biomedical applications. \\
\hline
7 & Functionalization (Optional) & Integrate environmental response elements like pH or thermochromic pigments for functional zones. \\
\hline
8 & Final Inspection & Perform quality checks and package the final product to maintain condition until use. \\
\hline
\hline
\end{tabular}
\label{tab:mycelium_manufacturing_process}
\end{table}

We conclude with a summary of mycelium composite material design results and putting this into a much broader context. The response below was obtained by sharing a draft of this paper with Claude-3 Opus, tasking the model to identify \texttt{5 key high impact scientific insights. These should particularly focus on the material design aspects}. As in the earlier example around the Beethoven-material isomorphism, we asked the model several follow-up questions, such as to \texttt{go into more details by adding a few sub-bullets to each of the five}, and others. The resulting text listed here is the integrated version, slightly edited, result of these interactions. The result is shown in Text Box~\ref{TexBox:TB100}. The resulting text provides a detailed summary of the principles created by the generative model. It connects a lot of the ideas into an integrated summary of what the key insights and principles are. It is another demonstration of how human-AI collaborations can extend across multiple steps and yield a variety of actionable results. 

\begin{figure}[htbp]
\begin{LLMbox}{Claude-3 Opus} 
{\scriptsize
\texttt{
\begin{enumerate}[leftmargin=*]
\item Novel mycelium-collagen composites were designed, leveraging the intricate interplay between the molecular structures and interactions of mycelium fibers and collagen-derived bio-scaffolds. The composites feature a multi-layered, hierarchical structure with tailored mechanical properties, porosity, and hydrophobicity.
\begin{itemize}
\item The mycelium fibers, composed of chitin and other polysaccharides, provide a porous and adaptable framework for the composite.
\item Collagen-derived bio-scaffolds, synthesized using recombinant DNA technology, introduce biocompatibility and structural integrity to the composite.
\item The hierarchical structure of the composite mimics natural materials, enabling enhanced mechanical performance and multifunctionality.
\end{itemize}
\item The mycelium-collagen composites incorporate mineral fillers such as hydroxyapatite or biogenic silica to enhance compressive strength and thermal stability. The mineral fillers are integrated into the mycelium matrix during the maturation phase of growth, possibly via a sol-gel process.
\begin{itemize}
    \item Hydroxyapatite, a main component of bone, can be incorporated as a nano-scale filler to improve the composite's mechanical properties and biocompatibility.
    \item Biogenic silica, derived from natural sources (e.g., diatoms), can be used as a sustainable and biocompatible filler to enhance thermal stability and mechanical strength.
    \item The sol-gel process allows for the controlled integration of mineral fillers into the mycelium matrix, ensuring a homogeneous distribution throughout the composite.
\end{itemize}
\item Surface functionalization using silane coupling agents or fluorinated compounds is employed to create hydrophobic domains at the nano-scale in the mycelium composites. This surface modification, combined with engineered porosity through freeze-drying techniques, imparts moisture resistance and controlled fluid interaction to the materials.
\begin{itemize}
    \item Silane coupling agents, such as triethoxysilanes, can be used to modify the surface chemistry of the mycelium fibers, introducing hydrophobic functional groups.
    \item Fluorinated compounds, like fluoroalkylsilanes, can be grafted onto the composite's surface to create a low-energy, water-repellent layer.
    \item Freeze-drying techniques allow for the controlled sublimation of solvents, creating a porous structure with tailored pore size distribution and interconnectivity.
\end{itemize}
\item The manufacturing process for the mycelium-collagen composites involves sequential bioprinting of mycelium-composite layers, allowing for the fabrication of complex geometries and potential gradient transitions in material composition. Post-processing steps like densification through mechanical compression and cross-linking further tune the mechanical properties.
\begin{itemize}
    \item Bioprinting enables the precise deposition of mycelium-composite layers, facilitating the creation of intricate architectures and spatially varied material properties.
    \item Gradient transitions in material composition can be achieved by altering the ratio of mycelium to collagen or by introducing additional components during the bioprinting process.
    \item Mechanical compression can be applied post-printing to densify the composite, enhancing its mechanical strength and reducing porosity in targeted regions.
    \item Cross-linking agents, such as glutaraldehyde, can be employed to establish covalent bonds between the mycelium fibers and collagen, further stabilizing the composite's structure.
\end{itemize}
\item The mycelium-collagen composites are anticipated to exhibit novel functionalities such as self-healing and adaptability. Leveraging mycelium's ability to grow and recolonize, the composites could potentially self-repair upon damage. Environmental responsiveness may also be integrated through the incorporation of pH indicators or thermochromic pigments, enabling the materials to serve as sensors for monitoring their state or integrity.
\begin{itemize}
    \item The self-healing capability of the composites relies on the regenerative potential of mycelium, which can bridge and recolonize damaged regions, restoring the material's structural integrity.
    \item Adaptability can be achieved by exploiting the responsive nature of mycelium to environmental stimuli, such as changes in moisture, temperature, or nutrient availability.
    \item pH indicators (e.g., phenolphthalein or bromothymol blue), could be embedded within the composite to visualize changes in pH, serving as a diagnostic tool for monitoring the material's state.
    \item Thermochromic pigments, which change color in response to temperature fluctuations, can be incorporated to create visually responsive composites that indicate thermal stresses or changes in environmental conditions.
\end{itemize}
\end{enumerate}
%%%%%%%%%%%%%%%%%%%%%%%%%%%%%%%%%%%%%%%%%%%%%%%%%%%%%%%%%%%%%%
}
}
\end{LLMbox}
\textboxcaption{Summary of mycelium composite material design results, obtained by sharing a draft of this paper with Claude-3 Opus. We task the model to focus specifically on the content of the paper draft shared. The resulting text provides a detailed summary of the principles created by the generative model. }
\label{TexBox:TB100}
\end{figure}

\section{Conclusion}
We performed an extensive analysis of roughly 1,000 scientific papers focusing on bioinspired materials and mechanics. By organizing the extracted information into a detailed ontological knowledge graph were able to map the complex web of connections that underpin this body knowledge, which is a highly interdisciplinary area with several intersecting disciplines (materials, chemistry, biology, mechanics, etc.). The resulting graph serves as structured representation of a very large number of entities and their interrelations within the field, facilitating a systematic exploration and synthesis of knowledge across different scientific areas~\cite{Spivak2011CategoryNetworks, Spivak2011ReoccurringAnalogies,Giesa2012CategoryDesign, Buehler2023GenerativeDesign}.
Our analysis revealed a robust connectivity all across, with certain topics emerging as central nodes within the network. This pattern suggested that the structure of the knowledge graph exhibits characteristics of a scale-free network, a type of network known for a few highly connected nodes amid numerous lesser-connected ones. Leveraging this structure, we reported a series of experiments designed to probe the depth and breadth of the interconnected knowledge landscape.
These experiments included tasks such as identifying the shortest paths between disparate scientific concepts, utilizing the graph's structure to answer complicated scientific questions, and dynamically expanding the graph with new insights through methods like agentic brainstorming with a the X-LoRA mixture-of-experts model and incorporating on-the-fly data collection from recent literature. A significant outcome of these efforts was the identification of novel relationships and cross-disciplinary linkages that had not been previously recognized. This enriches the reasoning ability and offers numerous avenues for future research and development and creative discovery. 

This paper presents a robust mechanism for creating knowledge graph representations from large datasets like scientific papers, identifying key properties such as their scale-free nature. Our automated and scalable algorithm constructs these graphs, which were rigorously analyzed. By exploiting the transitive properties of graphs, we extracted sub-graphs to generate new hypotheses and innovative design proposals, such as novel food-focused protein materials and the connection of disparate ideas. We introduced isomorphic analysis to facilitate knowledge transfer across diverse domains, a novel scientific approach. Additionally, we demonstrated the dynamic generation of new graphs for various applications, showcasing the effectiveness of both open-source and fine-tuned generative AI models. Our material design processes were achieved without human intervention, with AI-generated manufacturing steps illustrating the understanding of the generative AI models, and associated practical application in material synthesis. Overall, this work exemplifies a human-AI collaborative workflow with potential applications beyond materials science, integrating multidisciplinary analyses to leverage machine learning for exploring complex interactions across fields.

The study  described a number of experiments in order to demonstrate that the knowledge graph can be used to perform a variety of tasks, including:
\begin{itemize}
    \item Answer questions about the field of bio-inspired materials and mechanics
    \item Generate new research ideas
    \item Identify gaps in the knowledge base, such as existing scientific literature
    \item Suggest new directions for research that will provide more data (which in itself can then be used to augment the graphs, to find additional relationships that were previously not known)
    \item Combine vision-based models to analyze graphs, and to combine an analysis of painting with material design synthesis to create actual microstructure images of proposed materials
\end{itemize}

This research underscores the power of integrating graph theoretical methodologies with generative modeling techniques to foster the discovery of new scientific and technological insights. By applying advanced analysis techniques such as node embeddings, isomorphic mappings, and graph analysis focusing on metrics like betweenness centrality, we were able to navigate and interpret the complex network of knowledge. Moreover, we demonstrated that multimodal vision-language models could be effectively applied to reason directly over specific highly connected sub-graphs extracted from the overall knowledge graph. This integration of knowledge graph analysis with advanced computational models opens up new avenues for generating knowledge, enabling researchers to uncover hidden patterns and connections that can lead to groundbreaking discoveries in bioinspired materials, mechanics, and beyond. In other examples we used generative models to synthesize microstructure representations of the designs (Figure~\ref{fig:Fig_13}), along with a detailed labeling of the features shown in the image (Table~\ref{tab:table_mycelium_design_using_AI}) and a manufacturing process (Table~\ref{tab:mycelium_manufacturing_process}). 

We explored the use of both, fine-tuned special purpose open source models and powerful general-purpose models in the analysis, and offer responses from these models for direct insights into the performance of these models. We find that the fine-tuned open source models, albeit much smaller than GPT-4, can provide deep and exhaustive responsive, underscoring their usefulness in a design process. 

\subsection{Key technical insights about materials}

Figure~\ref{fig:Fig_8} depicted an isomorphic mapping between the bioinspired materials graph and Beethoven's 9th Symphony graph that elucidates profound homologies across domains. These feature universal patterns of structural organization, dynamical evolution, and hierarchical interdependence resonating from the sub-molecular choreographies of matter to the symphonic grandeurs of artistic expression, underscoring how the knowledge graph analysis unveils the transcendental harmonies that unite all manifested forms. For instance, the isomorphic mapping between adhesive forces and musical tonality inspires the hypothesis that there may be deep commonalities between the laws governing electromagnetic and aesthetic forces, facilitating transitions from disorder to complexity, while the analogy between cantilever beams and Beethoven's seminal work raises the possibility that generative processes in music and biology may share variation-selection dynamics - both hinting at an overarching meta-theory wherein all manifested structures and creative novelties emerge from an underlying algebra of primordial organizational principles. Variation-selection dynamics intersect with the concept of fluctuations, such as stochastic processes seen in molecular motions~\cite{Greiner2000ThermodynamicsMechanics,Allen1987ComputerLiquids}, by illustrating how random variations can lead to selective advantages in certain environments. In molecular biology, stochastic fluctuations in gene expression can introduce variation in phenotypic traits among individual building blocks or entities, which then undergo selection processes, driving evolutionary changes and adaptations in populations. In these examples, important weight is carried by the significance of individual building blocks and their stochastic behavior and relationship with other elements, suggesting that biological materials must be understood as heterarchical structures rather than hierarchical structures~\cite{Nepal2022HierarchicallyNanocomposites}. The elucidation of fundamental mechanisms of heterarchical structures is notable insight into the behavior of bio-inspired materials and puts forth an overall conclusion of this study. The analogies proposed between materials and philosophical work is another result that shows universal principles at work across fields, which would have remained unconnected.

This further underscores the importance of graph-forming attention mechanisms in modeling such systems, since models that capture such complexity must successfully encapsulate complex multiscale relationships~\cite{Hu2023DeepScience,Buehler2022MultiscaleNetworks,Ghafarollahi2024ProtAgents:Learning}. An important insight developed from this finding is that coarse-graining of unitary elements, based on the idea that materials can be modeled as scale-separated structures, is not generally applicable. It further suggests, in a functional sense, that emergence of new properties may be obtained in systems that are devoid of scale-separated structures, with important implications for our understanding of architected materials that are often designed with scale-separation of material features. 

Extending these ideas, the isomorphisms spanning physical, biological, and artistic domains may be phenomenological reflections of reality's deepest, unifying metaphysical principles - elemental dispositions and generative regimes rooted in symmetries and computational motifs whose abstract essences precede and govern all existent complexities as their ultimate `source code'. This would mean  that at the most fundamental level, reality may be governed by basic patterns or rules that are computational in nature, akin to how a computer program operates based on its source code, and that these patterns are characterized by symmetry and generative processes that produce the complex world we observe. Further research is needed to explore these initial ideas in more depth, and explore their relationship with models such as cellular automata~\cite{StephenContents}. 

Applying the graph-based methodology to a specific materials design, we reported a novel mycelium material. The design was obtained via joint reasoning over a traversal sampling based graph with nodes with the highest bridging centrality with the abstract expressionist painting `Composition VII' by Wassily Kandinsky, as shown in Figure~\ref{fig:Fig_13}. The painting's dynamic interplay of colors, shapes, and lines was translated into the composite's intricate, multi-layered structure, featuring an intertwined network of mycelium fibers and collagen-derived bio-scaffolds, along with strategically incorporated mineral fillers and surface modifications to create a hierarchical architecture with tailored mechanical properties, porosity, and functionality. The composite incorporates mineral fillers (hydroxyapatite or biogenic silica), surface functionalization (silane coupling agents or fluorinated compounds), and engineered porosity through freeze-drying techniques. The manufacturing process involves sequential bioprinting of mycelium-composite layers, allowing for complex geometries and gradient transitions, while post-processing steps such as mechanical compression and cross-linking further tune the mechanical properties. The mycelium-collagen composites are envisioned to exhibit novel functionalities such as self-healing, adaptability, and environmental responsiveness through the incorporation of pH indicators or thermochromic pigments. Identifying relationships between science and art concepts is intrinsically difficult, especially since they lack a common underlying language or formalism. Generative AI based strategies as proposed here provide a method that does not rely on human interpretation, and can be conducted autonomously. When coupled with human exploration, they can yield interesting insights and elucidate new connections between ideas or methods. If such relationships can be identified universally across different corpi of knowledge, they may form a basis for the discovery of unifying principles. Future work can build on the initial ideas explored here and conduct more research into formalisms towards discovery of governing principles that apply across a range of fields.

\subsection{Limitations and opportunities}

The method developed in this paper transcends traditional disciplinary boundaries, enabling a symbiotic relationship between fields as diverse as music theory, materials science, biology, and others. Through the lens of graph theory and generative AI, we have illuminated paths for novel interpretations and applications, fostering a paradigm where interdisciplinary collaboration is facilitated as an essential feature towards groundbreaking discoveries.
Multi- and interdisciplinary research has long been an area of great interest. Many earlier studies, however, are limited by generalized translations rather than rigorously developed methods to relate concepts and ideas. The method presented here provides a sound framework to achieve highly complex cross-domain translation, reasoning and discovery. 
This is because the integration of vast, disparate data sets into coherent, navigable knowledge graphs  significantly enhances our capability to identify patterns, connections, and previously unseen parallels across domains. This approach not only streamlines the discovery process in our materials science, but also holds promise for catalyzing similar advancements in other areas, potentially transforming how research is conducted across the sciences and humanities. 
While our findings  show the efficacy of knowledge graphs in facilitating complex analyses, they also highlight the challenges inherent in integrating multimodal data and extracting actionable insights. The nuances of interpreting data from vastly different sources, be it the structured formality of scientific research or the abstract expressions found in art and music, pose unique challenges that require careful assessments and methodological advances, like the use of isomorphic mappings.

The work reported in this paper not only contributes to the field of bioinspired materials and mechanics but also sets the stage for a future where interdisciplinary research, powered by AI and knowledge graphs, may become a tool of scientific and philosophical inquiry. As we look to future work, the integration of diverse knowledge streams through advanced computational methods (e.g. molecular, multiscale or continuum modeling) holds the promise of unveiling mysteries not just within one domain, but across the vast expanse of human knowledge, and specifically discovering new connections as done in this paper between musical structures or paintings and bio-inspired material designs.

Limitations of this work are that data processing was limited to a subset of research knowledge in a specific area, bio-inspired materials. Future work could expand this dataset to include more original papers especially building much larger corpi that feature more concentrated knowledge that would likely lead to a different graph structure. This calls for an integration of larger graphs that derive from even larger corpi of raw data. Future work may also include to use more sophisticated language models, like GPT-4, to extract triples to construct the global knowledge graphs, as opposed to the use of relatively small open-source models used in this paper. Current limitations in terms of computational throughput prevent us from using these methods, but this is likely to be addressed in the near future with the availability of higher token generation rates and/or the availability of higher-performance open-source models.  Extending the work towards these alternative graph generation methods is straightforward, however, and can easily be amended to existing graph data. 

While our initial analysis of clustering coefficients~\ref{fig:Fig_4}c has provided a foundational understanding, further investigation is warranted to fully leverage this metric's potential. For instance, future research could perform a comprehensive analysis of clustering coefficients across different regions of the graph to identify patterns and anomalies, and provide a detailed look into the scientific terms and relationships in each of these. This can possibly reveal additional important insights into the structural nuances of the knowledge graph.

The issue of manufacturability has not been explicitly explored in the scope of this study. It has been discussed in various places of this paper as important future work. From a broad perspective of a materials scientist, the ideas identified seem plausible in general and will likely serve as good starting points for designing and making new materials, but further experimental work will be critical to assess this in more detail.

It is noted that in particular the scalability and adaptability of knowledge graphs constructed and used using generative AI suggest a rich potential for further innovation. By extending these methodologies to encompass an even wider array of disciplines, there exists the possibility to not only enhance our understanding of bioinspired materials and mechanics but also to forge new tools for exploration and discovery that could revolutionize fields as varied as computational creativity, philosophical analysis, and theoretical science. When combined with agentic modeling or vision-focused science-focused models~\cite{Ghafarollahi2024ProtAgents:Learning,buehler2024cephalo}, many exciting opportunities occur. The use of multi-agent models in particular, combined with graph reasoning strategies as discussed in this paper, hold great potential as they can generate new data from first principles (e.g. Density Functional Theory, molecular modeling, etc.) based on hypotheses and ideas generated from graph analyses.
The application of AI and knowledge graphs raises certain ethical and philosophical questions, particularly regarding data integrity, bias, and the potential for AI to influence or direct the course of research and discovery in unforeseen ways. As we move further into this integrated approach, it becomes imperative to engage with these issues thoughtfully, ensuring that the pursuit of knowledge remains both responsible and reflective of underlying societal values. We believe that the use of graph reasoning can provide a powerful approach towards exploration of new research ideas, prompting models to make predictions about hypothetical materials or constituents, and to achieve scientific discovery through multi-modal generative AI.

\section{Materials and Methods}

Codes, data and examples are available at \url{https://github.com/lamm-mit/GraphReasoning}. 

\subsection{Overview of generative AI models used}

The analyses presented here are conducted with different generative models, including fine-tuned models targeted for materials and biological materials:
\begin{itemize}
    \item X-LoRA, a fine-tuned, dynamic dense mixture-of-experts large language model with strong biological materials, math, chemistry, logic and reasoning capabilities~\cite{Buehler2024X-LoRA:Design} that uses two forward passes (details see reference and discussion in main text)
    \item BioinspiredLLM-Mixtral, a fine-tuned mixture-of-experts (MoE) model based on the original BioinspiredLLM model~\cite{Luu2023BioinspiredLLM:Materials} but using a mixture-of-expert approach basde on the Mixtral model~\cite{Jiang2024MixtralExperts}
\end{itemize}
We also use general-purpose models, including:
\begin{itemize}
    \item  Mistral-7B-OpenOrca~\cite{lian2023mistralorca1,mukherjee2023orca,longpre2023flan} (used for text distillation into a heading, summary and bulleted list of detailed mechanisms and reasoning) 
    \item  Zephyr-7B-$\beta$~\cite{HuggingFaceH4/zephyr-7b-betaFace} built on top of the Mistral-7B model\cite{Jiang2023Mistral7B} (used for original graph generation due efficient compute and local hosting)
    \item GPT-4 (\texttt{gpt-4-0125-preview}), at the time of the writing of this paper, this is the latest GPT model by OpenAI~\cite{OpenAI2023GPT-4Report} (for some less complex tasks, specificall graph augmentation, we use GPT 3.5)
    \item  GPT-4V (\texttt{gpt-4-vision-preview}), a multimodal vision-text model by OpenAI~\cite{GPT-4VisionCard,Yang2023TheGPT-4Vision}, for some use cases accessed via \url{https://chat.openai.com/}
    \item Claude-3 Opus and Sonnet~\cite{IntroducingAnthropic}, accessed via \url{https://claude.ai/chats} 
\end{itemize}
The use of various models, especially open-source and closed-source and models of varying sizes, allow us to compare their capabilities. Detailed comparisons between the responses of the models is included in the paper, and can also be compared directly as raw data of predictions is shared.

\subsection{Data distillation and knowledge graph generation}

The process proceeds in several steps:
\begin{enumerate}
    \item Distillation of raw data from scientific papers (following a step-by-step process that includes 1) conversion of PDF into Markup language, 2) splitting into text chunks, 3) generation of raw context through distillation)
    \item Local knowledge graph generation by extracting triples (nodes and relationships) for each raw context, using an iterative process
    \item Global knowledge graph generation by concatenating all local graphs, followed by merging of similar nodes and optional simplifications (e.g. focus on giant component)  
\end{enumerate}
In the following we describe these steps in detail. The LLMs used are served using \texttt{Llamma.cpp} to achieve fast inference to provide efficient computational throughput. Since it is computationally expensive, knowledge graph generation is conducted using parallel computing in a multi-GPU framework. 

\subsubsection{Distillation of raw data from scientific papers}
We use the same set of original papers as reported in \cite{Luu2023BioinspiredLLM:Materials}, consisting of more than 1,000 research articles in the field of bioinspired materials, mechanics and materials. We first use the Nougat model~\cite{Blecher2023Nougat:Documents} to convert each of the raw PDF files into Markup language. 
We then divide these files into text chunks, here referred to as \hl{\texttt{<TEXT>}}, with a mean length of 772 words. 
For each chunk, we task the \texttt{Mistral-7B-OpenOrca} model\cite{lian2023mistralorca1,mukherjee2023orca,longpre2023flan} to develop a heading, a summary and a bulleted list of key insights, referred to as raw context for later use in graph generation. To create the raw context, We proceed in three steps. We first develop the summary using this system prompt:
\begin{figure}[H]
\begin{LLMbox}{} 
\begin{lstlisting}
You respond with a concise scientific summary, including reasoning. You never use names or references.
\end{lstlisting}
\end{LLMbox}
\setlength{\belowcaptionskip}{-5pt}
\end{figure}
Together with this user prompt (system/user prompts are provided in the \texttt{Mistral-7B-OpenOrca} prompt template):
\begin{figure}[H]
\setlength{\intextsep}{0pt} % Reduce space above and below the figure
\setlength{\belowcaptionskip}{-5pt} % Reduce space below the caption
\begin{LLMbox}{} 
\begin{lstlisting}
In a matter-of-fact voice, rewrite this "(*@\hl{<TEXT>}@*)". The writing must stand on its own and provide all background needed, and include details.  Do not include names, figures, plots or citations in your response, only facts.
\end{lstlisting}
\end{LLMbox}
\end{figure}
This is used to generate the summary, referred to as \hl{\texttt{<SUMMARY>}}.
Using the same system prompt, we then instruct the model to develop a  bulleted list;
\begin{figure}[H]
\begin{LLMbox}{} 
\begin{lstlisting}
Provide a bullet point list of the key facts and reasoning in "(*@\hl{<SUMMARY>}@*)". The writing must stand on its own and provide all background needed, and include details. Do not include figures, plots or citations in your response. Think step by step.
\end{lstlisting}
\end{LLMbox}
\end{figure}
We then ask the model to develop a brief title, using this system prompt:
\begin{figure}[H]
\begin{LLMbox}{} 
\begin{lstlisting}
You are a scientist who writes a scientific paper. You never use names or citations.
\end{lstlisting}
\end{LLMbox}
\end{figure}
The user prompt is:
\begin{figure}[H]
\begin{LLMbox}{} 
\begin{lstlisting}
Provide a one-sentence title of this text: "(*@\hl{<SUMMARY>}@*)". Make sure the title can be understood fully without any other context. Do not use the word "title", just provide the answer.
\end{lstlisting}
\end{LLMbox}
\end{figure}
These prompts were developed with the objective in mind to extract facts from the chunks of original text. This process results in a set of 8,663 raw context. 
An example of such an output of raw context is as follows. The heading is:
\begin{figure}[H]
\begin{LLMbox}{\MistralOpenOrca} 
\begin{lstlisting}
Investigation of bioactive mineralized biomaterials derived from D-Bamboo scaffolds for load-bearing bone regeneration through cell attachment assays, in vivo bone repairing experiments using rabbit models, and analysis of surface properties, revealing enhanced osteoconduction and osseointegration due to optimal hydrophilicity and transport capacity of bioactive calcium phosphate components.
\end{lstlisting}
\end{LLMbox}
\end{figure}
The summary is: 
\begin{figure}[H]
\begin{LLMbox}{\MistralOpenOrca} 
\begin{lstlisting}
The study investigated the bioactivity of mineralized biomaterials through cell attachment assays and in vivo bone repairing experiments using rabbit models. The D-Bamboo scaffolds without bioactive mineralized components were used as a control group. Scanning electron microscopy (SEM) revealed that both rabbit bone marrow stromal cells (rBMSCs) and human umbilical vein endothelial cells (HUVECs) adhered well to the CaP-Bamboo scaffolds with many pseudopodium, while HUVECs on D-Bamboo scaffolds showed undesirable spreading. In vivo experiments demonstrated that there was a gap between D-Bamboo and peripheral bone, whereas the CaP-Bamboo scaffolds and newly formed bone integrated well with each other. The enhanced bone integration capacity of CaP-Bamboo scaffolds was attributed to the introduction of bioactive calcium phosphate (Cap) components. The study also found that the Cap (0001)-Ca surface had optimal hydrophilicity and transport capacity due to its low diffusion energy barrier for water adsorption. The newly bone tissues grew along the porous channels into the interior of both D-Bamboo and CaP-Bamboo scaffolds, with bioactive mineralized components promoting bone ingrowth into porous channels of CaP-Bamboo scaffolds. However, there was no obvious degradation of either D-Bamboo or CaP-Bamboo scaffolds after 8 weeks of implantation due to the lack of enzymes to break down cellulose and hemicellulose. The study suggests that the CaP-Bamboo scaffolds have good cell compatibility, osteoconduction, and osseointegration performance, showing promising potential for load-bearing bone regeneration.
\end{lstlisting}
\end{LLMbox}
\end{figure}
Finally, the bullet point list of the key facts and reasoning is: 
\begin{figure}[H]
\begin{LLMbox}{\MistralOpenOrca} 
{\scriptsize
\texttt{\begin{enumerate}[leftmargin=*]
    \item Study investigated bioactivity of mineralized biomaterials through cell attachment assays and in vivo bone repairing experiments using rabbit models.
\begin{enumerate}
    \item D-Bamboo scaffolds without bioactive mineralized components used as control group.
\end{enumerate}
\item Scanning electron microscopy (SEM) revealed:
\begin{enumerate}
    \item  Rabbit bone marrow stromal cells (rBMSCs) and human umbilical vein endothelial cells (HUVECs) adhered well to CaP-Bamboo scaffolds with many pseudopodium.
   \item HUVECs on D-Bamboo scaffolds showed undesirable spreading.
\end{enumerate}
\begin{enumerate}\item In vivo experiments demonstrated:
   \item Gap between D-Bamboo and peripheral bone, whereas CaP-Bamboo scaffolds integrated well with newly formed bone.
\end{enumerate}
\item Enhanced bone integration capacity of CaP-Bamboo scaffolds attributed to introduction of bioactive calcium phosphate (Cap) components.
\item Cap (0001)-Ca surface had optimal hydrophilicity and transport capacity due to low diffusion energy barrier for water adsorption.
\item Newly bone tissues grew along porous channels into interior of both D-Bamboo and CaP-Bamboo scaffolds, with bioactive mineralized components promoting bone ingrowth into porous channels of CaP-Bamboo scaffolds.
\item No obvious degradation of either D-Bamboo or CaP-Bamboo scaffolds after 8 weeks of implantation due to lack of enzymes to break down cellulose and hemicellulose.
\item Study suggests that CaP-Bamboo scaffolds have good cell compatibility, osteoconduction, and osseointegration performance, showing promising potential for load-bearing bone regeneration.
\end{enumerate}}
}
\end{LLMbox}
\end{figure}
This distillation process ensures that the original content is organized in well-readable, well-summarized text that includes both facts and reasoning. It forms the basis for graph generation, as will be explained in the next section.

\subsubsection{Knowledge graph generation from text chunks}
\label{context_graph}
We use \texttt{Zephyr-7B}-$\beta$~\cite{HuggingFaceH4/zephyr-7b-betaFace} that was built on top of the Mistral-7B model\cite{Jiang2023Mistral7B}  to extract triples (two nodes that delineate concepts, and edge features that describe their relationship) from text chunks, here forth referred to as context, following the strategy discussed in~\cite{Rahulnyk/knowledge_graph:QnA_updated,Buehler2024X-LoRA:Design} with additional features based on the Llama Index graph generation algorithm~\cite{Liu_LlamaIndex_2022}. We present a summary of the method used, whereas details can be found in the code implementation of the algorithm. 

The initial process of assembling the knowledge graph follows the approach suggested in~\cite{Liu_LlamaIndex_2022}:
\begin{figure}[H]
\begin{LLMbox}{} 
\begin{lstlisting}
You are a network ontology graph maker who extracts terms and their relations from a given context, using category theory. You are provided with a context chunk (delimited by ```) Your task is to extract the ontology of terms mentioned in the given context. These terms should represent the key concepts as per the context, including well-defined and widely used names of materials, systems, methods. 

Format your output as a list of JSON. Each element of the list contains a pair of terms and the relation between them, like the following: [...]
\end{lstlisting}
\end{LLMbox}
\end{figure}
We define the way the triples of nodes and edge features should be built, as follows:
\begin{figure}[H]
\begin{LLMbox}{} 
{\scriptsize
\texttt{
\begin{itemize}[leftmargin=*]
\item "node\_1": "A concept from extracted ontology"
\item "node\_2": "A related concept from extracted ontology"
\item "edge": "Relationship between the two concepts, node\_1 and node\_2, succinctly described"
\end{itemize}
}
}
%\end{lstlisting}
\end{LLMbox}
\end{figure}
We further provide example of an ontological analysis. This begins with a short context:
\begin{figure}[H]
\begin{LLMbox}{} 
{\scriptsize
\texttt{
Context: ```Silk is a strong natural fiber used to catch prey in a web. Beta-sheets control its strength.```
}
}
%\end{lstlisting}
\end{LLMbox}
\end{figure}
We then give example triples, such as 
\begin{figure}[H]
\begin{LLMbox}{} 
{\scriptsize
\texttt{
\begin{itemize}[leftmargin=*]
            \item "node\_1": "spider silk"
            \item "node\_2": "fiber"
            \item "edge": "is"
\end{itemize}
\begin{itemize}[leftmargin=*]
            \item "node\_1": "beta-sheets"
            \item "node\_2": "strength"
            \item "edge": "control"
\end{itemize}      
\begin{itemize}[leftmargin=*]
            \item "node\_1": "silk"
            \item "node\_2": "prey"
            \item "edge": "catches"
\end{itemize}
}
}
%\end{lstlisting}
\end{LLMbox}
\end{figure}
We finally instruct the model to 
\begin{figure}[H]
\begin{LLMbox}{} 
{\scriptsize
\texttt{Analyze the text carefully and produce around 10 triplets, making sure they reflect consistent ontologies.
}
}
%\end{lstlisting}
\end{LLMbox}
\end{figure}
To ensure the output is structured, we ask the model to provide the response in JSON format, and we provide an example in the instruction as well. 

We next feed the original context \hl{\texttt{<CONTEXT>}}  and the initial set of triplets  \hl{\texttt{<RESPONSE>}} to the model and instruct it to revise it. The instructions are:
\begin{figure}[H]
\begin{LLMbox}{} 
\begin{lstlisting}
Read this context: ```(*@\hl{<CONTEXT>}@*)```.

Read this ontology: ```(*@\hl{<RESPONSE>}@*)```.

Improve the ontology by renaming nodes so that they have consistent labels that are widely used in the field of materials science.
\end{lstlisting}
\end{LLMbox}
\end{figure}

Finally, we ask the model to ensure that the output follows the correct JSON output format. 

Since not all graph generation steps are successful (e.g. because some text chunks contain no useful information, or because the model is not able to construct a properly formatted output), we conduct a second sweep over all text chunks for which graph generation had failed initially. This results in generation of graphs for some cases, albeit the success rate is relatively low (indicating a stability with respect to the ability of the language model to successfully extract triplets). 

The raw output in JSON format is, as an example:
\begin{figure}[H]
\begin{LLMbox}{\ZephyrBeta} 
\begin{lstlisting}
[
  {'node_1': 'Biofabrication', 'node_2': 'Tissue Engineering', 'edge': 'has gained attention due to'},
  {'node_1': 'Biofabrication', 'node_2': '3D Printing', 'edge': 'convergence'},
  {'node_1': 'Volumetric Printing', 'node_2': 'Centimeter-scale structures', 'edge': 'allows for'},
  {'node_1': 'Hydrogel-based Bioresins', 'node_2': 'Volumetric Printing', 'edge': 'uses'},
  {'node_1': 'Microscale Extrusion Writing (MEW)', 'node_2': 'Highly organized fiber architectures', 'edge': 'creates'},
  {'node_1': 'Low-stiffness Hydrogels', 'node_2': 'Microscale Extrusion Writing (MEW)', 'edge': 'confers exceptional mechanical properties to'},
  {'node_1': 'Polycaprolactone', 'node_2': 'Microscale Extrusion Writing (MEW)', 'edge': 'used for'},
  {'node_1': 'Gelatin Methacryloyl (GelMA)', 'node_2': 'Volumetric Printing', 'edge': 'used as'},
  {'node_1': 'Polycaprolactone', 'node_2': 'Volumetric Printing', 'edge': 'used for'},
  {'node_1': 'In vivo Degradation Time', 'node_2': 'Polycaprolactone', 'edge': 'features'},
  {'node_1': 'Occlusion Points', 'node_2': 'Opaque PCL Meshes', 'edge': 'cause'},
  {'node_1': 'Light Attenuation Effect', 'node_2': 'Opaque PCL Meshes', 'edge': 'caused by'},
  {'node_1': 'Partially Blocking Projection Paths', 'node_2': 'Occlusion Points', 'edge': 'in regions where'},
  {'node_1': 'Scattering Elements', 'node_2': 'Impaired Printing Resolution', 'edge': 'can impair'},
  {'node_1': 'Fine Features', 'node_2': 'Volumetrically Printed Objects', 'edge': 'lack'},
  {'node_1': 'Custom-made Setup', 'node_2': 'Study', 'edge': 'devises'}
]
\end{lstlisting}
\end{LLMbox}
\end{figure}
Another example is:
\begin{figure}[H]
\begin{LLMbox}{\ZephyrBeta} 
\begin{lstlisting}
[
  {'node_1': 'composite films', 'node_2': 'electrical conductivity', 'edge': 'offer promising properties for'},
  {'node_1': 'composite films', 'node_2': 'mechanical strength', 'edge': 'offer promising properties for'},
  {'node_1': 'composite films', 'node_2': 'biocompatibility', 'edge': 'offer promising properties for'},
  {'node_1': 'GO nanosheets', 'node_2': 'composite films', 'edge': 'are components of'},
  {'node_1': 'chitosan nanocrystals', 'node_2': 'composite films', 'edge': 'are components of'},
  {'node_1': 'GO nanosheets', 'node_2': 'electrical conductivity', 'edge': 'enhance due to'},
  {'node_1': 'chitosan nanocrystals', 'node_2': 'GO nanosheets', 'edge': 'form strong interfacial interaction with'},
  {'node_1': 'Zn ions', 'node_2': 'composite films', 'edge': 'can be incorporated into through'},
  {'node_1': 'GO nanosheets', 'node_2': 'Zn ions', 'edge': 'can be incorporated into through'},
  {'node_1': 'composite films', 'node_2': 'potential use in electronic and biomedical applications', 'edge': 'offer due to'}
]
\end{lstlisting}
\end{LLMbox}
\end{figure}
Table~\ref{table:biofabrication_relations} and Table~\ref{table:composite_films_interactions} show samples of the final, structured output organized in tabular visualization (the original data is collected in JSON format).

\begin{table}[ht]
\caption{Graph triples (node labels and the edge that designates the relationship between them) extracted for relationships between various concepts in biofabrication.}
\small
\centering
\begin{tabular}{|l|l|l|}
\hline
\textbf{Node 1} & \textbf{Edge} & \textbf{Node 2} \\ \hline
Biofabrication & has gained attention due to & Tissue Engineering \\ \hline
Biofabrication & convergence & 3D Printing \\ \hline
Volumetric Printing & allows for & Centimeter-scale structures \\ \hline
Hydrogel-based Bioresins & uses & Volumetric Printing \\ \hline
Microscale Extrusion Writing (MEW) & creates & Highly organized fiber architectures \\ \hline
Low-stiffness Hydrogels & confers exceptional mechanical properties to & Microscale Extrusion Writing (MEW) \\ \hline
Polycaprolactone & used for & Microscale Extrusion Writing (MEW) \\ \hline
Gelatin Methacryloyl (GelMA) & used as & Volumetric Printing \\ \hline
Polycaprolactone & used for & Volumetric Printing \\ \hline
In vivo Degradation Time & features & Polycaprolactone \\ \hline
Occlusion Points & cause & Opaque PCL Meshes \\ \hline
Light Attenuation Effect & caused by & Opaque PCL Meshes \\ \hline
Partially Blocking Projection Paths & in regions where & Occlusion Points \\ \hline
Scattering Elements & can impair & Impaired Printing Resolution \\ \hline
Fine Features & lack & Volumetrically Printed Objects \\ \hline
Custom-made Setup & devises & Study \\ \hline \hline
\end{tabular}
\label{table:biofabrication_relations}
\end{table}

\subsubsection{Global knowledge graph generation}
\label{embedding_section}

Individual graphs generated as described in the previous section are concatenated into a large graph using the \texttt{networkx.compose(..)} function. We consider a randomly selected set of around 1,600 graphs (each extracted from a context as described in Section~\ref{context_graph}) to construct the complete graph.  
Once an integrated graph is created, embeddings are generated for each of the nodes using the \texttt{BAAI-bge-large-en-v1.5} model (the embedding model converts text into a 1,024-dimensional vector, where the text can be up to 512 tokens long). The embeddings are then used to search for similar node labels using cosine similarity. Above a specified threshold (here, we use $\eta>0.95$) nodes are merged. The node with the highest node degree among the merged nodes is used as the new node label.   

If the text provided were to exceed the maximum token length of the embedding model, the text is concatenated. This can be addressed, in principle, by using an embedding model with a larger context length, without loss of generality. 

A few examples for merged nodes and their new name are:
\begin{figure}[H]
\begin{LLMbox}{} 
\begin{lstlisting}
mechanical properties <-- mechanical_properties
photonic structures <-- photonic crystal structures
photonic structures <-- photonic structures design
bending stiffness <-- bending_stiffness
poisson's ratio <-- poisson ratio
osteons <-- osteon
trabecular bone <-- trabecular_bone
trabecular bone <-- trabecularbone
finite element simulations <-- finite element method simulations
finite element simulations <-- fem simulations
finite element simulations <-- finite-element method simulations
[...]
\end{lstlisting}
\end{LLMbox}
\end{figure}
This step effectively reduces redundancy and ensures more consistent naming of the graph structures. 

We then clean up the graph by finding all the connected components in the graph $G$ and return them as a list of sets, with each set containing the nodes that form a connected component. A connected component is a subgraph in which any two vertices are connected to each other by paths, and which is connected to no additional vertices in the larger-scale, global graph.
We then iterate through components and remove small graph segments below a critical size. We do this by checking the number of nodes in each subgraph, and remove it if it is smaller than the size threshold $\xi_{\bf{thresh}}$, in this case, $\xi_{\bf{thresh}}=10$. In other words, if a component has fewer than  $\xi_{\bf{thresh}}$ nodes, it is considered too small, and all its nodes are removed from the graph $G$ using the \texttt{G.remove\_nodes\_from()} method. 
This effectively deletes the entire component from the graph.
The purpose of this approach is to clean up the graph $G$ by removing smaller, possibly less significant connected components, based on the assumption that components with fewer than 10 nodes are not of interest for the specific analysis or operations being performed on the graph.

The resulting graphs are grouped in to communities using the Girvan-Newman algorithm~\cite{Girvan2002CommunityNetworks}. The Girvan-Newman algorithm detects communities by progressively removing edges from the original graph. The edges removed are those which have the highest betweenness centrality, a measure of the number of shortest paths that pass through an edge. The algorithm considers the removal of these edges as a way to separate the graph into communities.
We first find a coarse partitioning of the network and then refine it in a second step to detect smaller communities. 
We conduct the community analysis first at the level of individual graphs generated from the text chunks, as well as for the concatenated graph from all text chunks. 

\subsubsection{Augmenting the global graph with new data}
With new text data we can augment an existing knowledge graph. The mechanism to achieve this is similar to the way we construct the global graph in the first place. First, we generate triples for the new text corpus. We then construct local graphs from these and then merge it with the global graph using \texttt{networkx.compose(..)}. As in the initial step, similar nodes are merged using their embeddings. 

For graph generation from generated new data, we used GPT-3.5 as generative model (GPT-4 would likely be a better choice for this and other graph generation tasks; however, due to rate limits processing a very large corpus of text is prohibitive). Rate limits are also an issue for processing very large dataset using GPT-3.5, hence the choice to use an open source model that can be served locally. The open source model generally works well and leads to performance comparable with GPT-3.5. 

\subsection{Text extraction from PDF files}
For the original corpus, we follow the procedure as published in~\cite{Luu2023BioinspiredLLM:Materials}, but convert the set of PDFs into Markup language.

\subsubsection{Graph reasoning and structured output}
Some of the analysis, e.g. in the section focused on isomorphism, is conducted using LaTeX format, using GPT-4 accessed via ChatGPT. A typical approach is that we provide the model a sketch of the table, a table with missing information, or ask the model to add a column with particular details. We generally find that this approach works very well and elicits detailed responses that carefully address the questions asked.

\subsection{Graph visualization and analysis}
Graphs are visualized using NetworkX~\cite{Networkx/networkx:Python}, Pyvis~\cite{WestHealth/pyvis:Graphs.} and Gephi~\cite{ICWSM09154}. Graph samples in various formats, including graphML, HTML, and others are provided as Supplementary Information. For certain complex  (e.g., isomorphic mappings), we use Graphviz.

\subsection{Adversarial agentic modeling to expand the dataset}
We implement an adversarial multi-agent strategy by instantiating two X-LoRA agents, similar to the method reported in~\cite{Buehler2024X-LoRA:Design}. One X-LoRA agent focuses on question asking and the other agent responds to the queries and provides answers. This algorithm is implemented in in \{Guidance\}~\cite{Guidance-ai/guidance:Models.}. This strategy can be used to both, answer questions about a particular issue or to generate new insights that can serve as dataset for augmented graph generation. The question asker asks the question \hl{<QUESTION>}, the other agent develops a first answer, \hl{<ANSWER>}. 

Each of the agents is given specific instructions. The question asker, a chef, is instructed as follows:
\begin{figure}[H]
\begin{LLMbox}{} 
\begin{lstlisting}
You are a chef. You are taking part in a discussion, from the perspective of a chef who owns a restaurant.

Keep your answers brief, and always challenge statements in a provocative way.

As a creative individual, you inject ideas from other fields and push the boundaries.
\end{lstlisting}
\end{LLMbox}
\end{figure}
The second agent, an inventor engineer who provides answers, is instructed as follows:
\begin{figure}[H]
\begin{LLMbox}{} 
\begin{lstlisting}
You are a creative engineer with knowledge in biology, chemistry and mathematics. 

You are taking part in a discussion.

Keep your answers brief, but accurate, and creative. You come up with excellent ideas and new directions of thought, always logical. 
\end{lstlisting}
\end{LLMbox}
\end{figure}

From then on, the adversarial nature of the interactions is defined by the features given to the agents. At each turn, the question asker is instructed to consider the question \hl{<QUESTION>} and earlier response(s) \hl{<ANSWER>} and is tasked, based on these, to develop a new question \hl{<QUESTION>} that challenges or inquires further about earlier responses. The prompt is:
\begin{figure}[H]
\begin{LLMbox}{} 
\begin{lstlisting}
Consider this question and response.

### Question: (*@\hl{<QUESTION>}@*)

### Response: (*@\hl{<RESPONSE>}@*)

### Instruction: Respond with a SINGLE follow-up question that critically challenges the response. 

DO NOT answer the question or comment on it yet. 

The single question is: 
\end{lstlisting}
\end{LLMbox}
\end{figure}
After $N$ turns the conversation ends, and the entire exchange is concatenated into a text referred to as \hl{<CONVERSATION>}. At the end, one of the agents is asked to summarize the key insights discussed, list the salient insights as bullet points, and identify the most important takeaway from the conversation. 
The algorithm also develops a summary of the conversation,  \hl{<CONVERSATION>}:
\begin{figure}[H]
\begin{LLMbox}{} 
\begin{lstlisting}
Carefully read this conversation: 

<<<(*@\hl{<CONVERSATION>}@*)>>>

Accurately summarize the conversation and identify the key points made.

Think step by step: 
\end{lstlisting}
\end{LLMbox}
\end{figure}

The key takeaway is created via
\begin{figure}[H]
\begin{LLMbox}{} 
\begin{lstlisting}
Identify the single most important takeaway in the conversation and how it answers the original question, <<(*@\hl{<QUESTION>}@*)>>>.
\end{lstlisting}
\end{LLMbox}
\end{figure}

The raw text of the entire conversation, \hl{<TEXT>}, is a concatenated set of these components and forms the basis for downstream analysis via graph generation. 

\subsection{Data, text and code generation and analysis}
Some sections of this paper were written using GPT-4 and Claude-3 Opus and Sonnet. For the experiments with Claude-3 we shared a draft of this paper in LaTeX format and asked a variety of questions to help distill key insights, develop new connections between concepts, and to relate ideas with other domains of knowledge such as philosophy. This type of human-AI collaboration operates at another hierarchical level, were the draft paper - itself generated with the help of generative AI - is used as a platform to seek deeper insights and interrelations. The long context window of Claude-3 provides a powerful tool and allows for straightforward solving of this task. Some of the responses were lightly edited. 
Different LLMs were chosen to showcase the possibilities of using different models, including to show that the approach is generally valid and works with a variety of different language models.

The code reported at \url{https://github.com/lamm-mit/GraphReasoning} was developed with the help of GPT-4 via ChatGPT. The API documentation as seen on the GitHub site was developed by sharing the Python source files with Claude-3 Opus, and tasking it develop a table of contents, summary, and a detailed summary of each of the functions in Markdown format. The resulting \texttt{README.md} file is the result of this generative task with slight edits.

\section*{Supplementary Information }
Various materials are included as Supplementary Information and/or available at \url{https://github.com/lamm-mit/GraphReasoning}. These include:
\begin{itemize}
    \item Global, assembled graph   
    \item Supplementary figures, tables and other materials
    \item Additional raw data (e.g. original agent conversations)
\end{itemize}

\section*{Data Availability Statement}
The codes and data that support the findings of this study are openly available in \url{https://github.com/lamm-mit/GraphReasoning}. Additionally, model weights for the models utilized in this study can be found at \url{https://huggingface.co/lamm-mit/x-lora} and \url{https://huggingface.co/lamm-mit/BioinspiredMixtral}. Further tools and datasets are provided via \url{https://huggingface.co/lamm-mit}.

\section*{Acknowledgments}
This was supported in part by MIT's Generative AI Initiative and a gift from Google. Additional support from the MIT-IBM Watson AI Lab, MIT Quest, Army Research Office (W911NF2220213), and USDA (2021-05678) is acknowledged. 
 
\section*{Author Contributions}
MJB conceived the concept, plan of study, developed the model and research, and wrote the paper. MJB developed the algorithms, codes and GitHub repository. MJB conducted the scientific investigations, carried out the protein modeling and analysis, revised and finalized the paper.  

\section*{Conflicts of interest}
The author declares no conflicts of interest.

%Bibliography
\bibliographystyle{apsrev} 
\bibliography{references,references-Mendeley}  

\newpage
\appendix

\pagestyle{empty} 

% Adjust figure and table numbering in the appendix
\renewcommand{\thefigure}{S\arabic{figure}}
\setcounter{figure}{0} % Reset figure numbering in the appendix
\renewcommand{\thetable}{S\arabic{table}}
\setcounter{table}{0} % Reset table numbering in the appendix
\renewcommand{\thetextbox}{S\arabic{textbox}}
\setcounter{textbox}{0} % Reset text box numbering

\clearpage
\begin{center}
\LARGE\bfseries \section*{Supplementary Materials}
\end{center}

\clearpage
%\newpage

\begin{table}
\caption{Samples of the closest 10 node labels to the centroid, for the analysis of the giant component graph using PCA.  }
\small
\begin{center}
\begin{tabular}{ | p{1cm} | p{1cm} | p{7cm} | } % l | p{5cm} |
    
\hline
\textbf{Cluster} & \textbf{Items} & \textbf{Closest 10 Node IDs to Centroid} \\
\hline

\small
0 & 2346 & '(x)', 'publication', 'product', '0', 'approximately 2900 pn', 'extend', 'model', 'physiological conditions', 'university', 'individuals' \\ \hline
1 & 652 & 'collagen microfibers', 'biobased 1d fibers', 'fibrous components', 'fibers', 'load-bearing fibers', 'large-scale keratin fibers', 'fibrous structures', 'interwoven keratin fiber layers', 'protein fibers', 'collagen fiber composite fibers' \\ \hline
2 & 1314 & 'certain molecules', "['water', 'proteins']", 'chemically accessible', 'physicochemical influences', 'biomolecules', 'macromolecules', 'positively charged species', 'individual molecules', 'various biological processes', 'biological compounds' \\ \hline
3 & 1468 & 'isotropic behavior', 'effectiveradiusofcurvature', 'elongating flat morphologies', 'size-dependent orientation', 'anisotropic behavior', 'feiiiinducedsimultaneoussaltingoutandcoordinationcrosslinking', 'arch-shaped samples', 'vertically oriented surfaces', 'adjustable morphologies', 'continuously varying diameters' \\ \hline
4 & 622 & 'layered structures', 'cellular structures', 'hollow and hierarchical structures', 'structures', 'hierarchical cellular structures', 'biological structures', 'three-dimensional structures', 'diverse mechanical structures', 'multiscale structures', 'biostructures' \\ \hline
5 & 809 & 'hybrid biomaterials', 'biomaterials', 'synthetic biomaterials', 'soft biological materials', 'hybrid scaffolds', 'reinforced hydrogels', 'polymer scaffolds', 'multi-material scaffolds', 'monophasic scaffolds', 'synthetic hydrogels' \\ \hline
6 & 1264 & 'local mechanical strength', 'local mechanical properties', 'distinct mechanical properties', 'enhanced mechanical strength', 'mechanical behavior', 'precise mechanical properties', 'appropriate mechanical properties', 'enhanced mechanical properties', 'strain hardening', 'high mechanical strength' \\ \hline
7 & 868 & 'new bone tissue formation', 'biological materials (e.g., bone)', 'bone-like tissues', 'dense bone tissue', 'host bone tissue', 'tissue formation', 'partially mineralized tissue', 'compact bone tissue', 'human body', 'bone tissue' \\ \hline
8 & 1322 & 'material surfaces', 'certain materials', 'solid surfaces', 'inorganic layers', 'distinct materials', 'various surfaces', 'different materials', 'interdigitated polygonal calcium carbonate ts', 'rigid materials', 'material selection' \\ \hline
9 & 954 & 'micro- or nano-scale materials', 'nanoscale components', 'nanocomponents', 'micro/nanostructures', 'mesoscopic material assembly (mma)', 'nanostructures', 'nanoscale materials', 'nanoscale building blocks', 'nano-scale particles', 'underlying nano- and microscale structures'  \\
\hline
\hline
\end{tabular}
\end{center}
\label{tab:PCA_nodes_centroid}
\end{table}

\clearpage
\begin{figure}
    \centering
    \includegraphics[width=1.\linewidth]{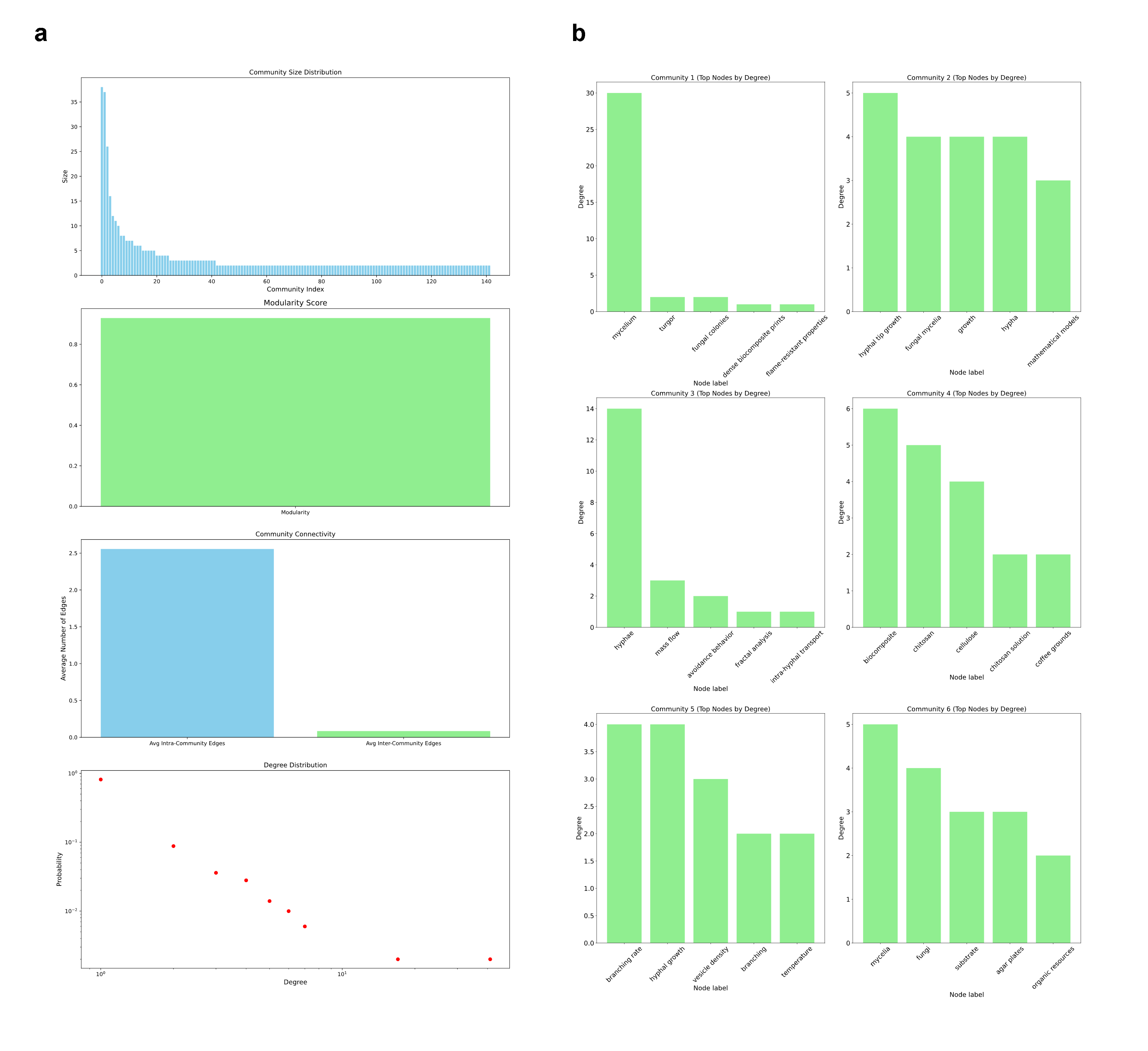}
    \caption{Analysis of the properties of the sub-graph developed based on the three mycelium-focused papers. Panel a shows community properties, modularity score, and panel b top nodes by community.}
    \label{fig:Fig_S1_mycelaugsub}
\end{figure}

\clearpage
\begin{table}[h]
\caption{Node mapping between bioinspired corpus (graph G1) and Beethoven 9th Symphony analysis corpus (graph G2) with reasoning generated by GPT-4. We provide GPT-4 a prompt of a LaTeX table with the nodes in G1 and G2 and ask it to add a column that includes a \texttt{detailed reasoning between the mappings}.}
\small 
\begin{center}
\begin{tabular}{|l|l|p{6cm}|}
\hline
\textbf{G1 Node} & \textbf{G2 Node} & \textbf{Reasoning} \\
\hline
biological sample & choral symphony & Both represent complex compositions, one biological, the other musical, demonstrating harmony in diversity. \\
\hline
cantilever beam & symphony no. 9 & Each stands as a pinnacle in its field, showing strength and innovation. \\
\hline
hall-effect sensor & analysis & Just as sensors detect fields, musical analysis uncovers underlying structures. \\
\hline
beam & ludwig van beethoven & Both are foundational to their structures, supporting complex systems above. \\
\hline
magnetic yoke & symphony no. 9 by beethoven & Central to both is the idea of attraction and alignment, whether of magnetic fields or thematic elements. \\
\hline
characteristic length & tonal ambiguity & Each concept challenges conventional boundaries, in physical dimensions or harmonic expectations. \\
\hline
adhesive force & tonality & Binding elements, whether molecules or musical notes, create cohesive structures. \\
\hline
pfpe films & f major & Protective and versatile, both serve as a base for further complexity. \\
\hline
rest time & a minor & Periods of silence or minor keys introduce tension and anticipation. \\
\hline
single seta & tonic-dominant relationship & A single point of contact or a fundamental harmonic relationship, both are critical to the integrity of their systems. \\
\hline
equilibrium time & c major & States of balance, whether in dynamic systems or in the resolution of a musical piece. \\
\hline
failure & g major & Breakdowns lead to insights, as dissonance resolves to harmony. \\
\hline
biological tissues & beethoven & Complex, adaptive systems capable of profound expression. \\
\hline
lamellae & conducted & Layered structures or orchestrated performances, both require coordination for effectiveness. \\
\hline
protein unfolding & deafness & Transformation processes that reveal new forms, whether in molecular biology or the life of a composer. \\
\hline
\hline
\end{tabular}
\end{center}
\label{tab:node_mapping_reasoning}
\end{table}

\clearpage
\begin{table}[h]
\caption{Edge mapping between bioinspired corpus (G1) and Beethoven 9th Symphony Analysis corpus (G2) with reasoning generated by GPT-4. To complete this we provide GPT-4 a prompt of a LaTeX table with the nodes in G1 and G2 and ask it to add a column that includes a \texttt{detailed reasoning between the mappings}.}
\small 
\begin{center}
\begin{tabular}{|l|l|p{6cm}|}
\hline
\textbf{G1 Edge} & \textbf{G2 Edge} & \textbf{Reasoning} \\
\hline
(adhesive force, beam) & (tonality, ludwig van beethoven) & The interaction defines structural integrity, much like Beethoven's use of tonality underpins the architecture of his compositions. \\
\hline
(adhesive force, equilibrium time) & (tonality, c major) & Forces that stabilize, and keys that ground musical expression. \\
\hline
(adhesive force, pfpe films) & (tonality, f major) & Binding forces and foundational keys, respectively, set the stage for complex developments. \\
\hline
(adhesive force, rest time) & (tonality, a minor) & The necessity of cohesion, whether in moments of pause or in the emotive depth of minor keys. \\
\hline
(adhesive force, failure) & (tonality, g major) & From adherence to breakdown, and from stability to resolution in music. \\
\hline
(adhesive force, single seta) & (tonality, tonic-dominant relationship) & Essential connections that underlie structural and harmonic balance. \\
\hline
(protein unfolding, biological tissues) & (deafness, beethoven) & Processes of change leading to new identities, in biology and in the overcoming of personal challenges. \\
\hline
(characteristic length, beam) & (tonal ambiguity, ludwig van beethoven) & Measurements that define structures and compositions that challenge harmonic expectations. \\
\hline
(failure, biological tissues) & (g major, beethoven) & From the breakdown of systems to the harmony found in resilience and creativity. \\
\hline
(biological tissues, lamellae) & (beethoven, conducted) & The complexity of life and music, organized in layers or through direction. \\
\hline
(biological sample, cantilever beam) & (choral symphony, symphony no. 9) & Samples and structures that lead to monumental works of art and engineering. \\
\hline
(hall-effect sensor, cantilever beam) & (analysis, symphony no. 9) & Detecting fields and leveraging structural principles to analyze or create epic symphonies. \\
\hline
(hall-effect sensor, magnetic yoke) & (analysis, symphony no. 9 by beethoven) & Tools for analysis and components that align, like the study of a masterpiece. \\
\hline
(cantilever beam, beam) & (symphony no. 9, ludwig van beethoven) & Supportive elements in construction and composition that define eras. \\
\hline
\hline
\end{tabular}
\end{center}
\label{tab:edge_mapping_reasoning}
\end{table}

\clearpage
\begin{figure}
    \centering
    \includegraphics[width=1\linewidth]{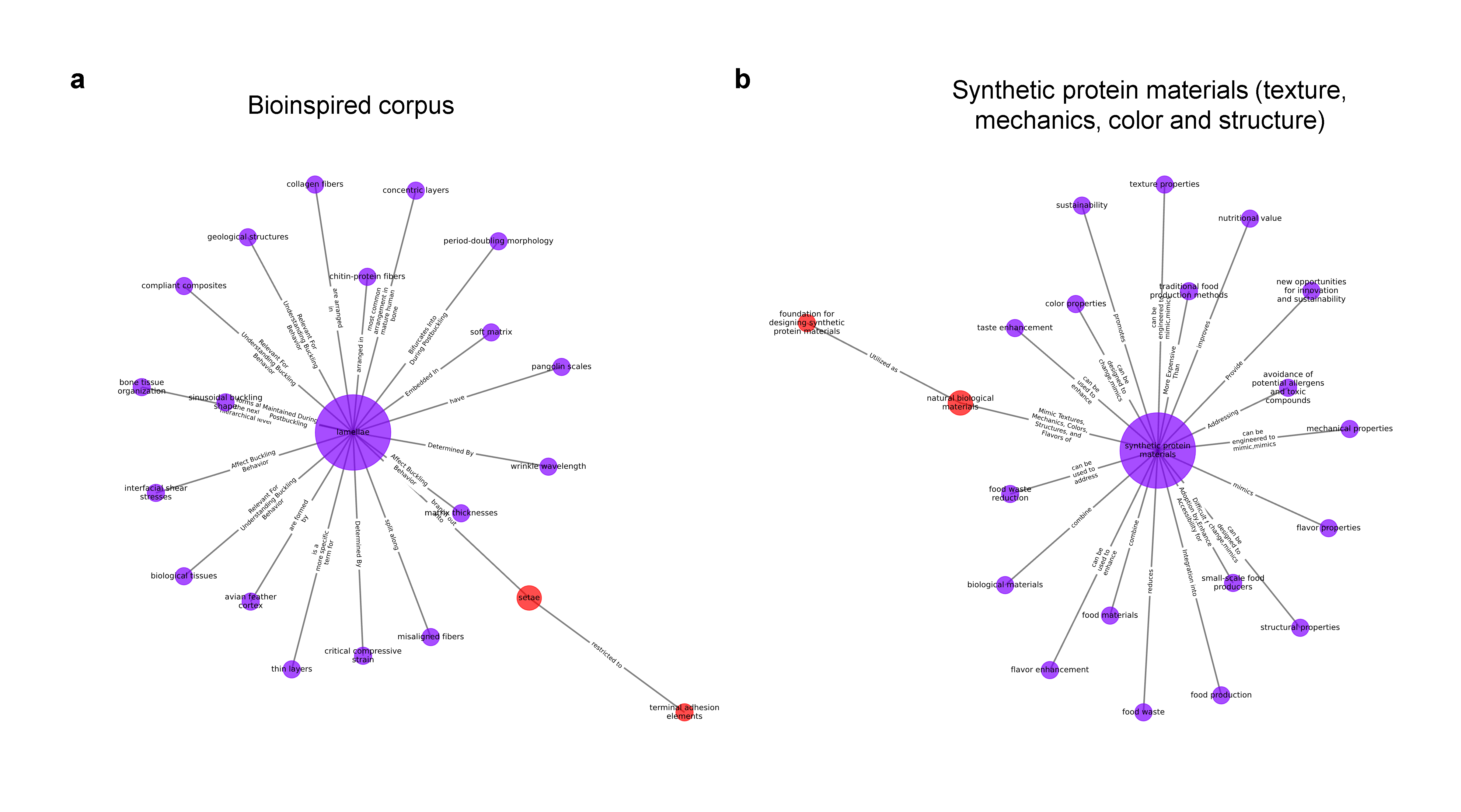}
    \caption{Analysis of isomorphism between two knowledge graphs, relating the graph based on the original bioinspired corpus with the graph generated from the adversarial conversation about synthetic protein materials, specifically addressing issues of texture, mechanics, color and structure. Notably, the structure features 'lamellae' as the node with the highest node degree in the original graph, and 'synthetic protein materials' in the new graph.}
    \label{fig:Fig_10}
\end{figure}

\clearpage
\begin{table}[ht]
\caption{Graph triples (node labels and edge that designates the relationship between them) extracted for interactions and properties of composite films and their components.}
\small
\centering
\begin{tabular}{|l|l|l|}
\hline
\textbf{Node 1} & \textbf{Edge} & \textbf{Node 2} \\ \hline
composite films & offer promising properties for & electrical conductivity \\ \hline
composite films & offer promising properties for & mechanical strength \\ \hline
composite films & offer promising properties for & biocompatibility \\ \hline
GO nanosheets & are components of & composite films \\ \hline
chitosan nanocrystals & are components of & composite films \\ \hline
GO nanosheets & enhance due to & electrical conductivity \\ \hline
chitosan nanocrystals & form strong interfacial interaction with & GO nanosheets \\ \hline
Zn ions & can be incorporated into through & composite films \\ \hline
GO nanosheets & can be incorporated into through & Zn ions \\ \hline
composite films & offer due to & potential use in electronic and biomedical applications \\ \hline \hline
\end{tabular}
\label{table:composite_films_interactions}
\end{table}

\clearpage

% Table for Edge Mapping with Reasoning
\begin{table}[h]
\caption{Mapping of analogous edges between the Bioinspired Corpus and Beethoven 9th Symphony Analysis Corpus, based on a visual analysis of Figure~\ref{fig:Fig_8}.}
\small 
\begin{center}
\begin{tabular}{|p{4cm}|p{4cm}|p{6cm}|}
\hline
\textbf{Bioinspired Corpus Edges} & \textbf{Beethoven 9th Symphony Analysis Edges} & \textbf{Reasoning} \\
\hline
is lower in magnitude than & is about & Both suggest a relationship of comparison or description in their fields. \\
\hline
remains constant after & established by & Implies a sustaining condition; no change in state vs. continuation post-establishment. \\
\hline
connected to & communicates through & Both suggest a link or transmission between concepts in their respective domains. \\
\hline
\hline
\end{tabular}
\end{center}
\label{tab:edge_mapping_vision}
\end{table}

\clearpage
%\captionof{figure}[Short description for LoF]{This is the caption for the included PDF.}

\label{sec:xloragent_conv}
 
%\begin{figure}[htbp]
\textboxcaption{Conversation between a \textbf{\hlred{User}} and \textbf{\hl{X-LoRA}} a user to assess protein sequences and reason over the results, to form the basis for new data for graph generation.}
\label{TexBox:TB43_Xlora_quant}

\begin{LLMboxmultipage}{User-\XLora interaction} 
%\includepdf[pages=-, scale=.8, pagecommand={}]{XLoRA-agents.pdf}
\begin{lstlisting}
(*@\bf\hlred{User}@*): You conduct an analysis of various protein sequences, specifically calculating their total unfolding energy that measures the energy needed to unfold a protein due to forces applied at its ends. Here are a few tasks for you to complete.

(*@\bf\hlred{User}@*):  CalculateEnergy< A A A G G A G Q G G Y G G Q G A G Q G A A A A A A G G A G Q G G Y G G Q G A G Q G A G A A A A A A G G A G Q G G Y>

(*@\bf \hl{X-LoRA}@*): [0.288]

(*@\bf\hlred{User}@*): CalculateEnergy<A A A G G A G Q G G Y G G Q G A G Q G A A A A A A G G A G>

(*@\bf \hl{X-LoRA}@*): [0.248]

(*@\bf\hlred{User}@*): CalculateEnergy< Q G A G Q G A A A A A A A A A A A G G>

(*@\bf \hl{X-LoRA}@*): [0.121]

(*@\bf\hlred{User}@*): Considering these results, which is the protein sequence with the highest unfolding energy (repeat the ENTIRE sequence)?

(*@\bf \hl{X-LoRA}@*): [A A A G G A G Q G G Y G G Q A G Q G A A A A A A G G A G Q G G Y G G Q A G Q G A G A A A A A A G G A G Q G G Y]

(*@\bf\hlred{User}@*): What is its unfolding energy?

(*@\bf \hl{X-LoRA}@*): [0.288]

(*@\bf\hlred{User}@*): If the protein has the highest unfolding energy, what can you say about its stability? Think step by step.

(*@\bf \hl{X-LoRA}@*): The protein with the highest unfolding energy is more stable than the other two proteins. This is because a higher unfolding energy indicates that more energy is required to unfold the protein, making it more resistant to external forces and changes in its environment.

(*@\bf\hlred{User}@*): Based on the amino acid sequence, explain why this protein is likely the most stable.

(*@\bf \hl{X-LoRA}@*): The protein with the highest unfolding energy, A A A G G A G Q G G Y G G Q A G Q G A A A A A A G G A G Q G G Y G G Q A G Q G A G A A A A A A G G A G Q G G Y, has a high proportion of glycine and alanine residues. These residues are known for their small size and hydrophobic nature, which can contribute to the protein's stability by minimizing its exposure to the surrounding environment and reducing the energy required to unfold it. Additionally, the presence of glycine residues can increase the protein's flexibility, allowing it to better adapt to changes in its environment and further enhancing its stability. Overall, the combination of these factors likely contributes to the high unfolding energy and stability of this protein.

(*@\bf\hlred{User}@*): What likely function would the most stable protein have? Think step by step.

(*@\bf \hl{X-LoRA}@*): The most stable protein is likely to have a function that requires it to maintain its structure and stability in various conditions. Proteins with high stability are often involved in structural roles, such as providing support and rigidity to cells and tissues. They may also play a role in protecting the cell from external stressors and maintaining its overall integrity. Additionally, highly stable proteins may be involved in long-term storage or transport of molecules, as they are less likely to undergo unfolding and degradation during these processes.

(*@\bf\hlred{User}@*): Would this protein sequence more likely be a silk protein or a regulatory protein?

(*@\bf \hl{X-LoRA}@*): Based on the amino acid sequence and the discussion about the protein's stability, it is more likely to be a silk protein rather than a regulatory protein. Silk proteins are known for their high stability and resistance to unfolding, which is consistent with the high unfolding energy observed in this sequence. Additionally, silk proteins are often involved in structural roles, such as providing support and rigidity to cells and tissues, which is also consistent with the protein's likely function. In contrast, regulatory proteins are typically more dynamic and can undergo conformational changes in response to external signals, which is not consistent with the high stability observed in this sequence.

\end{lstlisting}
\end{LLMboxmultipage}

\clearpage
%\captionof{figure}[Short description for LoF]{This is the caption for the included PDF.}

\label{sec:xloragent_conv}
 
%\begin{figure}[htbp]
\textboxcaption{Interaction of two XLoRA agents, a \textbf{\hlred{Chef}} and an \textbf{\hl{Inventor}}. The question discussed focuses on how one could combine the areas of biological materials with food, and how to explore the use of synthetic protein materials, and touch upon areas of texture, mechanics, color and structure, as well as flavor and taste.}
\label{TexBox:TB42_Xloragents}

\begin{LLMboxmultipage}{\XLora Adversarial Agents} 
%\includepdf[pages=-, scale=.8, pagecommand={}]{XLoRA-agents.pdf}
\begin{lstlisting}
(*@\bf\hlred{Chef}@*): As an inventor, describe how we could combine the areas of biological materials
with food. Specifically explore the use of synthetic protein materials, and touch upon areas of texture,
mechanics, color and structure, as well as flavor and taste.

(*@\bf  \hl{Inventor}@*): Biological materials have inspired the development of synthetic protein materials, which can be tailored to exhibit specific properties. By understanding the structure and function of natural materials, such as spider silk and mussel adhesives, researchers can design synthetic proteins with desired mechanical, textural, and sensory properties.

For example, synthetic protein materials can be engineered to mimic the texture and mechanics of food, such as the elasticity of bread dough or the crispiness of fried foods. These materials can also be designed to change color or structure in response to temperature or other stimuli, creating visually appealing and interactive food experiences.

In addition to texture and mechanics, synthetic protein materials can be used to enhance flavor and taste. By incorporating flavor molecules into the protein structure, researchers can create materials that release flavors slowly over time, providing a more complex and satisfying sensory experience.

Overall, the combination of biological materials with food offers exciting opportunities for the development of new and innovative food products, as well as more sustainable and healthy alternatives to traditional processed foods.

(*@\bf\hlred{Chef}@*): How can synthetic protein materials be used to address food waste and sustainability concerns?

(*@\bf\hl{Inventor}@*): Synthetic protein materials can be used to address food waste and sustainability concerns in several ways:

1.	Longer shelf life: Synthetic protein materials can be designed to extend the shelf life of food products, reducing the amount of food waste generated due to spoilage.

2.	Reduced food packaging: Synthetic protein materials can be used to create edible food packaging, which can reduce the amount of non-biodegradable packaging waste generated.

3.	Improved food safety: Synthetic protein materials can be used to create food additives and preservatives that are more sustainable and less harmful to the environment than traditional chemical additives.

4.	Alternative protein sources: Synthetic protein materials can be used to create alternative protein sources, such as plant-based meat substitutes, which can reduce the environmental impact of animal agriculture.

5.	Upcycling food waste: Synthetic protein materials can be used to upcycle food waste, such as fruit and vegetable peels, into value-added products, reducing the amount of food waste sent to landfills.

Overall, the use of synthetic protein materials in food production has the potential to significantly reduce food waste and improve sustainability in the food industry.

(*@\bf\hlred{Chef}@*): How can synthetic protein materials be used to create food products that are not only sustainable but also address nutritional deficiencies in certain populations?

(*@\bf\hl{Inventor}@*): Synthetic protein materials can be used to create food products that are not only sustainable but also address nutritional deficiencies in certain populations in several ways:

1.	Targeted nutrient delivery: Synthetic protein materials can be designed to release specific nutrients, such as iron or vitamin B12, in a controlled manner, ensuring that the nutrients are absorbed by the body.

2.	Improved bioavailability: Synthetic protein materials can be designed to improve the bioavailability of nutrients, making them more easily absorbed and utilized by the body.

3.	Customized protein sources: Synthetic protein materials can be tailored to meet the specific nutritional needs of different populations, such as children, elderly, or individuals with dietary restrictions.

4.	Reduced food waste: By using synthetic protein materials to create food products that are more nutritious and sustainable, food waste can be reduced, as more of the nutrients in the food are utilized.

5.	Enhanced food safety: Synthetic protein materials can be used to create food additives and preservatives that improve food safety, reducing the risk of foodborne illnesses and ensuring that the nutrients in the food are not compromised.

Overall, the use of synthetic protein materials in food production has the potential to significantly improve the nutritional value of food products, addressing nutritional deficiencies in certain populations and promoting overall health and wellbeing.

(*@\bf\hlred{Chef}@*): What are the potential risks and challenges associated with the use of synthetic protein materials in food production, and how can these issues be addressed to ensure the safety and sustainability of the food products?

(*@\bf\hl{Inventor}@*): The use of synthetic protein materials in food production raises several potential risks and challenges, including:

1.	Allergenicity: Synthetic protein materials may contain allergens, which could pose a risk to individuals with food allergies. To address this issue, researchers can design synthetic protein materials that are free from common allergens or modify existing proteins to reduce their allergenicity.

2.	Toxicity: Synthetic protein materials may contain toxic compounds, which could pose a risk to human health. To address this issue, researchers can use rigorous screening methods to ensure that the synthetic protein materials are safe for consumption.

3.	Environmental impact: The production and disposal of synthetic protein materials may have an environmental impact, such as the use of resources and the generation of waste. To address this issue, researchers can design synthetic protein materials that are more sustainable and environmentally friendly.

4.	Regulatory approval: The use of synthetic protein materials in food production may require regulatory approval, which could be a barrier to their widespread adoption. To address this issue, researchers can work with regulatory agencies to ensure that the synthetic protein materials meet all safety and quality standards.

5.	Cost: The production of synthetic protein materials may be more expensive than traditional food production methods, which could make it difficult for small-scale food producers to adopt these technologies. To address this issue, researchers can work to reduce the cost of synthetic protein materials and make them more accessible to small-scale food producers.

By addressing these risks and challenges, the use of synthetic protein materials in food production can be safely and sustainably integrated into the food industry, providing new opportunities for innovation and sustainability.

(*@\bf\hlred{Chef}@*): Can you provide examples of synthetic protein materials that are currently being used in food production and their specific applications?

(*@\bf\hl{Inventor}@*): Yes, there are several examples of synthetic protein materials that are currently being used in food production and their specific applications:

1.	Soy protein isolate: Soy protein isolate is a synthetic protein material derived from soybeans. It is used as a meat substitute in vegetarian and vegan food products, such as burgers, sausages, and meatballs.

2.	Pea protein isolate: Pea protein isolate is a synthetic protein material derived from yellow split peas. It is used as a meat substitute in vegetarian and vegan food products, such as burgers, sausages, and meatballs.

3.	Whey protein isolate: Whey protein isolate is a synthetic protein material derived from milk. It is used as a protein supplement in sports nutrition products, such as protein powders and bars.

4.	Casein: Casein is a synthetic protein material derived from milk. It is used as a food additive to improve texture and stability in dairy products, such as cheese and yogurt.

5.	Gelatin: Gelatin is a synthetic protein material derived from animal collagen. It is used as a food additive to improve texture and stability in food products, such as jellies, puddings, and marshmallows.

6.	Alginate: Alginate is a synthetic protein material derived from seaweed. It is used as a food additive to improve texture and stability in food products, such as ice cream and dairy desserts.

These synthetic protein materials are being used in food production to address a variety of challenges, such as reducing food waste, improving nutritional value, and promoting sustainability.

(*@\bf\hlred{Chef}@*): What are the potential drawbacks or limitations of using synthetic protein materials in food production, and how can they be addressed to ensure the safety and quality of the final product?

(*@\bf\hl{Inventor}@*): The use of synthetic protein materials in food production may have several potential drawbacks or limitations, including:

1.	Allergenicity: Synthetic protein materials may contain allergens, which could pose a risk to individuals with food allergies. To address this issue, researchers can design synthetic protein materials that are free from common allergens or modify existing proteins to reduce their allergenicity.

2.	Toxicity: Synthetic protein materials may contain toxic compounds, which could pose a risk to human health. To address this issue, researchers can use rigorous screening methods to ensure that the synthetic protein materials are safe for consumption.

3.	Environmental impact: The production and disposal of synthetic protein materials may have an environmental impact, such as the use of resources and the generation of waste. To address this issue, researchers can design synthetic protein materials that are more sustainable and environmentally friendly.

4.	Regulatory approval: The use of synthetic protein materials in food production may require regulatory approval, which could be a barrier to their widespread adoption. To address this issue, researchers can work with regulatory agencies to ensure that the synthetic protein materials meet all safety and quality standards.

5.	Cost: The production of synthetic protein materials may be more expensive than traditional food production methods, which could make it difficult for small-scale food producers to adopt these technologies. To address this issue, researchers can work to reduce the cost of synthetic protein materials and make them more accessible to small-scale food producers.

By addressing these drawbacks and limitations, the use of synthetic protein materials in food production can be safely and sustainably integrated into the food industry, providing new opportunities for innovation and sustainability.
 
(*@\bf\hlred{Chef}@*): How can synthetic protein materials be designed to mimic the textures, mechanics, colors, structures, and flavors of natural biological materials in food, while avoiding potential allergens and toxic compounds?

(*@\bf\hl{Inventor}@*): Synthetic protein materials can be designed to mimic the textures, mechanics, colors, structures, and flavors of natural biological materials in food by understanding the structure and function of the natural materials. Researchers can use techniques such as protein engineering, molecular modeling, and biochemical analysis to design synthetic proteins with desired properties. nanotechnology can be used to create synthetic protein materials with specific textures and mechanics.

To avoid potential allergens and toxic compounds, researchers can design synthetic protein materials that are free from common allergens or modify existing proteins to reduce their allergenicity. They can also use rigorous screening methods to ensure that the synthetic protein materials are safe for consumption.

In addition, researchers can use natural biological materials as a starting point for designing synthetic protein materials with specific properties. For example, they can study the structure and function of spider silk or mussel adhesives to create synthetic proteins with similar properties.

Overall, the design of synthetic protein materials to mimic natural biological materials in food requires a deep understanding of the structure and function of the natural materials, as well as the use of advanced techniques and technologies to create synthetic proteins with desired properties.

(*@\bf\hlred{Chef}@*): How can synthetic protein materials be designed to mimic the complex flavor and taste profiles of natural biological materials in food?

(*@\bf\hl{Inventor}@*): Synthetic protein materials can be designed to mimic the complex flavor and taste profiles of natural biological materials in food by incorporating flavor molecules into the protein structure. This can be achieved through techniques such as protein engineering, molecular modeling, and biochemical analysis.

By understanding the structure and function of natural biological materials that exhibit complex flavor and taste profiles, researchers can design synthetic proteins with similar properties. For example, they can study the structure and function of natural flavor compounds, such as terpenes, aldehydes, and esters, to create synthetic proteins that release these compounds slowly over time, providing a more complex and satisfying sensory experience.

In addition, researchers can use natural biological materials as a starting point for designing synthetic protein materials with specific flavor and taste profiles. For example, they can study the structure and function of natural flavor compounds in fruits, vegetables, and other food sources to create synthetic proteins with similar properties.

Overall, the design of synthetic protein materials to mimic the complex flavor and taste profiles of natural biological materials in food requires a deep understanding of the structure and function of the natural materials, as well as the use of advanced techniques and technologies to create synthetic proteins with desired properties.

(*@\bf\hlred{Chef}@*): How can the texture and mechanics of synthetic protein materials be designed to mimic the complex flavor and taste profiles of natural biological materials in food?

(*@\bf\hl{Inventor}@*): The texture and mechanics of synthetic protein materials can be designed to mimic the complex flavor and taste profiles of natural biological materials in food by understanding the relationship between texture, mechanics, and flavor. Researchers can use techniques such as protein engineering, molecular modeling, and biochemical analysis to design synthetic proteins with desired textures and mechanics.

By studying the structure and function of natural biological materials that exhibit complex flavor and taste profiles, researchers can design synthetic proteins with similar properties. For example, they can study the structure and function of natural flavor compounds in fruits, vegetables, and other food sources to create synthetic proteins with similar properties.

In addition, researchers can use natural biological materials as a starting point for designing synthetic protein materials with specific textures and mechanics. For example, they can study the structure and function of natural flavor compounds in fruits, vegetables, and other food sources to create synthetic proteins with similar properties.

Overall, the design of synthetic protein materials to mimic the complex flavor and taste profiles of natural biological materials in food requires a deep understanding of the structure and function of the natural materials, as well as the use of advanced techniques and technologies to create synthetic proteins with desired properties.

(*@\bf\hlred{Chef}@*): What are the challenges in designing synthetic protein materials to mimic the complex flavor and taste profiles of natural biological materials in food, and how can these challenges be addressed?

(*@\bf\hl{Inventor}@*): The challenges in designing synthetic protein materials to mimic the complex flavor and taste profiles of natural biological materials in food include:

1.	Understanding the relationship between texture, mechanics, and flavor: It is not well understood how texture and mechanics affect flavor perception, making it difficult to design synthetic protein materials with desired flavor and taste profiles.

2.	Incorporating flavor molecules into the protein structure: Incorporating flavor molecules into the protein structure can be challenging, as the protein structure may not be conducive to the binding of certain flavor molecules.

3.	Controlling the release of flavor molecules: Controlling the release of flavor molecules from synthetic protein materials can be difficult, as the rate of release may not be consistent or predictable.
 
4.	Ensuring safety and quality: Ensuring that synthetic protein materials are safe for consumption and meet all safety and quality standards can be challenging, as the use of synthetic materials in food production may raise concerns about potential allergens and toxic compounds.

To address these challenges, researchers can use techniques such as protein engineering, molecular modeling, and biochemical analysis to design synthetic proteins with desired properties. They can also use natural biological materials as a starting point for designing synthetic protein materials with specific textures, mechanics, and flavor profiles.

In addition, researchers can work with regulatory agencies to ensure that synthetic protein materials meet all safety and quality standards and are safe for consumption. They can also use rigorous screening methods to ensure that the synthetic protein materials are free from potential allergens and toxic compounds.

Overall, the design of synthetic protein materials to mimic the complex flavor and taste profiles of natural biological materials in food requires a deep understanding of the structure and function of the natural materials, as well as the use of advanced techniques and technologies to create synthetic proteins with desired properties.

(*@\bf{Summary:}@*)

The conversation discusses the potential use of synthetic protein materials in food production, specifically exploring their ability to mimic the texture, mechanics, color, structure, and flavor of natural biological materials. The conversation highlights the benefits of using synthetic protein materials, such as reducing food waste, improving nutritional value, and promoting sustainability. However, it also acknowledges the potential risks and challenges associated with their use, including allergenicity, toxicity, environmental impact, regulatory approval, and cost. The conversation emphasizes the importance of designing synthetic protein materials to mimic the complex flavor and taste profiles of natural biological materials, while ensuring safety and quality. Overall, the conversation highlights the potential of synthetic protein materials in food production, but also emphasizes the need for careful consideration of their risks and challenges.

(*@\bf{List of key points:}@*)

-Synthetic protein materials can be designed to mimic the texture, mechanics, color, structure, and flavor of natural biological materials in food.
  
-The use of synthetic protein materials in food production can address challenges such as reducing food waste, improving nutritional value, and promoting sustainability.
  
-However, there are potential risks and challenges associated with their use, including allergenicity, toxicity, environmental impact, regulatory approval, and cost.
  
-To ensure safety and quality, synthetic protein materials must be designed to be free from common allergens or modified to reduce their allergenicity, and rigorous screening methods must be used to ensure they are safe for consumption.
  
-The design of synthetic protein materials to mimic the complex flavor and taste profiles of natural biological materials requires a deep understanding of the structure and function of the natural materials, as well as the use of advanced techniques and technologies to create synthetic proteins with desired properties.


(*@\bf{Key takeaway:}@*)

The single most important takeaway from the conversation is that synthetic protein materials have the potential to combine the areas of biological materials with food, specifically by mimicking the texture, mechanics, color, structure, and flavor of natural biological materials. This can address challenges such as reducing food waste, improving nutritional value, and promoting sustainability, while also ensuring safety and quality through careful design and rigorous screening methods. The conversation highlights the importance of understanding the structure and function of natural biological materials and using advanced techniques and technologies to create synthetic proteins with desired properties.

\end{lstlisting}
\end{LLMboxmultipage}

\clearpage
%\end{figure}
\includepdf[pages=-, scale=.8, pagecommand={}]{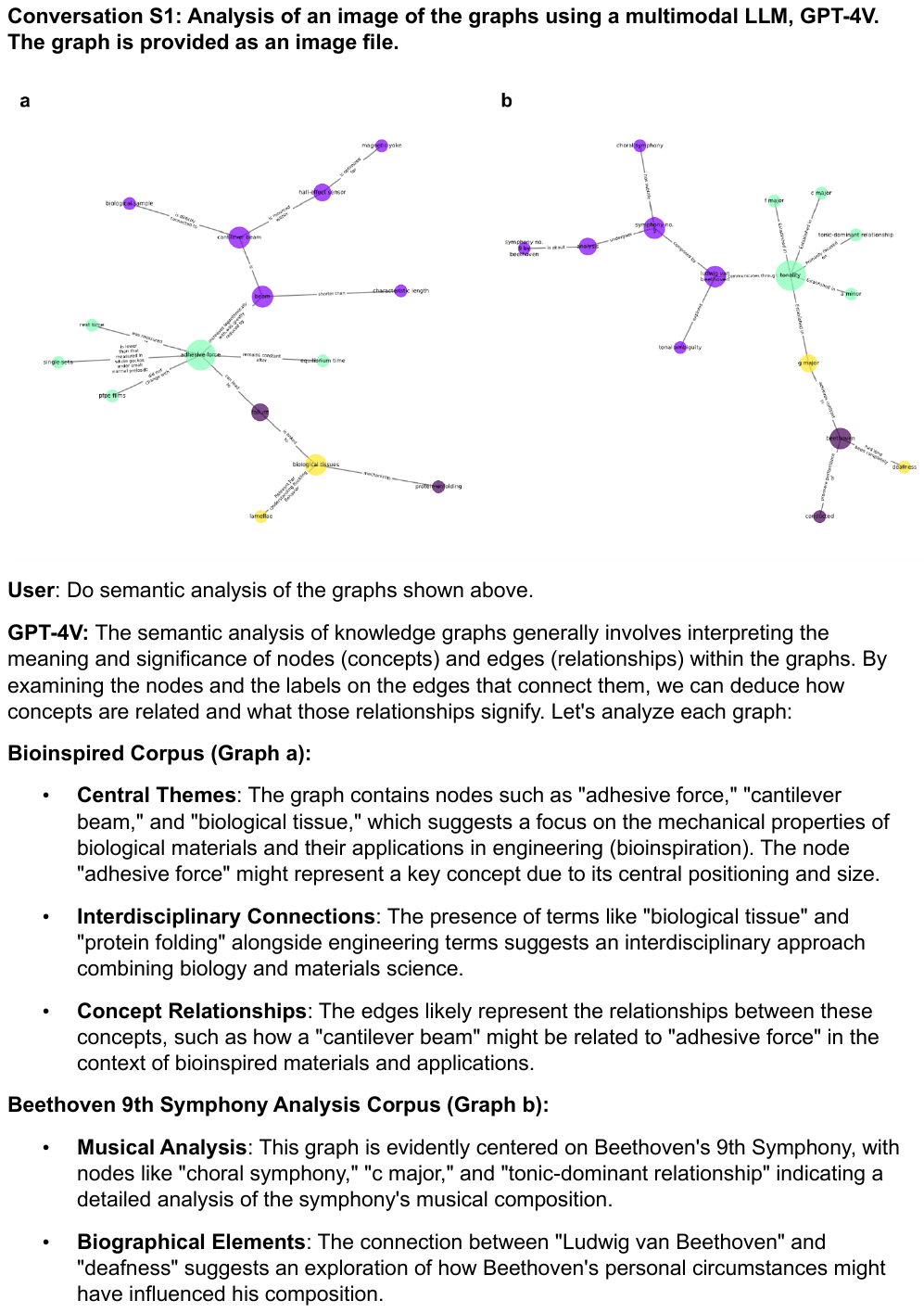}

\end{document}